\documentclass[journal]{IEEEtranTIE}
\usepackage{graphicx}
\usepackage{cite}
\usepackage{picinpar}
\usepackage{amsmath}
\usepackage{url}
\usepackage{flushend}
\usepackage[latin1]{inputenc}
\usepackage{colortbl}
\usepackage{soul}
\usepackage{multirow}
\usepackage{pifont}
\usepackage{color}
\usepackage{alltt}
\usepackage[hidelinks]{hyperref}
\usepackage{enumerate}
\usepackage{siunitx}
\usepackage{breakurl}
\usepackage{epstopdf}
\usepackage{pbox}

\usepackage{algorithm}
\usepackage{algorithmic}
\usepackage{subfigure}
\usepackage{booktabs}
\usepackage{amsfonts}

\begin{document}
	\title{Hierarchical Representation via Message Propagation for Robust Model Fitting}
	
	\author{
		Shuyuan Lin, 
		Xing Wang, 
		Guobao Xiao,
		Yan Yan, \emph{Member, IEEE},
		and Hanzi Wang, \emph{Senior Member, IEEE}
		\thanks{
			This work was supported by the National Natural Science Foundation of China under Grant U1605252, Grant 61872307, Grant 61702431, and by the National Key Research and Development Program of China No. 2017YFB1302400. (Corresponding author: Hanzi Wang.)
			
			S. Lin, X. Wang, Y. Yan and H. Wang are with the Fujian Key Laboratory of Sensing and Computing for Smart City, the School of Informatics, Xiamen	University, Xiamen 361005, China (e-mail: swin.shuyuan.lin@gmail.com; xingwang\_xmu@163.com; yanyan@xmu.edu.cn; Wang.Hanzi@gmail.com).			
			
			G. Xiao is with the College of Computer and Control Engineering, Minjiang University, Fuzhou 350108, China  (e-mail: gbx@mju.edu.cn).
		}
		\vspace{-5ex}		
	}
	
	\maketitle	
	
	\begin{abstract}
		In this paper, we propose a novel hierarchical representation via message propagation (HRMP) method for robust model fitting, which simultaneously takes advantages of both the consensus analysis and the preference analysis to estimate the parameters of multiple model instances from data corrupted by outliers, for robust model fitting. Instead of analyzing the information of each data point or each model hypothesis  independently, we formulate the consensus information and the preference information as a hierarchical representation to alleviate the sensitivity to gross outliers. Specifically, we firstly construct a hierarchical representation, which consists of a model hypothesis layer and a data point layer. The model hypothesis layer is used to remove insignificant model hypotheses and the data point layer is used to remove gross outliers. Then, based on the hierarchical representation, we propose an effective hierarchical message propagation (HMP) algorithm and an improved affinity propagation (IAP) algorithm to prune insignificant vertices and cluster the remaining data points, respectively. The proposed HRMP can not only accurately estimate the number and parameters of multiple model instances, but also handle multi-structural data contaminated with a large number of outliers. Experimental results on both synthetic data and real images show that the proposed HRMP significantly outperforms several state-of-the-art model fitting methods in terms of fitting accuracy and speed.
	\end{abstract}
	
	
	\markboth{}%
	{}
	
	\definecolor{limegreen}{rgb}{0.2, 0.8, 0.2}
	\definecolor{forestgreen}{rgb}{0.13, 0.55, 0.13}
	\definecolor{greenhtml}{rgb}{0.0, 0.5, 0.0}

	\section{Introduction}
	Robust geometric model fitting, which aims to estimate model parameters from data contaminated with outliers and noise, is a fundamental problem in computer vision. Robust model fitting is to extract meaningful structures from images, such as the shapes of objects in images, the geometric transformation between two images, the position and motion trajectory of moving objects in an image sequence, etc. In the above computer vision tasks, model parameters can be estimated by using robust model fitting methods. Moreover, the observed data in real-world scenes are usually corrupted by noise and outliers, which may be caused by measurement sensors, mismatches, multiple structures, and so on. These data often contain the unknown number of model instances. These challenges may result in biased solutions in electronic industry systems. It is worth mentioning that model fitting methods can effectively improve the tolerance to noise and outliers, and they have been applied to various applications in modern electronic industries, including mobile robots \cite{li2018visual}, automatic parking \cite{suhr2016automatic}, lane detection \cite{wu2009dynamic}, motion segmentation \cite{hoang2017motion}, fundamental/homography matrix estimation \cite{raguram2013usac}, etc. In practice, robust geometric model fitting can be used to improve the stability and reliability of electronic industry systems, especially when multi-structural data are corrupted by severe noise, pseudo- and gross-outliers \cite{lai2018robust}.

	During the past few decades, a large number of robust geometric model fitting methods (e.g., \cite{FischlerBolles1981,chin2009robust,wang2012simultaneously,wang2019searching,torr2000mlesac,chum2003locally,chum2005matching,frahm2006ransac}) have been proposed. For example, random sample consensus (RANSAC) \cite{FischlerBolles1981}, which is one of the most popular model fitting methods, and its variants \cite{torr2000mlesac,chum2003locally,chum2005matching,frahm2006ransac} have been widely used to estimate model parameters due to their robustness to noise and outliers. Specifically, a set of model hypotheses is firstly generated by randomly sampling some minimal subsets of data points (e.g., three points for a circle, four correspondences for a homography matrix). Then, the model parameters are estimated from the generated model hypotheses. 	The above two steps are repeated to handle multi-structural data until all model instances are estimated. Nevertheless, if the current model parameters are incorrectly estimated, the remaining model parameters cannot be correctly estimated in the subsequent process.
	
	Generally speaking, robust geometric model fitting methods \cite{chin2009robust,toldo2008robust,magri2014t,wang2012simultaneously,wang2019searching,magri2015robust,magri2016multiple} can be coarsely classified into the consensus analysis based methods and the preference analysis based methods. The consensus analysis based fitting methods (e.g., adaptive kernel-scale weighted hypotheses (AKSWH) \cite{wang2012simultaneously}, density guided sampling and consensus (DGSAC) \cite{tiwari2018dgsac}, random sample coverage (RansaCov) \cite{magri2016multiple} and mode-seeking on hypergraphs fitting (MSHF) \cite{wang2019searching}) exploit the consensus set corresponding to each model hypothesis for distinguishing inliers from outliers. The preference analysis based fitting methods (e.g., kernel fitting (KF) \cite{chin2009robust}, Jaccard distance based clustering (J-linkage) \cite{toldo2008robust}, Tanimoto distance based clustering (T-linkage) \cite{magri2014t}, robust preference analysis (RPA) \cite{magri2015robust} and density preference analysis (DPA) \cite{tiwari2016robust}) focus on the preference of each data point to the significant model hypotheses for clustering data points. However, there are some limitations when only using the consensus analysis or the preference analysis. That is, the consensus analysis based methods are sensitive to redundant model instances (i.e., the model instances with slightly different parameters to the true structures), while the preference analysis based methods usually require a user-specified threshold to remove outliers.
	
	To solve the above problems, we present a new and effective hierarchical representation via message propagation based fitting method, called HRMP. The proposed HRMP aims to efficiently and accurately fit multi-structural data contaminated with noise and outliers. Firstly, we propose a hierarchical representation, which consists of a model hypothesis layer (where each vertex corresponds to a model hypothesis), a data point layer (where each vertex corresponds to a data point) and edges (where each edge has a weighting score). The hierarchical representation effectively combines both advantages of the consensus analysis and the preference analysis. Secondly, we develop a hierarchical message propagation (HMP) algorithm to prune the two layers of the hierarchical representation. The messages (including both the preference information  and the consensus information) are propagated between the vertices of the two layers for pruning insignificant vertices. Such a way can effectively reduce the computational complexity of HRMP. Thirdly, we propose an improved affinity propagation (IAP) algorithm, which introduces a sparse graph to obtain significant clusters for affinity propagation (AP) \cite{frey2007clustering}. This makes the proposed IAP more efficient than AP in the message propagation process. The main contributions of HRMP are summarized as follows:
	\begin{itemize}
		\item We present an effective hierarchical representation, which includes a model hypothesis layer and a data point layer, by taking advantages of both the consensus analysis and the preference analysis for model fitting. HRMP can effectively fit multi-structural data contaminated by a large proportion of outliers. 
		\item We propose a hierarchical message propagation algorithm based on the hierarchical representation to prune insignificant vertices of the two layers, which utilizes both the preference information and the consensus information, respectively. 
		\item We also propose an improved affinity propagation algorithm to cluster the pruned data points by using the Tanimoto-like similarity. 
	\end{itemize}
	
	As a result, the proposed HRMP can effectively handle both pseudo-outliers (which are the inliers to a structure of interest but they are outliers to the other structures) and gross outliers (which are the outliers that do not belong to any structures). 
	
	The remainder of the paper is organized as follows: In Section \ref{sec:relatedwork}, we review the related work. In Section \ref{sec:theproposedethod}, we propose the complete HRMP method. In Section \ref{sec:theproposedethod}, we present the experimental results on both synthetic data and real images. In Section \ref{sec:Conclusion}, we draw the conclusions.

	\section{Related Work}
	\label{sec:relatedwork}
	In this section, we briefly review some related work on robust geometric model fitting, which contains model hypothesis pruning based methods, data point pruning based methods and message propagation based methods.	
	
	\vspace{-2ex}
	\subsection{Model Hypothesis Pruning based Methods}
	\label{sec:modelhypothesispruningmethod}
	Model fitting methods often need to generate many model hypotheses (which include a large proportion of insignificant ones) to hit all true model instances in data. Therefore, model hypothesis pruning plays an important role in robust geometric model fitting as it can improve the computational efficiency and reduce the influence of bad model hypotheses. Recently, some mode seeking-based fitting methods \cite{xu1990new,comaniciu2002mean,wang2019searching,subbarao2009nonlinear} have been proposed to prune model hypotheses according to their weighting scores. These methods estimate model parameters by seeking for the peaks of model hypothesis distribution in the parameter space. For example, the randomized Hough transform method \cite{xu1990new} makes significant model hypotheses to vote for the global optimal bin in the ``Hough space'' to improve the performance of the traditional Hough transform. Mean shift \cite{comaniciu2002mean,subbarao2009nonlinear} attempt to estimate model parameters, which correspond to the peaks in the parameter space. Note that, the performance of these methods largely depends on the proportion of the generated good model hypotheses for fitting multiple model instances in data.

	\vspace{-2ex}
	\subsection{Data Point Pruning based Methods}
	\label{sec:datapointpruningmethod}
	Data point pruning methods (i.e., outlier removal) \cite{toldo2008robust,chin2009robust,magri2014t,bustos2018guaranteed} can effectively improve the performance of robust geometric model fitting, since input data are often contaminated by a large number of outliers. For example, kernel fitting (KF) \cite{chin2009robust} introduces a mercer kernel-based statistical learning approach to measure the similarity between data points, by which it can effectively prune outliers in data. However, KF needs to calculate the similarity between each pair of data points to produce an affinity matrix, which is time-consuming. J-Linkage \cite{toldo2008robust} and T-linkage \cite{magri2014t} first measure the similarity between all data points by using the Jaccard distance and the Tanimoto distance, respectively. The agglomerative clustering algorithm is then applied to identify the potential model instances in multi-structural data. However, these methods suffer from the high computational cost and they cannot properly handle the intersection of two model instances.
	The guaranteed outlier removal (GORE) algorithm \cite{bustos2018guaranteed} introduces a guaranteed outlier removal technique based on the maximum consensus, which can greatly reduce the computational complexity of the method. However, GORE relies on the mixed integer programming technique for the branch-and-bound optimization, which may lead to a compromised global optimal solution.

	\vspace{-2ex}
	\subsection{Message Propagation based Methods}
	\label{sec:MessagePropagationMethods}
	Affinity propagation (AP) \cite{frey2007clustering} is an efficient exemplar-based clustering algorithm. There are two kinds of message propagation (i.e., the responsibility propagation and the availability propagation) between data points and candidate exemplars (i.e., each examplar corresponds to a clustering center). AP exploits the advantages of the pair-wise clustering algorithm based on a similarity matrix computed by using all data points. However, the limitation of AP is that it is difficult to determine the preference values of the input data for the optimal clustering solution.	A fast affinity propagation clustering (FAP) algorithm \cite{shang2012fast} is proposed based on a multi-level graph partitioning scheme, which combines the global and local features in data. FAP improves the quality, speed and memory usage of AP in three phases (i.e., the coarsening phase, the exemplar-clustering phase and the refining phase). The subspace affinity propagation (SSAP) algorithm \cite{gan2015subspace} is proposed for clustering subspaces by using both the fuzzy subspace clustering algorithm and the attribute weights in AP. However, SSAP needs to iteratively calculate a similarity matrix, which increases the computational cost. A facility location for subspace segmentation (FLOSS) algorithm \cite{lazic2009floss} is proposed to handle the robust geometric model fitting problem by considering the trade-off between facilities (i.e., the model complexity) and customers (i.e., the data integrity). 
	Unfortunately, FLOSS requires an extra outlier detection algorithm to deal with contaminated data such as outliers. Moreover, FLOSS may fail when the outlier percentage is more than 35\%.
	
	Note that both the proposed HRMP and the two-stage message passing (TSMP) algorithm \cite{Wang2016} employ the message propagation based manner to handle the robust fitting problem. However, there are some significant differences between them: (1) The model hypotheses generated by sampling usually contain a large number of insignificant model hypotheses, which will affect the performance of model fitting. Therefore, the proposed HRMP introduces the message propagation to prune both model hypotheses and outliers to alleviate their influence, whereas TSMP only uses the message propagation to prune outliers. (2) The proposed HRMP presents an improved affinity propagation algorithm to cluster data points by selecting and refining significant clusters, while TSMP directly applies the affinity propagation algorithm to detect clustering. Furthermore, the differences between the proposed HRMP and the hierarchical voting scheme based fitting (HVF) algorithm \cite{xiao2017hierarchical} are mainly in that: (1) Instead of utilizing random sampling in HVF, HRMP employs proximity sampling to effectively generate more all-inlier minimal subsets. (2) HVF requires a user-specified cut-off threshold to remove outliers. In contrast, HRMP is able to adaptively distinguish inliers from outliers by using the GMM. (3) HVF uses a hierarchical voting scheme to vote each node, while HRMP proposes to propagate the information on the two layers of the hierarchical representation to effectively weigh the quality of the edges. (4) HVF introduces a continuous relaxation based clustering algorithm to cluster data points. However, HRMP proposes a sparse graph based affinity propagation algorithm to group the data points, by which the robustness of HRMP can be effectively improved.
	
	In summary, the proposed HRMP not only exploits the hierarchical representation for multi-structural data, but also utilizes the message propagation to remove insignificant model hypotheses and outliers for higher efficiency. Moreover, HRMP does not rely on the trade-off between facilities and customers, but it directly estimates the parameters of model instances without requiring a user to specify the number of model instances in data. Meanwhile, HRMP can effectively handle contaminated data, especially for data with a majority of outliers.

	\section{Hierarchical Representation based Modeling}
	\label{sec:theproposedethod}
	In this section, we firstly present a novel hierarchical representation via message propagation (HRMP) method, which combines the consensus information and the preference information, for robust geometric model fitting. We then propose a hierarchical message propagation (HMP) algorithm to prune insignificant model hypotheses and gross outliers. Finally, we propose an improved affinity propagation (IAP) algorithm to adaptively cluster data points in the pruned data point layer without requiring a user to specify the number of clusters, by which the number of model instances and model parameters can be estimated.
	
	\begin{figure}[!t]\centering
		\vspace{-4ex}
		\begin{minipage}[t]{.3\textwidth}
			\centering
			\centerline{\includegraphics[width=1\textwidth]{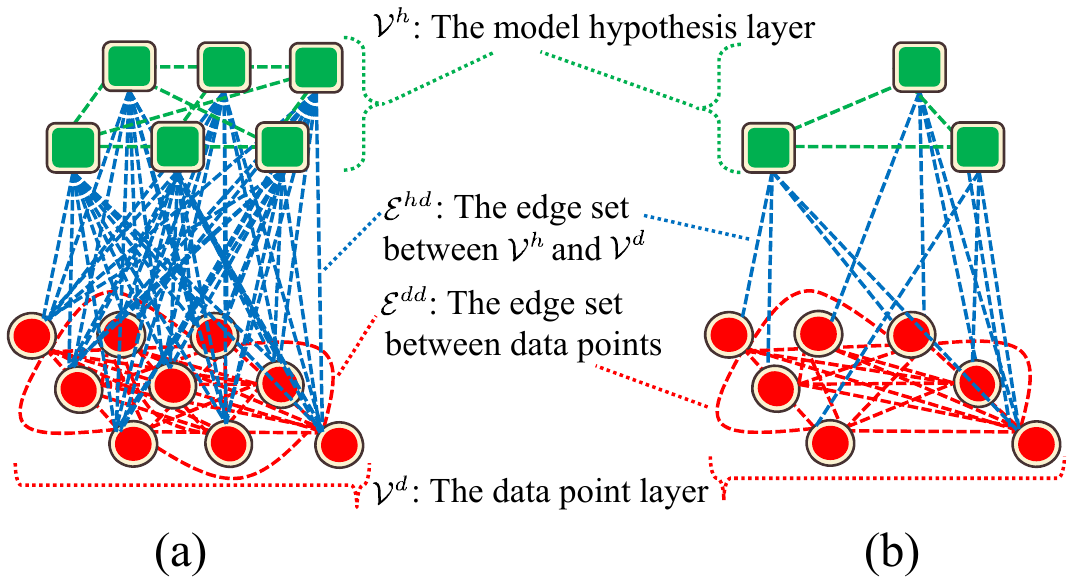}}
		\end{minipage}%
		\vspace{-1.5ex}
		\caption{An illustration of hierarchical representation. (a) The constructed hierarchical representation includes a model hypothesis layer (i.e., $\mathcal{V}^{h}$, marked by the squares in the green colors), a data point layer (i.e., $\mathcal{V}^{d}$, marked by the circles in the red colors) and the corresponding edge sets (i.e., $\mathcal{E}^{hd}$ and $\mathcal{E}^{dd}$). (b) The pruned hierarchical representation.}
		\label{fig:HRMP}
	\end{figure}

	\vspace{-2ex}
	\subsection{Overview of the Hierarchical Representation}
	\label{twolayernetworks}
	A hierarchical representation is defined as $\mathcal{H}=\{\mathcal{V}^{h},\mathcal{V}^{d},\mathcal{E}^{hd},\mathcal{E}^{dd},\mathcal{W}^{hd},\mathcal{W}^{dd}\}$, where $\mathcal{V}^{h}$ and $\mathcal{V}^{d}$ are two vertex sets corresponding to the model hypothesis layer and the data point layer, respectively. 
	$\mathcal{E}^{hd}$ and $\mathcal{W}^{hd}$ represent the edge set and the corresponding weighting score set between $\mathcal{V}^{h}$ and $\mathcal{V}^{d}$.
	$\mathcal{E}^{dd}$ and $\mathcal{W}^{dd}$ represent the edge set and the corresponding weighting score set between two vertices in $\mathcal{V}^{d}$. 	Each edge of the hierarchical representation is encoded as follows:	Firstly, $M$ model hypotheses $\pmb{\theta}=\{\theta_m\}_{m=1}^M$ (i.e., each model hypothesis corresponds to a vertex) are generated by employing the proximity sampling \cite{toldo2008robust} from data $\mathbf{X}=\{x_i\}_{i=1}^N$ (where $N$ is the number of data points), and they are used to construct a model hypothesis layer $\mathcal{V}^{h}$. After that, all data points are used to construct as a data point layer $\mathcal{V}^{d}$. Next, each vertex of $\mathcal{V}^{h}$ and $\mathcal{V}^{d}$ is also connected to construct the edge set $\mathcal{E}^{hd}$. Each pair of vertices in $\mathcal{V}^{d}$ is connected with each other to construct the edge set $\mathcal{E}^{dd}$ (see Fig. \ref{fig:HRMP} (a)). 
	
	In order to measure the quality of each edge between $\mathcal{V}^{h}$ and $\mathcal{V}^{d}$, we introduce a weighting function as follows:
	\begin{equation}
		\begin{aligned}
			\label{equ:weightingfunction}
			\omega^{hd}_{(i, m)} &=\left\{ \begin{array} {r@{\quad \quad} l}
				e^{(-r^h_{(i, m)}/\sigma_m )}, & if~ r^h_{(i, m)}\leq 2.5 \sigma_m,\\
				{0}~~~~,&otherwise,
			\end{array}\right.
		\end{aligned}
	\end{equation}
	where $e^{(\cdot)}$ is the exponential function; $r^h_{(i, m)}$ is the residual value of $x_i$ with regard to  $\theta_m$; $\sigma_m$ is the inlier scale estimated by using the iterative K-th ordered scale estimator (IKOSE) \cite{wang2012simultaneously}; $\omega^{hd}_{(i, m)}$ indicates the weighting score of an edge ($\varepsilon^{hd}_{(i, m)} \in \mathcal{E}^{hd}$) connecting the $i$-th vertex of $\mathcal{V}^{d}$ (corresponding to $x_i$) and the $m$-th vertex of $\mathcal{V}^{h}$ (corresponding to $\theta_m$). The proposed hierarchical representation can effectively characterize the complex relationship between $v^{h}_m \in \mathcal{V}^{h}$ and $v^{d}_i \in \mathcal{V}^{d}$. 
	In practice, an edge connecting a vertex $v^h_m$ to another vertex $v^d_i$ will have a higher weighting score ($\omega^{hd}_{(i, m)} \in \mathcal{W}^{hd}$) if the vertex $v^d_i$ is an inlier of the true model instance. Similarly, an edge ($\varepsilon^{dd}_{(i, i')} \in \mathcal{E}^{dd}$) between a pair of vertices ($v^d_i,v^d_{i'}$) of the same model instance in $\mathcal{V}^{d}$ will have a higher weighting score ($\omega^{dd}_{(i, i')} \in \mathcal{W}^{dd}$) if the vertices are derived from the same cluster (more details are described in Sec. \ref{AffinityPropagationClustering}). 
	
	\begin{figure}[!t]\centering
		\vspace{-4ex}
		\centerline{\includegraphics[width=0.3\textwidth]{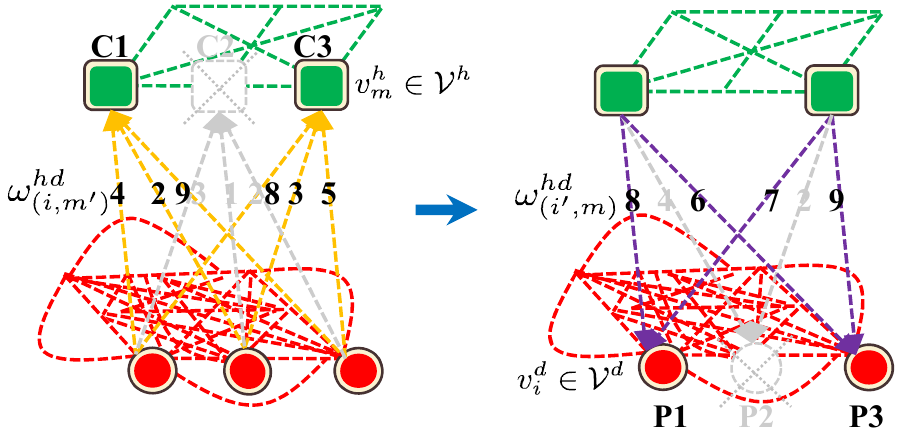}}
		\caption{An illustration of hierarchical message propagation for pruning layers. The consensus information (marked by the dotted lines in the yellow colors) and the preference information (marked by the dotted lines in the purple colors) are propagated between $\mathcal{V}^{h}$ and $\mathcal{V}^{d}$ for pruning the vertices with the lower weighting scores (i.e., C2 and P2) and the corresponding edges (marked by the dotted lines in the gray colors).}
		\label{fig:HRMP2}
		\vspace{-3ex}
	\end{figure}	
	
	\vspace{-2ex}
	\subsection{Hierarchical Representation for Model Fitting}
	\label{HierarchicalRepresentationforModelFitting}
	In this subsection, we present an effective hierarchical message propagation (HMP) algorithm, which aims to remove insignificant vertices with low weighting scores and the corresponding edges. Based on the proposed hierarchical representation, the robust fitting is designed as a message propagation problem. On one hand, the messages are propagated between the vertices of $\mathcal{V}^{h}$ and $\mathcal{V}^{d}$ for weighing the quality of each edge. 	On the other hand, we evaluate the weighting score of each vertex in the two layers to prune insignificant vertices. Thus, the meaningful information is retained while the influence of insignificant vertices is alleviated.
	
	\subsubsection{Hierarchical Message Propagation}
	\label{messagepassingbetweenthetwolayer}
	In order to extract the vertices with the high weighting scores, we propose an efficient algorithm (i.e., HMP), which iteratively propagates the two types of messages along the edges between $\mathcal{V}^{h}$ and $\mathcal{V}^{d}$. Specifically, the two types of messages are described as the consensus information $C(v_i^d,v_m^h)$ and the preference information $P(v_i^d,v_m^h)$, respectively. 
	The consensus information reveals the possibility that a vertex $v_i^d$ becomes an adjacent vertex of $v_m^h$. Meanwhile, the preference information reveals the degree that a vertex $v_i^d$ is supported by an adjacent vertex $v_m^h$.	On one hand, the quality of model hypotheses can be evaluated by using the information of data points connected to the model hypotheses within a specified inlier scale. Here, we introduce the consensus information $C(v_i^d, v_m^h)$, which calculates the preferences and weighting scores of data points connected to the model hypotheses (i.e., the messages are delivered from the vertices of $\mathcal{V}^{d}$ to those of $\mathcal{V}^{h}$). On the other hand, whether data points belong to the inliers of model hypotheses can be estimated by utilizing the model hypotheses. Similarly, we utilize the preference information $P(v_i^d,v_m^h)$, which calculates the consensus information and weighting scores of model hypotheses (i.e., the messages are delivered from the vertices of $\mathcal{V}^h$ to the those of $\mathcal{V}^d$).
	
	In practice, we initialize $P(v_i^d,v_m^h)=1/M$ (where $M$ is the number of the model hypotheses). Then, the consensus information $C(v_i^d,v_m^h)$ is delivered from the vertex $v_i^d$ to the vertex $v_m^h$. We define a consensus update rule that each vertex of $\mathcal{V}^{h}$ is suported by
	the vertices of  $\mathcal{V}^{d}$ as follows:
	\begin{equation}
		\label{consensus}
		C(v_i^d,v_m^h)=\sum_{m'=1}^{M} P(v_i^d,v_{m'}^h)\cdot \omega^{hd}_{(i, m')},
	\end{equation}
	where $\omega^{hd}_{(i, m')}$ represents the weighting score between the vertex $v_i^d$ and the vertex $v_{m'}^h$.
	
	Moreover, the preference information $P(v_i^d,v_m^h)$ is delivered from the vertex $v_m^h$ to the vertex $v_i^d$. If a vertex of $\mathcal{V}^{d}$ is strongly supported by a vertex of $\mathcal{V}^{h}$, the preference information will be assigned a large value. We also define a preference update rule that the preference information is gradually updated as follows:

	\begin{equation}
		\label{preference}
		P(v_i^d,v_m^h)=\sum_{i'=1}^{N}C(v_{i'}^d,v_m^h)\cdot \omega^{hd}_{(i', m)}.
	\end{equation}
	
	Subsequently, good vertices (i.e., significant vertices with high weighting scores) should obtain more messages than bad vertices (i.e., insignificant vertices with low weighting scores) during the propagation process. In our experiments, we find that the above iterative process usually obtains a solution within a few (e.g., two or three) iterations.

	\begin{figure}[!t]\centering
		\vspace{-4ex}
		\begin{minipage}[t]{.135\textwidth}
			\vspace{-13.5ex}
			\centering
			\centerline{\includegraphics[width=0.93\textwidth]{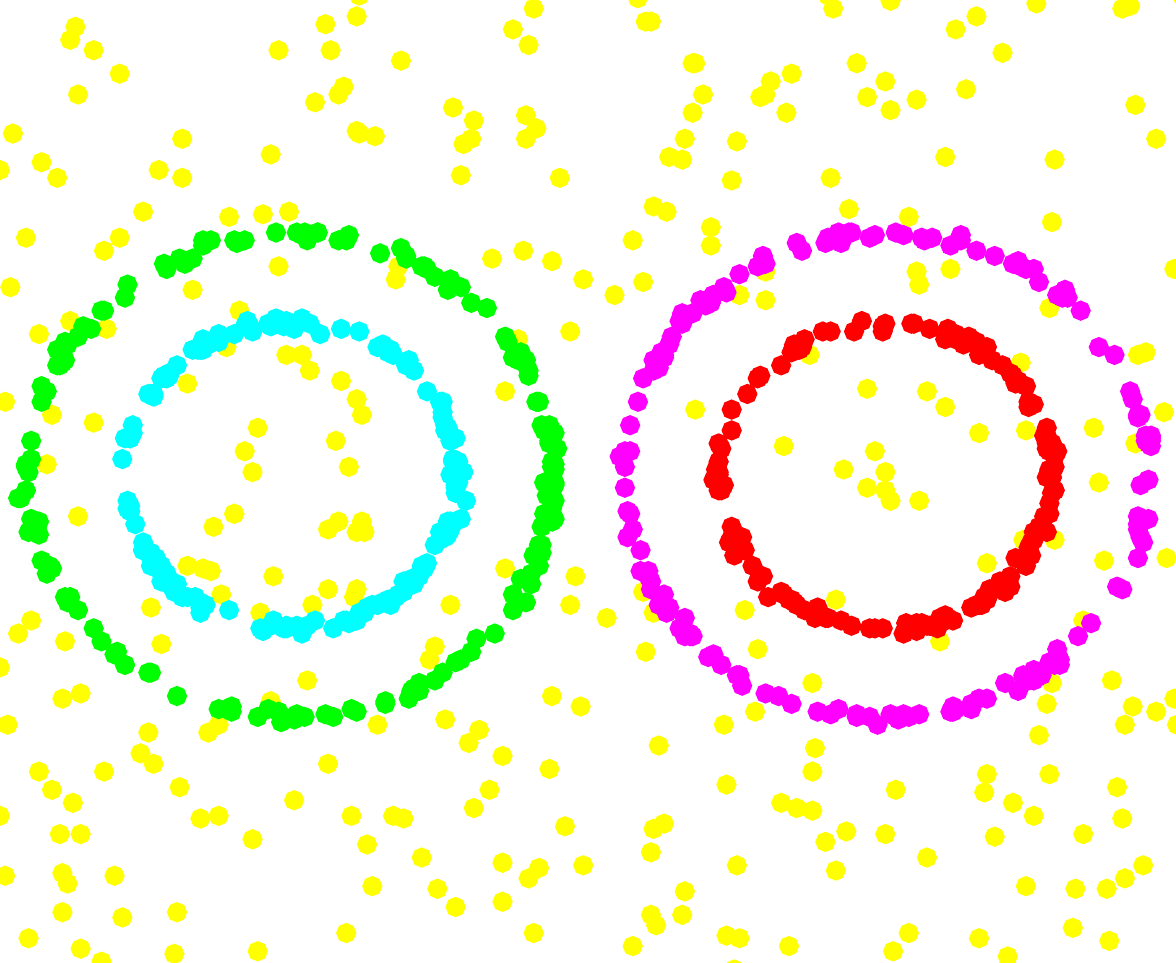}\hspace{2ex}}
			\vspace{1.7ex}
			\centerline{(a)}\medskip
		\end{minipage}%
		\begin{minipage}[t]{.17\textwidth}
			\centering
			\centerline{\includegraphics[width=0.91\textwidth]{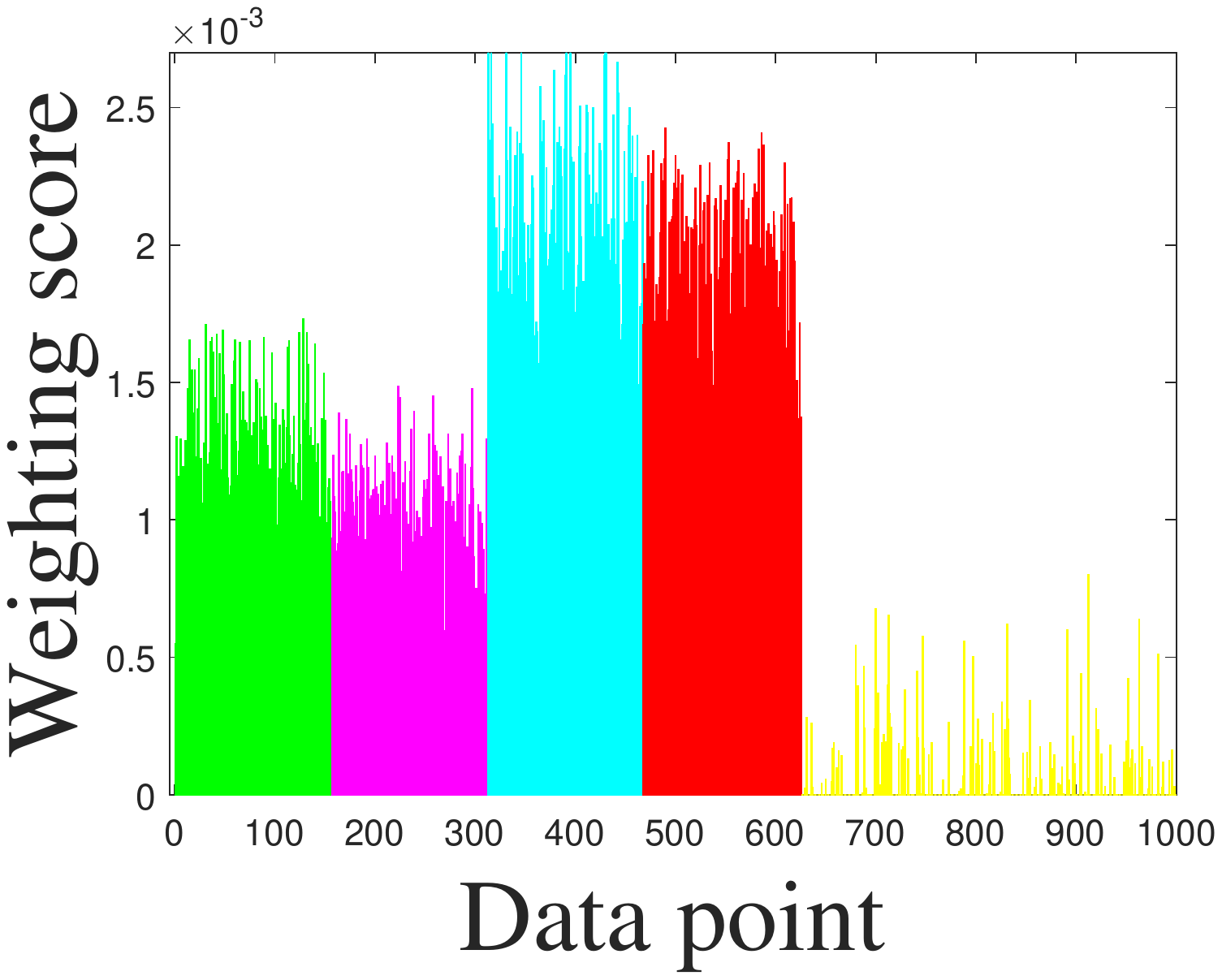}}
			\centerline{(b)}\medskip
		\end{minipage}%
		\begin{minipage}[t]{.17\textwidth}
			\centering
			\centerline{\includegraphics[width=0.86\textwidth]{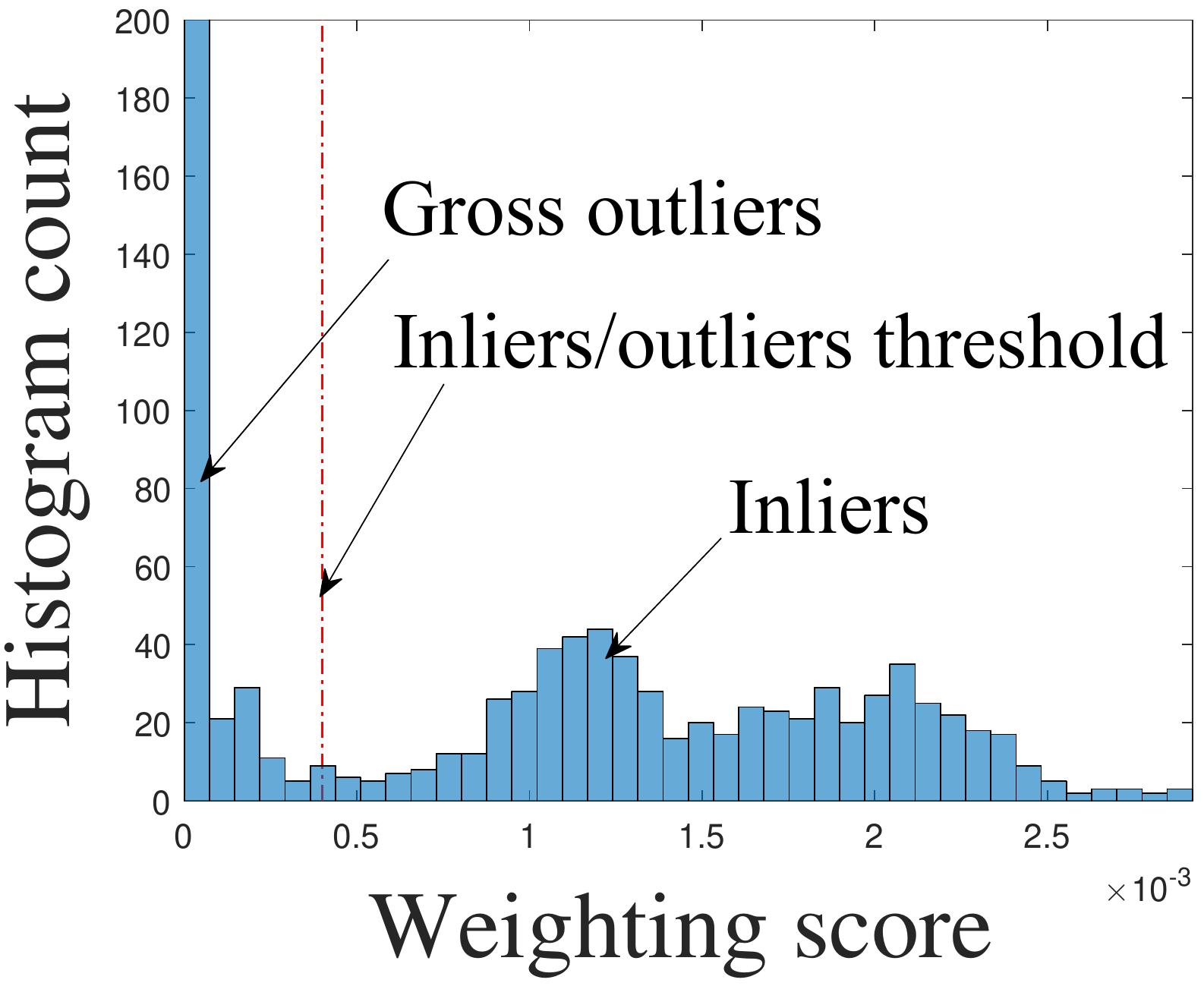}}
			\centerline{(c)}\medskip
		\end{minipage}
		\vspace{-1.5ex}
		\caption{An illustration of gross outliers pruning for the ``4 Circles'' data. (a) The data with gross outliers (marked by the solid dots in the yellow colors). (b) The weighting scores of the vertices corresponding to the 4 circles. (c) The histogram of the weighting scores.}
		\label{weightandhist}
		\vspace{-2ex}
	\end{figure}
	
	\subsubsection{Layer Pruning}
	\label{pruningthefirstlayer}
	In order to remove insignificant model hypotheses and gross outliers, we propose to prune insignificant vertices of both $\mathcal{V}^{h}$ and $\mathcal{V}^{d}$ for alleviating the sensitivity to gross outliers and pseudo-outliers. As shown in Fig. \ref{fig:HRMP2}, each vertex $v^h_m$ collects the preference information of each vertex $v^d_i$ (i.e., the preference information is delivered from $\mathcal{V}^{d}$ to $\mathcal{V}^{h}$). The vertices with low weighting scores (i.e., the insignificant model hypotheses) are then pruned by using an entropy thresholding algorithm. Next, each vertex $v^d_i$ collects the consensus information of each vertex $v^h_m$ (i.e., the consensus information is delivered from $\mathcal{V}^{h}$ to $\mathcal{V}^{d}$), and outliers are then pruned by using a Gaussian mixing model.
	
	Firstly, we introduce a weighting strategy based on the density estimation and the inlier scale estimation techniques to calculate the weighting score $\omega^{h} (v^h_m)$ of each vertex $v^h_m$ in $\mathcal{V}^{h}$ as follows:
	\begin{equation}
		\label{equ:whd}
		\omega^{h} (v^h_m) = \frac{1}{N} \sum_{i=1}^{N} \frac{\mathbb{K} (r^h_{(i, m)}/b^h)} {\sigma_m b^h},
	\end{equation}
	where $\mathbb{K}(\cdot)$ is the Epanechnikov kernel; $b^h$ is a bandwidth; $N$ is the number of the vertices in $\mathcal{V}^{d}$, which are connected to the vertex $v_m^h$. Then, the entropy thresholding algorithm \cite{wang2012simultaneously} is employed to filter out the insignificant model hypotheses.
	
	Secondly, we propose to utilize a preference-based weighting function to weigh each vertex ${\{v^{d}_i\}}_{i=1}^N$ in $\mathcal{V}^{d}$.
	The weighting score $\omega^{d}(v_i^d)$ of each vertex $v_i^d$ is calculated from the preferences of all vertices in $\mathcal{V}^{h}$ as follows:
	\begin{equation}
		\label{equ:wdd}
		\omega^{d}(v_i^d)=\sum_{m=1}^M P(v_i^d,v_m^h),
	\end{equation}
	where $\omega^{d}(v_i^d)$ represents the weighting score of the vertex $v_i^d$ in $\mathcal{V}^{d}$.
	The preference-based weighting score means that a vertex in $\mathcal{V}^{d}$, which is an inlier, should have a high weighting score since the vertex is supported by all the corresponding vertices of $\mathcal{V}^{h}$ (see Fig. \ref{weightandhist} (b)). Furthermore, we note that the histogram of the weighting scores in $\mathcal{V}^{d}$ shows two different peaks, which respectively correspond to the inliers and the outliers (see Fig. \ref{weightandhist} (c)). 
	Therefore, similar to \cite{chin2009robust}, we can separate the inliers from the outliers by using a 1D Gaussian mixture model (GMM) as follows:
	\begin{equation}
		\label{threshold}
		\mathcal{G}(\hat{\omega})=\sum_{c=1}^2 \alpha_c \mathcal{P}\{\omega^{d}(v_i^d)\big|\varphi_c,\sigma_c\},
	\end{equation}
	where $\alpha_c$ is a coefficient; $\mathcal{P}$ is a Gaussian distribution with the average $\varphi_c$ and the standard variance $\sigma_c$. Then, the average of $\varphi_1$ and $\varphi_2$ in the GMM is taken as a threshold $\phi$ for removing gross outliers.
	
	Based on the proposed pruning algorithm, the vertices with the weighting scores (in both $\mathcal{V}^{h}$ and $\mathcal{V}^{d}$) lower than the threshold $\phi$ and the corresponding edges are removed to obtain more effective pruned hierarchical representation (see Fig. \ref{fig:HRMP} (b)).

	\vspace{-2ex}
	\subsection{Cluster Detection via the Improved Affinity Propagation}
	\label{AffinityPropagationClustering}
	The efficiency of the original AP algorithm is greatly reduced when data contaminated with a great many of outliers. This is mainly because the original AP algorithm needs to handle $\gamma \times N \times N$ matrices (where $\gamma$ is the number of iterations). Similar to the message propagation based on the hierarchical representation as described in Sec. \ref{HierarchicalRepresentationforModelFitting}, we propose an improved affinity propagation (IAP) algorithm to cluster the pruned data points in $\mathcal{V}^{d}$ by using the Tanimoto-like similarity \cite{tanimoto1958elementary}. Given a set of vertices $\tilde{\mathcal{V}}^d=\{\tilde{v}_i^d\}_{i=1}^{N'}$ (where $N'\ll N$ is the number of the pruned vertices), the similarity between two vertices $\tilde{v}_i^d$ and $\tilde{v}_{i'}^d$ in $\mathcal{V}^{d}$, which are connected by an edge, is computed as follows:
	\begin{equation}
		\resizebox{.902\linewidth}{!}{$
			\begin{aligned}
				\label{similarity}
				S(\tilde{v}_i^d,\tilde{v}_{i'}^d)= \frac{\langle\mathbf{P}(\tilde{v}_i^d),\mathbf{P}(\tilde{v}_{i'}^d)\rangle}{\left \| \mathbf{P}(\tilde{v}_i^d) \right \|^2+\left \| \mathbf{P}(\tilde{v}_{i'}^d) \right \|^2 -\langle\mathbf{P}(\tilde{v}_i^d),\mathbf{P}(\tilde{v}_{i'}^d)\rangle^\prime}-1
			\end{aligned}
			$},
	\end{equation}
	where   $\mathbf{P}(\tilde{v}_i^d)=[P(\tilde{v}_i^d,v_1^h),...,P(\tilde{v}_i^d,v_{M'}^h)]$ is the preference vector of the model hypothesis layer to the data point layer (where $M'\ll M$ is the number of the pruned model hypotheses $\tilde{\mathcal{V}}^{h}$); $\|\cdot\|$ represents the induced norm; $\langle\cdot,\cdot\rangle$ represents the inner product. $S(\tilde{v}_i^d,\tilde{v}_{i'}^d)$ indicates that an edge connecting two vertices (which derive from the inliers of a true structure) has a high similarity value (corresponding to the weighting score $\omega_{(i, i')}^{dd}$).
	
\begin{figure}[!t]\centering
		\vspace{-5ex}
		\begin{minipage}[t]{.14\textwidth}
			\centering
			\centerline{\includegraphics[width=\textwidth]{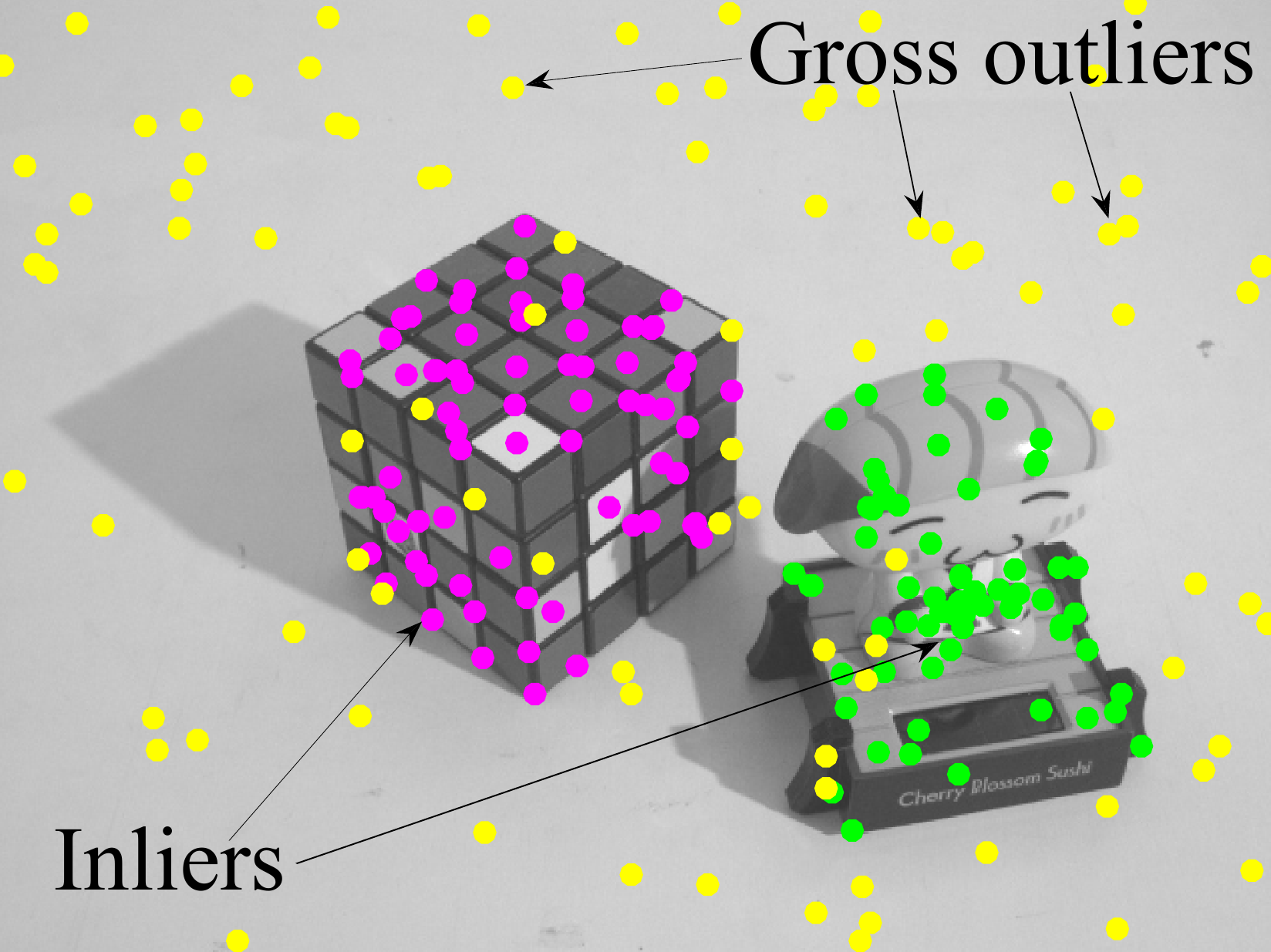}\hspace{2ex}}
			\centerline{(a)}\medskip
		\end{minipage}%
		\begin{minipage}[t]{.135\textwidth}
			\centering
			\centerline{\includegraphics[width=\textwidth]{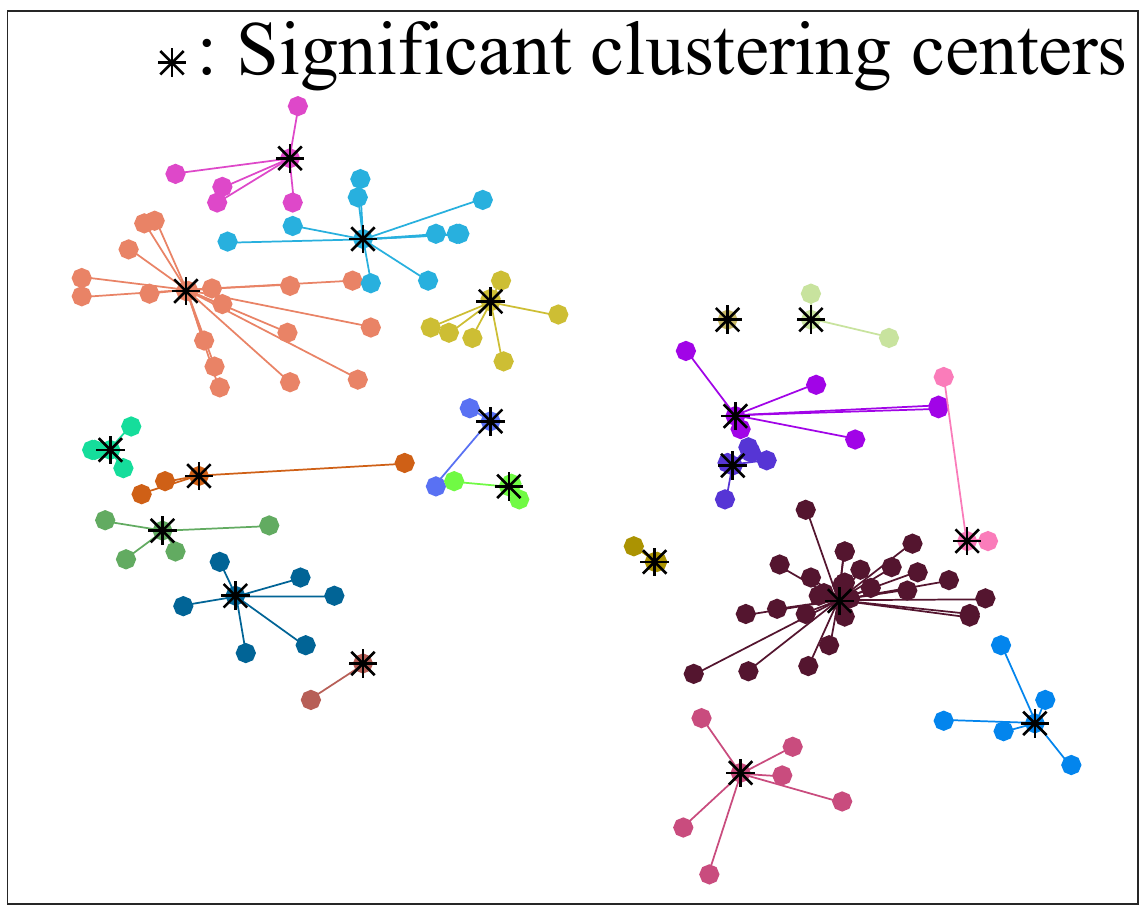}}
			\centerline{(b)}\medskip
		\end{minipage}
		\begin{minipage}[t]{.135\textwidth}
			\centering
			\centerline{\includegraphics[width=\textwidth]{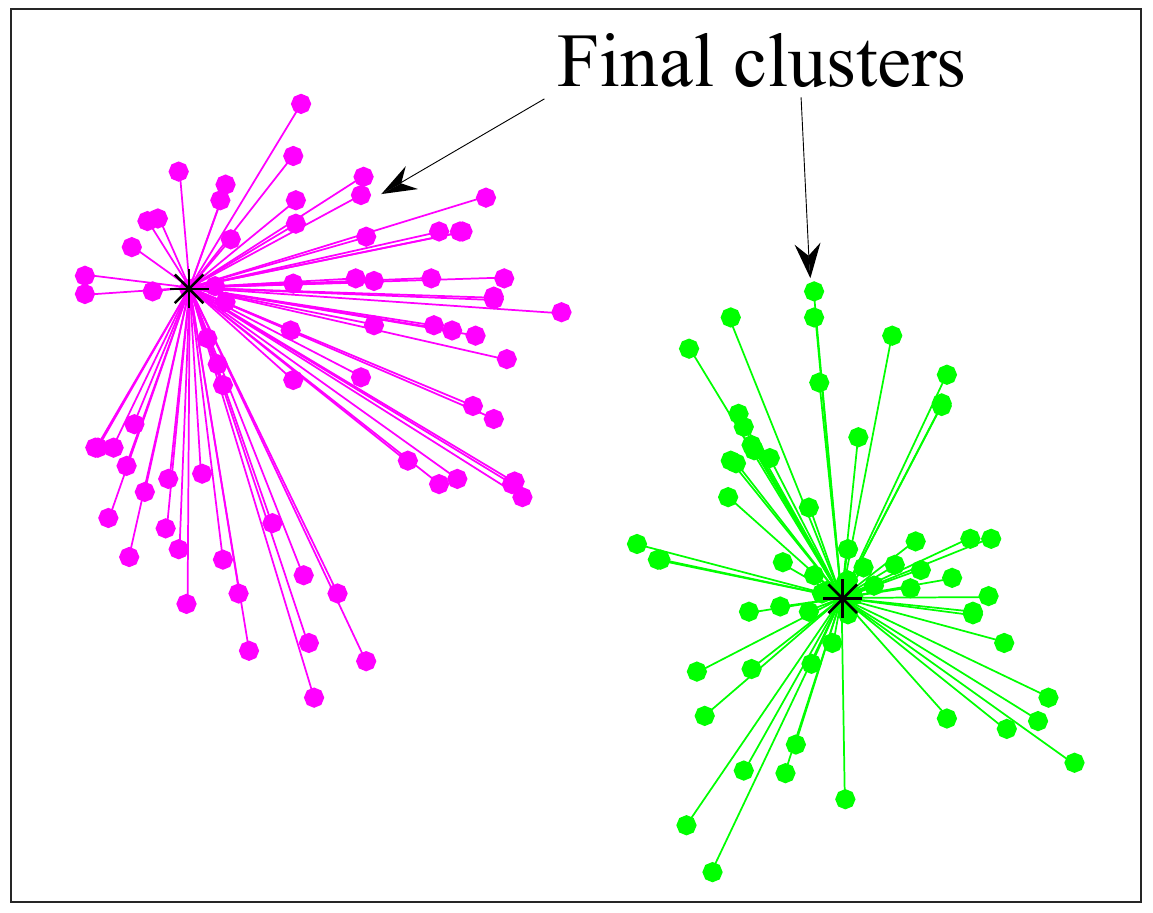}}
			\centerline{(c)}\medskip
		\end{minipage}
		\vspace{-1.5ex}
		\caption{An illustration of the improved affinity propagation on the ``Cubetoy'' image pair (only one image is shown due to the space limit. (a) The data with two structures (respectively marked in the green and purple colors) and gross outliers (marked in the yellow colors). (b) Some significant clusters (marked in different colors) are obtained by using the message propagation on the sparse graph after pruning the gross outliers. (c) The final two clusters (marked in magenta and green colors, respectively) are obtained from the significant clusters, and they indicate two estimated model instances.}
		\label{improveAP}
		\vspace{-3ex}
	\end{figure}

	After that, the details of IAP are described as follows (see Fig. \ref{improveAP}): Firstly, the $\tau$-nearest neighbor sparse graph is constructed based on the similarity measure in $\tilde{\mathcal{V}}^{d}$. Secondly, the Sparse AP (SAP) algorithm \cite{jia2008finding} is used to select the significant clusters on the sparse graph (see Fig. \ref{improveAP} (b)). This is because the sparse graph with fewer edges is faster than a normal graph during the message propagation process.
	Finally, SAP is used to refine the final clusters. In practice, each cluster corresponds to a true model instance of multi-structural data, and each pruned data point derives from one of the clusters (see Fig. \ref{improveAP} (c)).	

	We summarize the complete steps of the proposed HRMP in Alg. \ref{algorithm 1}. The computational complexity of HRMP is primarily governed by Steps 5 and 6 of the algorithm. For Step 5, the complexity of calculating the similarity between each pair of vertices is $O$($N'^2$). For Step 6, the complexity of the sparse AP algorithm is approximately $O$($N'^2$) + $O$($N''^2$) (where $N''$ is the number of the significant clusters). The other steps of HRMP take much less time than Steps 5 and 6. Therefore, the total complexity of HRMP approximately amounts to  2$O$($N'^2$) + $O$($N''^2$) ($N'' \ll N'$).
	
	\begin{algorithm}[!b]
		\caption{ The hierarchical representation via message propagation (HRMP) method}
		\label{algorithm 1}
		\begin{algorithmic}[1]
			\REQUIRE $\mathbf{X}=\{x_i\}_{i=1}^N$, the number of model hypotheses $M$, the $K$ value for IKOSE and the $\tau$ value for the sparse graph.
			\ENSURE The number of model instances and model parameters.
			\STATE Generate $M$ model hypotheses $\pmb{\theta}$ and estimate their inlier scales by IKOSE.
			\STATE Construct a hierarchical representation connected to the generated $\pmb{\theta}$ and $\mathbf{X}$ (described in Sec. \ref{twolayernetworks}).
			\STATE Propagate the information on the two layers of the hierarchical representation to weigh the quality of each edge. (described in Sec. \ref{HierarchicalRepresentationforModelFitting}).
			\STATE Calculate the weighting scores to remove the insignificant model hypotheses and gross outliers (described in Sec. \ref{HierarchicalRepresentationforModelFitting}).
			\STATE Construct the $\tau$-nearest neighbor sparse graph based on the similarities between the pruned vertices.
			\STATE Propagate the messages among the vertices of the $\tau$-nearest neighbor sparse graph to cluster the pruned data points (described in Sec. \ref{AffinityPropagationClustering}).
			\label{algorithmpassmessage2}
			\STATE  Estimate the number of model instances and model parameters based on the clusters.
		\end{algorithmic}
	\end{algorithm}	
	
	It is worth noting that T-Linkage \cite{magri2014t} and the proposed HRMP are significantly different. The main differences are: (1) $\mathbf{P}(\tilde{v}_i^d)$ in the proposed HRMP not only gathers the preference information between the vertices in both $\tilde{\mathcal{V}}^{h}$ and $\tilde{\mathcal{V}}^{d}$, but also considers the relationships between the vertices in $\tilde{\mathcal{V}}^{d}$ during the message propagation process. In contrast, T-Linkage only considers the preference information of the data points towards the model hypotheses. (2) HRMP measures the similarities of the vertices of $\tilde{\mathcal{V}}^{d}$, while T-Linkage calculates the similarities using all the data points. Therefore, HRMP is more effective and efficient than T-Linkage, because the influence of both insignificant model hypotheses and outliers are alleviated.	
	
	\section{Experiments}
	\label{sec:experiments}
	In this section, we firstly evaluate the influence of the different components of HRMP on the performance. Then, we compare the proposed HRMP with eleven state-of-the-art model fitting methods, including J-Linkage \cite{toldo2008robust}, KF \cite{chin2009robust}, AKSWH \cite{wang2012simultaneously}, T-Linkage \cite{magri2014t}, RPA \cite{magri2015robust}, DPA \cite{tiwari2016robust}, RansaCov \cite{magri2016multiple}, TSMP \cite{Wang2016}, HVF \cite{xiao2017hierarchical}, DGSAC \cite{tiwari2018dgsac} and MSHF \cite{wang2019searching} on both challenging synthetic data and real images. For fair comparison, the model hypotheses are generated by employing the proximity sampling \cite{toldo2008robust} for all the fitting methods. Similar to AKSWH, for synthetic data, we generate 5,000 and 10,000 model hypotheses for line and circle fitting, respectively. For real images, we generate 10,000 and 20,000 model hypotheses for two-view motion and multi-homography segmentation on the AdelaideRMF dataset \cite{wong2011dynamic}, respectively. All experiments are repeated 50 times, and then the mean fitting error \cite{magri2014t}, the standard deviation of fitting errors, the total average and total median fitting errors, the mean CPU time and the total average CPU time are used to evaluate the performance of these methods.
	\begin{figure}[!t]
		\vspace{-4ex}
		\begin{center}
			\begin{minipage}[b]{0.3\textwidth}
				\includegraphics[width=1.\textwidth]{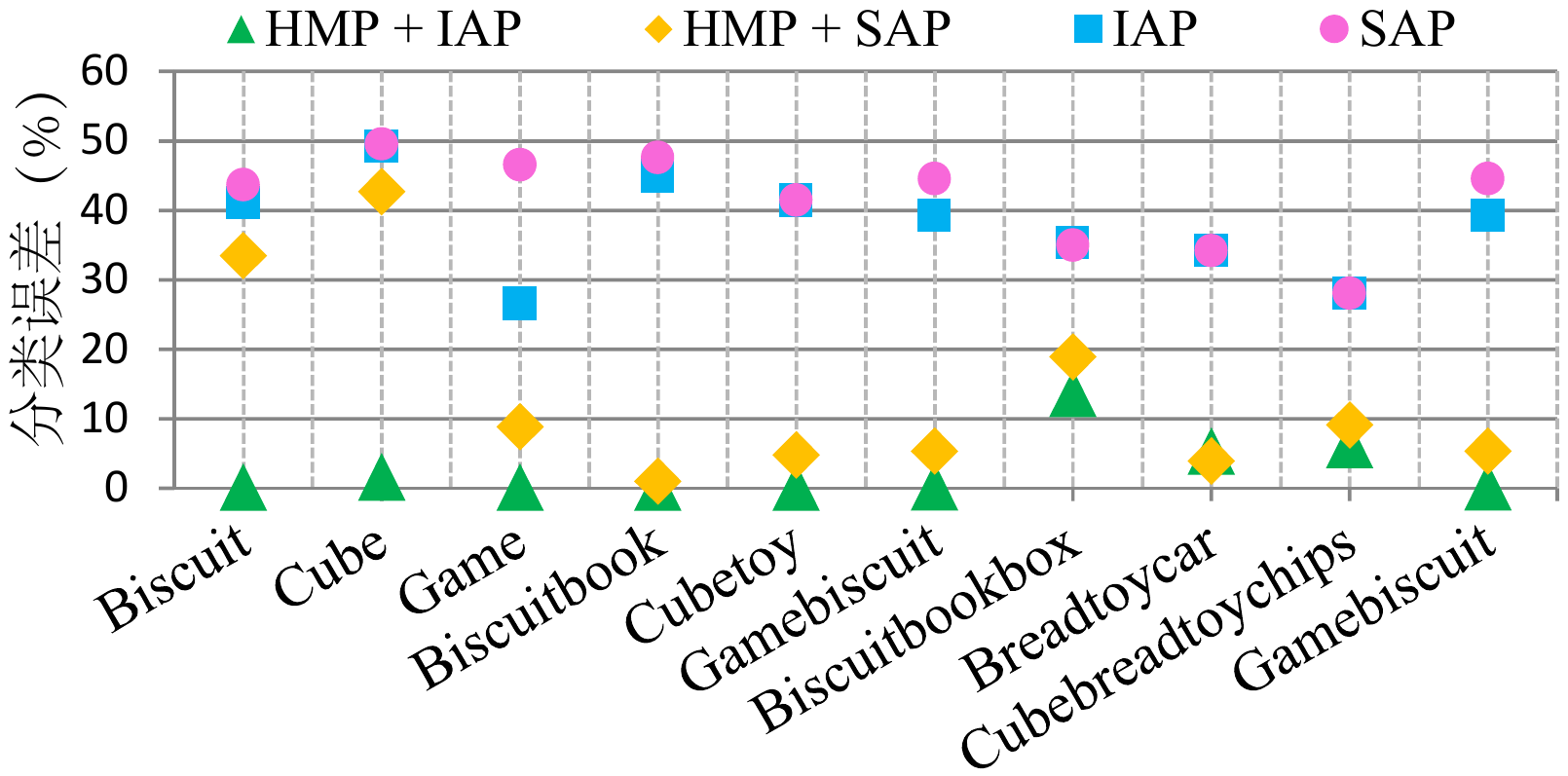}
			\end{minipage}
			\vspace{-2ex}
			\caption{The influence of different components on the performance of HRMP.}
			\label{components1}
		\end{center}
		\vspace{-3ex}
	\end{figure}

	\vspace{-2ex}
	\subsection{Influence of Components}	
	\label{AblationStudy}
	In this subsection, we evaluate the influence of HRMP with different components, including HRMP using both the hierarchical message propagation algorithm and the improved affinity propagation algorithm (i.e., HMP+IAP), with only the hierarchical message propagation algorithm (i.e., HMP+SAP), with only the improved affinity propagation algorithm (i.e., IAP), and with the sparse AP algorithm (i.e., SAP, which is used as a benchmark). 
	Fig. \ref{components1} shows the influence of the different components on the performance of HRMP.
	As shown in Fig. \ref{components1}, SAP obtains the worst performance. IAP improves the performance of SAP, which shows the effectiveness of IAP. HMP+SAP achieves the second best performance because HMP effectively alleviates the influence of gross outliers.	The proposed HRMP (i.e., HMP+IAP) achieves the best performance due to the contributions of both the proposed hierarchical message propagation algorithm and the improved affinity propagation algorithm.
	
\begin{figure}[!t]
		\vspace{-5ex}
		\begin{center}
			\begin{minipage}[b]{0.35\textwidth}
				\includegraphics[width=1.\textwidth]{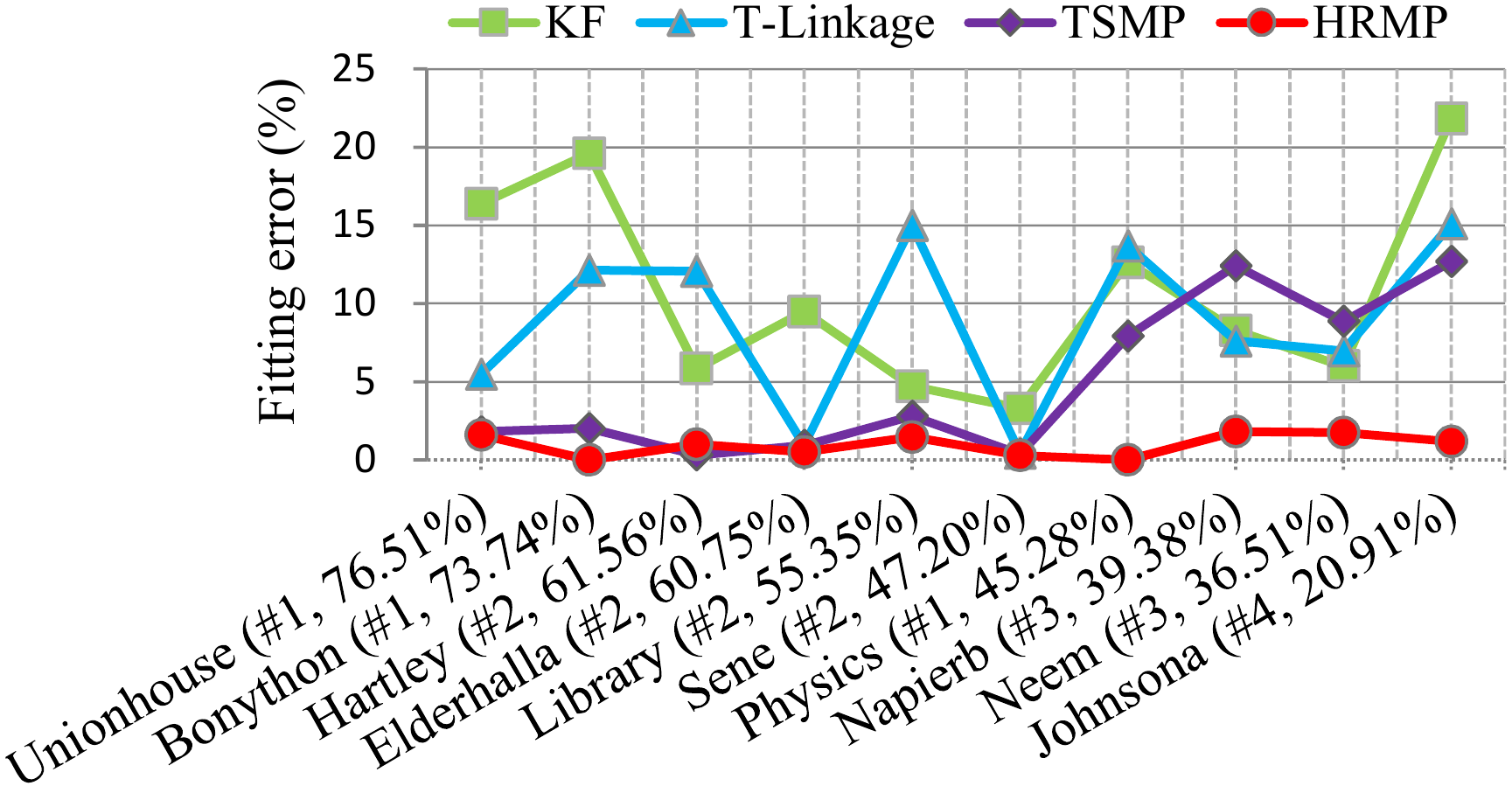}	
			\end{minipage}
			\vspace{-2ex}
			\caption{The influence of different gross outlier percentages (\%) on KF, T-Linkage, TSMP and HRMP. \# indicates the true number of model instances. The percentages in brackets means the gross outlier percentages that the corresponding data include.}
			\label{components2}
		\end{center}
		\vspace{-3ex}
	\end{figure}

	In addition, we further evaluate the influence of different gross outlier percentages on the HMP component of the proposed HRMP.	Fig. \ref{components2} shows the gross outlier removal results obtained by HRMP and the other competing methods with gross outlier removal components (i.e., KF, T-Linkage and TSMP). From Fig. \ref{components2}, we can see that KF and T-Linkage obtain the bad results, although they can remove gross outliers. This is because that gross outliers are usually accompanied by multiple model instances (i.e., the more number of real structures the input data include, the less gross outlier percentages the data have), resulting in some gross outliers being incorrectly identified as the inliers of real structures. TSMP obtains the second best results, but it cannot yet effectively handle the gross outliers and  multi-structural data (e.g., when the true number of model instances is 3 or 4). In contrast, the proposed HRMP achieves the best results. This is due to the fact that gross outliers can be effectively removed during the message propagation process, so that the influence of gross outliers on the performance of HRMP is alleviated.
	
	\vspace{-2ex}
	\subsection{Line Fitting}
	\label{lineFitting}
\begin{figure}[!t]
		\vspace{-3.7ex}
		\begin{center}
			\subfigure[4 Lines]{
				\begin{minipage}[b]{0.13\textwidth}
					\includegraphics[width=1.\textwidth]{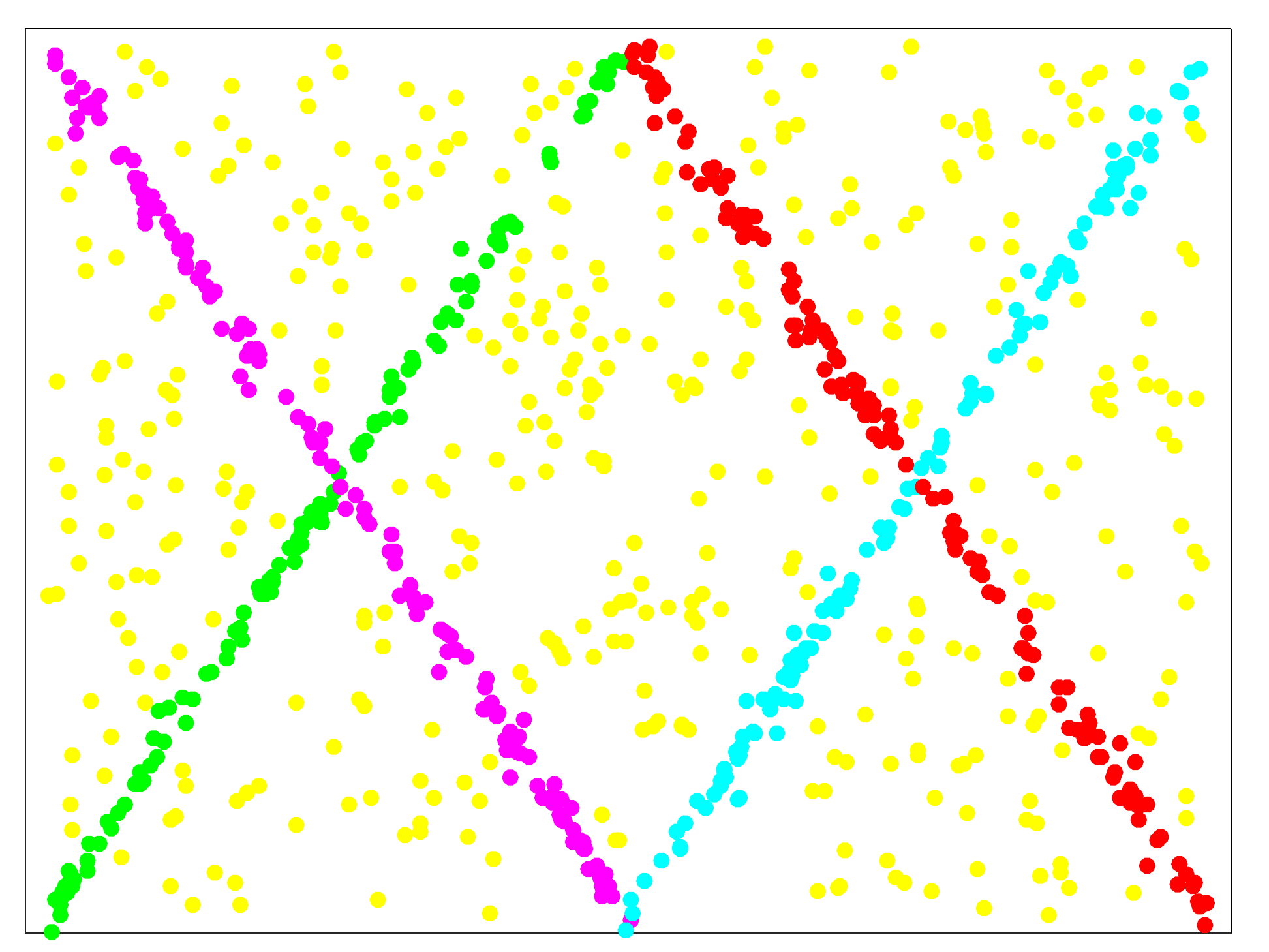}\vspace{0.5ex}
					\includegraphics[width=1.\textwidth]{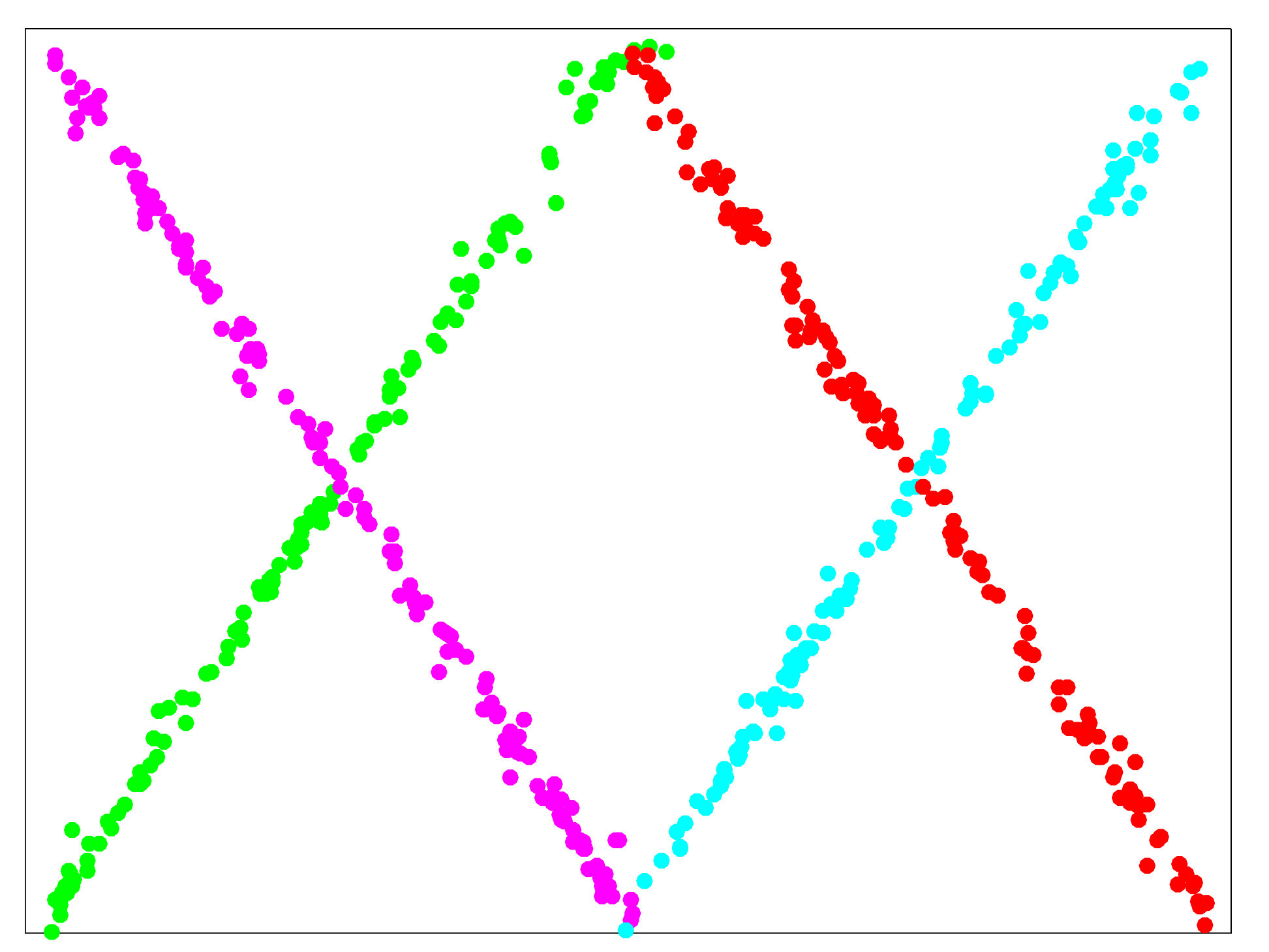}
			\end{minipage}}\hspace{0.5ex}
			\subfigure[5 Lines]{
				\begin{minipage}[b]{0.13\textwidth}
					\includegraphics[width=1.\textwidth]{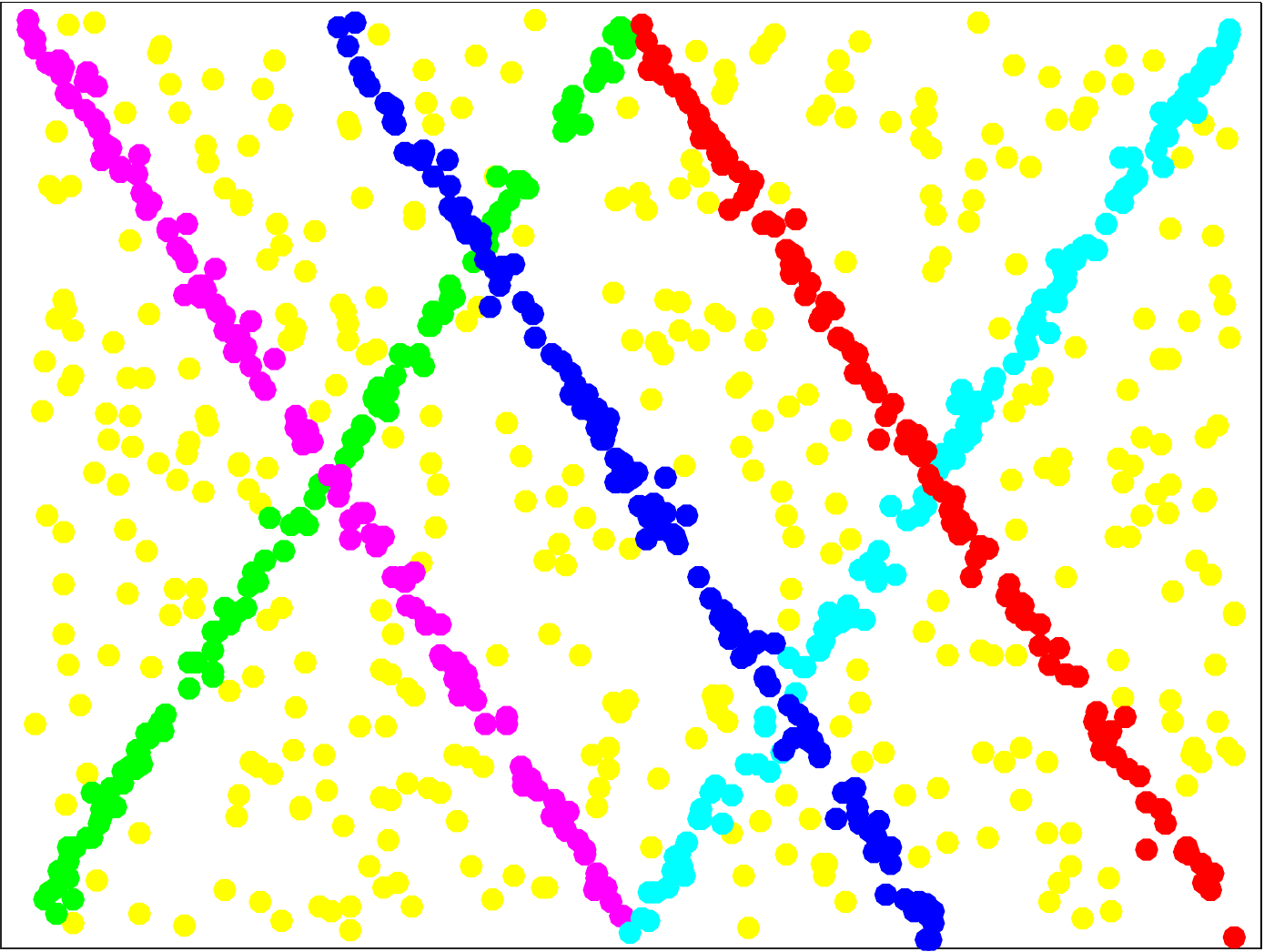}\vspace{0.5ex}
					\includegraphics[width=1.\textwidth]{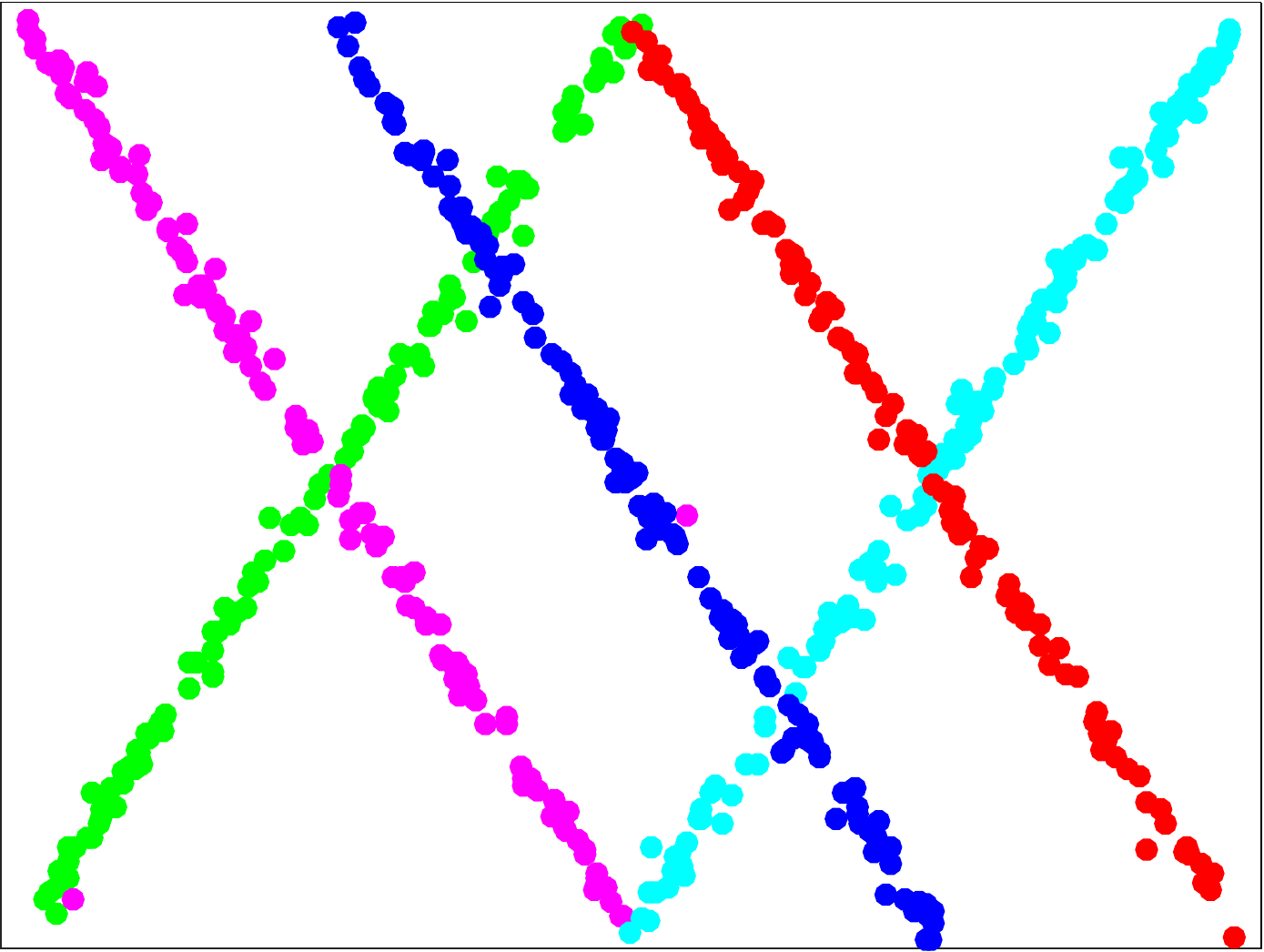}
			\end{minipage}}\hspace{0.5ex}
			\subfigure[6 Lines]{
				\begin{minipage}[b]{0.13\textwidth}
					\includegraphics[width=1.\textwidth]{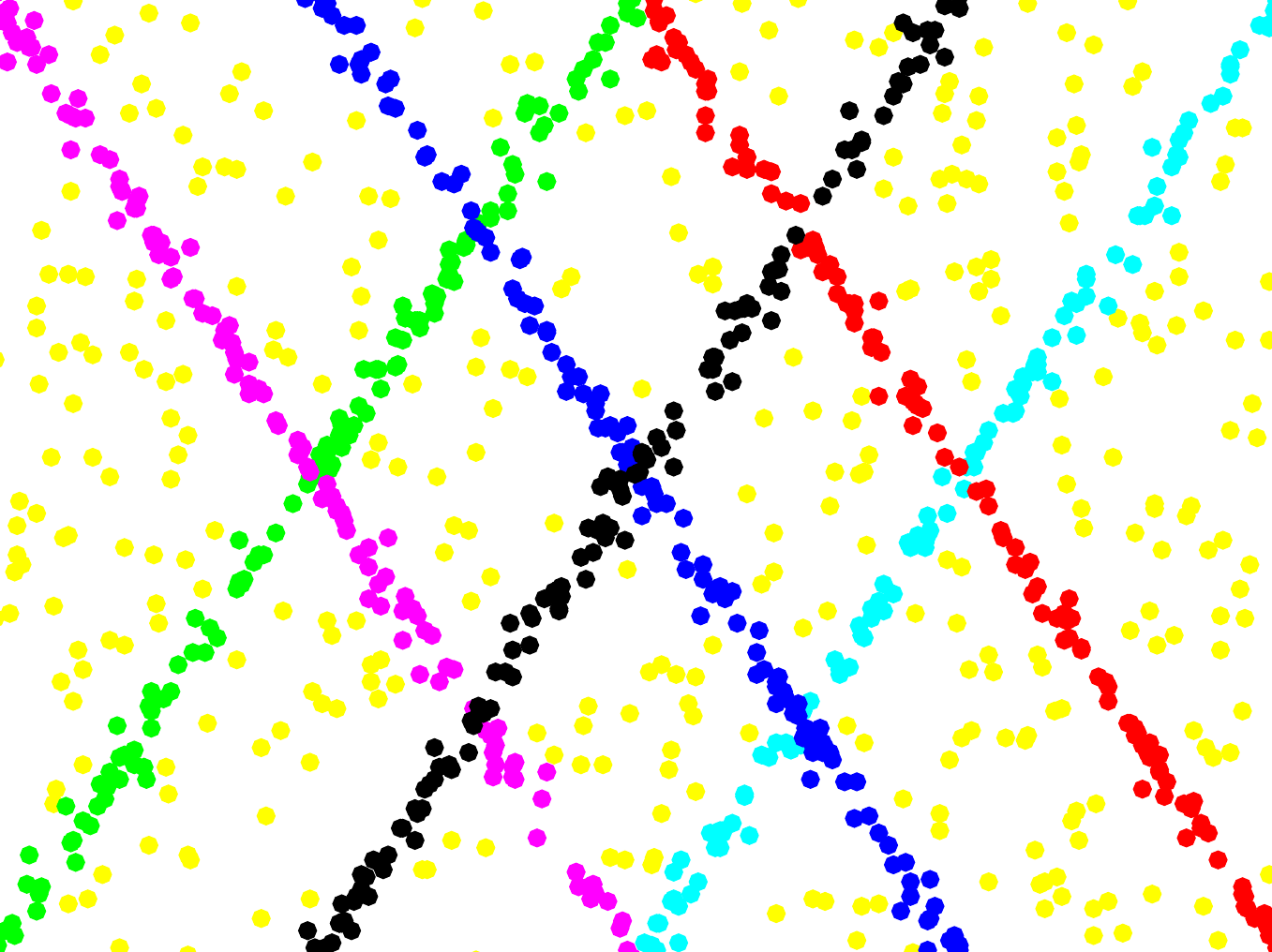}\vspace{0.5ex}
					\includegraphics[width=1.\textwidth]{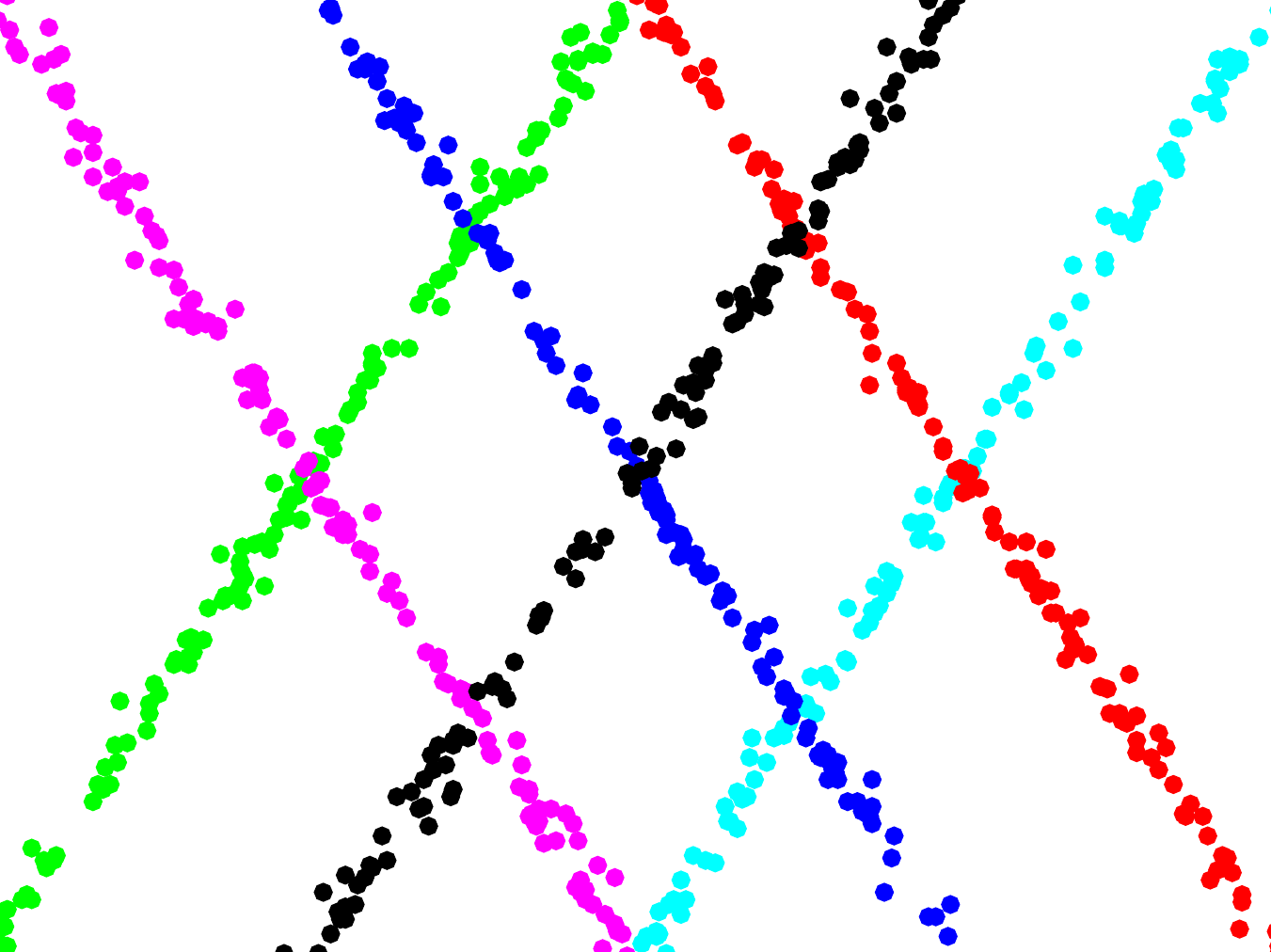}
			\end{minipage}}	
			\vspace{-1.5ex}
			\caption{Examples for line fitting. The $1^{st}$ and $2^{nd}$ rows are the Ground-truth fitting results and the results obtained by HRMP, respectively. The gross outliers are marked in the yellow color. The inliers belonging to different model instances are marked in the other different colors.}	
			\label{line}
		\end{center}
		\vspace{-3ex}
	\end{figure}

	In this subsection, we evaluate the performance of all the competing methods for line fitting on the synthetic data.	From Fig. \ref{line} and Table \ref{tableLine}, we can see that HRMP obtains the best results among all the competing methods. For the 3-line data and the 4-line data, all the competing methods correctly estimate the parameters of lines. For the 5-line data and the 6-line data, AKSWH, TSMP, RansaCov, DGSAC and MSHF correctly estimate the number of lines within a reasonable time, while TSMP obtains the second lowest total average and total median fitting errors among all the competing methods due to its robustness to outliers.T-Linkage, RPA and HVF obtain similar results. In particular, T-Linkage alleviates the influence of the intersections of different model instances by seeking the maximum cardinality intervals, but it cannot handle data with a large number of pseudo-outliers. RPA takes much time in the preference analysis and the matrix decomposition. HVF relies on a cut-off threshold to distinguish inliers from outliers. Moreover, both J-Linkage and DPA obtain the high standard deviation of fitting errors, the high total average and total median fitting errors among all the competing methods. J-Linkage and KF fail to fit the 6-line data. This is because that J-Linkage and KF cannot properly handle the data points around the intersections of different model instances. DPA is sensitive to the selection of the inlier-outlier boundary, especially for data with high outlier rates.

	\begin{table*}[!t]
		\vspace{-6ex}
		\caption{Quantitative comparison results of line fitting on four data. \# indicates the true number of model instances included in data. Mean and Time represent the mean fitting error and the mean CPU time, respectively. The best results are boldfaced.}
		\label{tableLine}
		\vspace{-5.5ex}
		\begin{center}
			\scalebox{0.8}{\tabcolsep0.03in
				\begin{tabular}{cccrrrrrrrrrrrrrrrrrrrrrrrr}
					\toprule
					\multicolumn{1}{c}{\multirow{2}[4]{*}{Data}} & \multicolumn{1}{c}{\multirow{2}[4]{*}{\shortstack{Outlier\\rate \\(\%)}}} & \multicolumn{1}{c}{\multirow{2}[4]{*}{\shortstack{Gross\\outlier\\rate (\%)}}} & \multicolumn{2}{c}{J-Linkage} & \multicolumn{2}{c}{KF} & \multicolumn{2}{c}{AKSWH} & \multicolumn{2}{c}{T-Linkage} & \multicolumn{2}{c}{{RPA}} & \multicolumn{2}{c}{{DPA}} & \multicolumn{2}{c}{RansaCov} & \multicolumn{2}{c}{TSMP} & \multicolumn{2}{c}{{HVF}} & \multicolumn{2}{c}{{DGSAC}} & \multicolumn{2}{c}{MSHF} & \multicolumn{2}{c}{HRMP} \\
					\cmidrule{4-27}          &       &       & \multicolumn{1}{c}{Mean} & \multicolumn{1}{c}{Time} & \multicolumn{1}{c}{Mean} & \multicolumn{1}{c}{Time} & \multicolumn{1}{c}{Mean} & \multicolumn{1}{c}{Time} & \multicolumn{1}{c}{Mean} & \multicolumn{1}{c}{Time} & \multicolumn{1}{c}{{Mean}} & \multicolumn{1}{c}{{Time}} & \multicolumn{1}{c}{{Mean}} & \multicolumn{1}{c}{{Time}} & \multicolumn{1}{c}{Mean} & \multicolumn{1}{c}{Time} & \multicolumn{1}{c}{Mean} & \multicolumn{1}{c}{Time} & \multicolumn{1}{c}{{Mean}} & \multicolumn{1}{c}{{Time}} & \multicolumn{1}{c}{{Mean}} & \multicolumn{1}{c}{{Time}} & \multicolumn{1}{c}{Mean} & \multicolumn{1}{c}{Time} & \multicolumn{1}{c}{Mean} & \multicolumn{1}{c}{Time} \\
					\midrule
					\multicolumn{1}{l}{3 Lines (\#3)} & 84.25  & 52.63  & 9.25  & 3.06  & 12.11  & 4.92  & 2.02  & 0.44  & 8.43  & 70.65  & {6.69} & {93.03} & {13.50} & {54.34} & 2.71  & 3.58  & 2.14  & 1.04  & {6.25} & {10.21} & {6.13} & {69.01} & 2.91  & 0.90  & \textbf{1.64} & \textbf{0.32} \\
					\multicolumn{1}{l}{4 Lines (\#4)} & 86.00  & 42.22  & 16.57  & 3.81  & 15.17  & 6.92  & 4.67  & 0.54  & 7.88  & 93.98  & {9.36} & {116.59} & {14.44} & {66.85} & 3.92  & 3.79  & 2.81  & 1.66  & {7.86} & {8.73} & {6.87} & {76.92} & 4.62  & 1.23  & \textbf{1.83} & \textbf{0.39} \\
					\multicolumn{1}{l}{5 Lines (\#5)} & 87.80  & 36.68  & 29.19  & 4.56  & 17.18  & 10.05  & 8.03  & 0.49  & 10.20  & 117.26  & {11.26} & {185.63} & {33.60} & {80.08} & 5.06  & 4.25  & 3.52  & 2.38  & {13.20} & {13.14} & {5.98} & {93.99} & 6.33  & 1.32  & \textbf{1.92} & \textbf{0.48} \\
					\multicolumn{1}{l}{6 Lines (\#6)} & 90.12  & 31.72  & 31.81  & 5.41  & 23.05  & 14.39  & 8.47  & 0.77  & 11.15  & 139.47  & {10.99} & {293.14} & {40.18} & {95.11} & 6.17  & 4.60  & 4.17  & 3.29  & {13.71} & {17.77} & {8.38} & {107.84} & 6.79  & 1.43  & \textbf{2.60} & \textbf{0.58} \\
					\midrule
					Std.  &       &       & 10.64 & -     & 4.61  & -     & 3.04  & -     & 1.52  & -     & {2.10} & {-} & {13.51} & {-} & 1.49  & -     & 0.88  & -     & {3.76} & {-} & {1.10} & {-} & 1.77  & -     & \textbf{0.42} & - \\
					{Total median} & {} & {} & {22.88} & {-} & {16.17} & {-} & {6.35} & {-} & {9.32} & {-} & {10.17} & {-} & {24.02} & {-} & {4.49} & {-} & {3.17} & {-} & {10.53} & {-} & {6.50} & {-} & {5.48} & {-} & {\textbf{1.88}} & {-} \\
					Total average &       &       & 21.70 & 4.21  & 16.88 & 9.07  & 5.80  & 0.56  & 9.42  & 105.34 & {9.57} & {172.09} & {25.43} & {74.10} & 4.47  & 4.06  & 3.16  & 2.09  & {10.25} & {12.46} & {6.84} & {86.94} & 5.16  & 1.22  & \textbf{2.00} & \textbf{0.44} \\
					\bottomrule
				\end{tabular}
			}
		\end{center}
		\vspace{-1ex}
	\end{table*}

	\begin{table*}[!t]
		\vspace{-3ex}
		\caption{Quantitative comparison results of circle fitting on four data. \# indicates the true number of model instances included in data. Mean and Time represent the mean fitting error and the mean CPU time, respectively. The best results are boldfaced.}
		\label{tableCircle}
		\vspace{-5.5ex}
		\begin{center}
			\scalebox{0.8}{\tabcolsep0.0245in
				\begin{tabular}{cccrrrrrrrrrrrrrrrrrrrrrrrr}
					\toprule
					\multicolumn{1}{c}{\multirow{2}[4]{*}{Data}} & \multicolumn{1}{c}{\multirow{2}[4]{*}{\shortstack{Outlier\\rate \\(\%)}}} & \multicolumn{1}{c}{\multirow{2}[4]{*}{\shortstack{Gross\\outlier\\rate (\%)}}} & \multicolumn{2}{c}{J-Linkage} & \multicolumn{2}{c}{KF} & \multicolumn{2}{c}{AKSWH} & \multicolumn{2}{c}{T-Linkage} & \multicolumn{2}{c}{{RPA}} & \multicolumn{2}{c}{{DPA}} & \multicolumn{2}{c}{RansaCov} & \multicolumn{2}{c}{TSMP} & \multicolumn{2}{c}{{HVF}} & \multicolumn{2}{c}{{DGSAC}} & \multicolumn{2}{c}{MSHF} & \multicolumn{2}{c}{HRMP} \\
					\cmidrule{4-27}          &       &       & \multicolumn{1}{c}{Mean} & \multicolumn{1}{c}{Time} & \multicolumn{1}{c}{Mean} & \multicolumn{1}{c}{Time} & \multicolumn{1}{c}{Mean} & \multicolumn{1}{c}{Time} & \multicolumn{1}{c}{Mean} & \multicolumn{1}{c}{Time} & \multicolumn{1}{c}{{Mean}} & \multicolumn{1}{c}{{Time}} & \multicolumn{1}{c}{{Mean}} & \multicolumn{1}{c}{{Time}} & \multicolumn{1}{c}{Mean} & \multicolumn{1}{c}{Time} & \multicolumn{1}{c}{Mean} & \multicolumn{1}{c}{Time} & \multicolumn{1}{c}{{Mean}} & \multicolumn{1}{c}{{Time}} & \multicolumn{1}{c}{{Mean}} & \multicolumn{1}{c}{{Time}} & \multicolumn{1}{c}{Mean} & \multicolumn{1}{c}{Time} & \multicolumn{1}{c}{Mean} & \multicolumn{1}{c}{Time} \\
					\midrule
					\multicolumn{1}{l}{4 Circles (\#4)} & 84.70  & 36.60  & 10.26  & 32.24  & 13.68  & 54.54  & 1.05  & 1.28  & 6.64  & 291.92  & {13.72} & {267.23} & {9.90} & {153.00} & 1.29  & 21.33  & 3.58  & 3.99  & {13.75} & {31.88} & {1.35} & {131.06} & 1.87  & 1.86  & \textbf{0.78} & \textbf{0.94} \\
					\multicolumn{1}{l}{5 Circles (\#5)} & 86.87  & 32.60  & 22.55  & 37.32  & 19.97  & 58.70  & 2.07  & 1.41  & 7.40  & 345.12  & {20.61} & {522.36} & {19.65} & {190.33} & 4.25  & 20.74  & 8.23  & 6.41  & {14.22} & {49.43} & {2.00} & {171.18} & 1.91  & 1.96  & \textbf{1.69} & \textbf{1.08} \\
					\multicolumn{1}{l}{6 Circles (\#6)} & 88.54  & 28.61  & 27.78  & 41.84  & 22.60  & 81.37  & 3.76  & 1.51  & 8.45  & 393.81  & {31.85} & {746.40} & {26.69} & {232.56} & 3.76  & 20.14  & 6.89  & 8.04  & {14.85} & {65.81} & {1.88} & {221.99} & 1.95  & 2.10  & \textbf{1.78} & \textbf{1.10} \\
					\multicolumn{1}{l}{7 Circles (\#7)} & 89.59  & 24.34  & 34.56  & 66.50  & 27.67  & 108.29  & 2.95  & 3.19  & 10.71  & 636.55  & {34.32} & {1273.34} & {29.24} & {279.76} & 4.99  & 30.26  & 5.27  & 11.64  & {15.82} & {87.54} & {3.77} & {256.50} & 2.10  & 2.63  & \textbf{1.92} & \textbf{1.46} \\
					\midrule
					Std.  &       &       & 10.27 & -     & 5.82  & -     & 1.17  & -     & 1.77  & -     & {9.66} & {-} & {8.66} & {-} & 1.60  & -     & 2.01  & -     & {0.90} & {-} & {1.05} & {-} & \textbf{0.10} & -     & 0.52  & - \\
					{Total median} & {} & {} & {25.16} & {-} & {21.28} & {-} & {2.51} & {-} & {7.93} & {-} & {26.23} & {-} & {23.17} & {-} & {4.01} & {-} & {6.08} & {-} & {14.54} & {-} & {1.94} & {-} & {1.93} & {-} & {\textbf{1.74}} & {-} \\
					Total average &       &       & 23.79 & 44.48 & 20.98 & 75.72 & 2.46  & 1.85  & 8.30  & 416.85 & {25.12} & {702.33} & {21.37} & {213.91} & 3.57  & 23.12 & 5.99  & 7.52  & {14.66} & {58.67} & {2.25} & {195.18} & 1.96  & 2.14  & \textbf{1.54} & \textbf{1.15} \\
					\bottomrule
				\end{tabular}
			}
		\end{center}
		\vspace{-4ex}
	\end{table*}

	\vspace{-2ex}
	\subsection{Circle Fitting}
	\label{circleFitting}
\begin{figure}[!t]
		\vspace{-3.5ex}
		\begin{center}
			\subfigure[4 Circles]{
				\begin{minipage}[b]{0.13\textwidth}
					\includegraphics[width=0.95\textwidth]{figures/4circle_temp_G}\vspace{0.5ex}
					\includegraphics[width=0.95\textwidth]{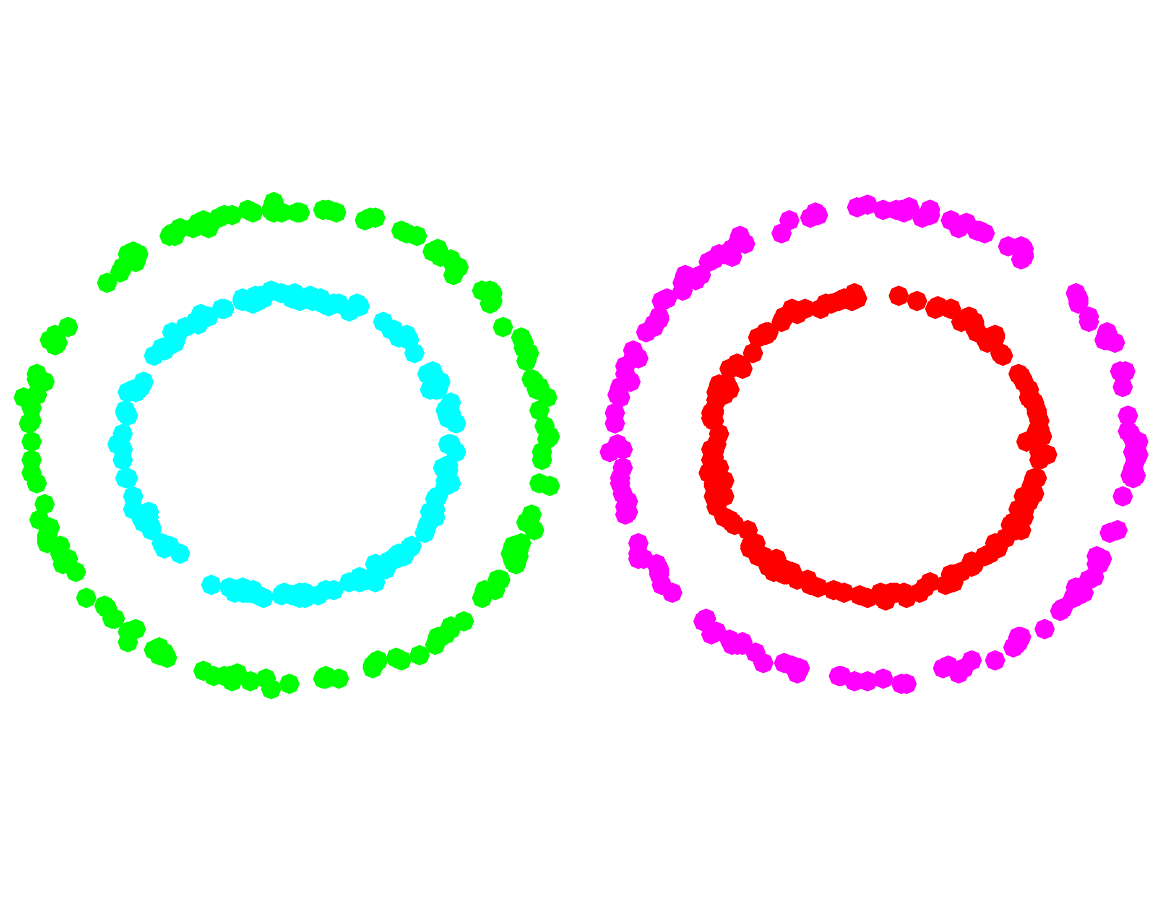}
			\end{minipage}}\hspace{0.9ex}
			\subfigure[5 Circles]{
				\begin{minipage}[b]{0.13\textwidth}
					\includegraphics[width=0.95\textwidth]{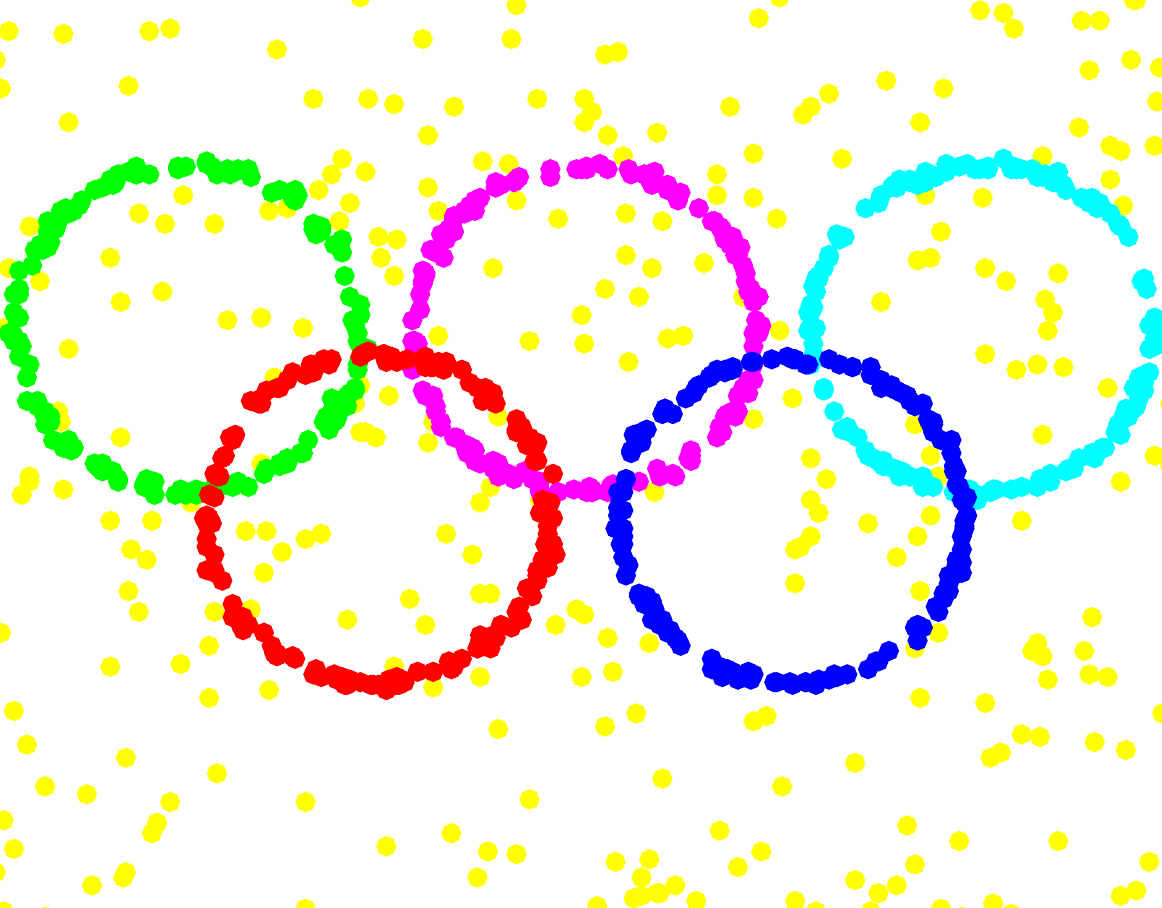}\vspace{0.5ex}
					\includegraphics[width=0.95\textwidth]{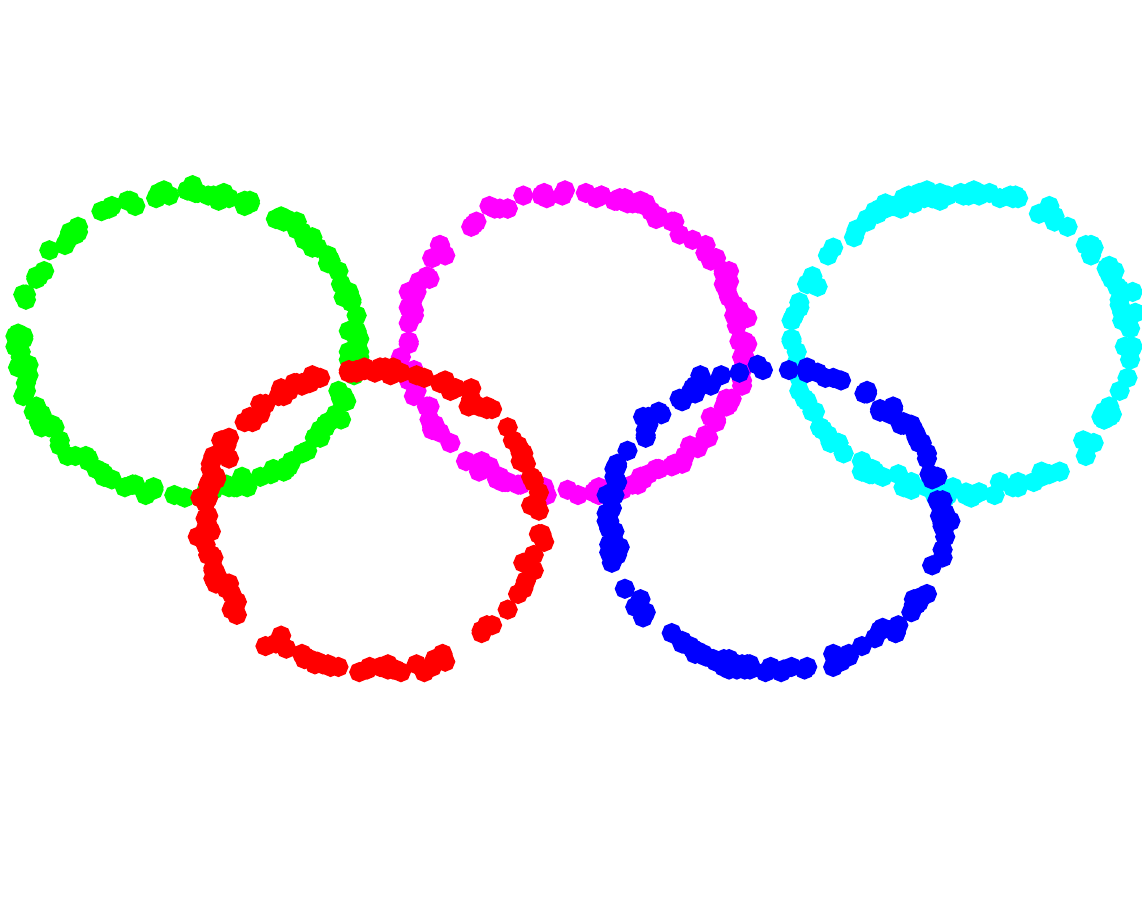}
			\end{minipage}}\hspace{0.7ex}
			\subfigure[6 Circles]{
				\begin{minipage}[b]{0.13\textwidth}
					\includegraphics[width=0.95\textwidth]{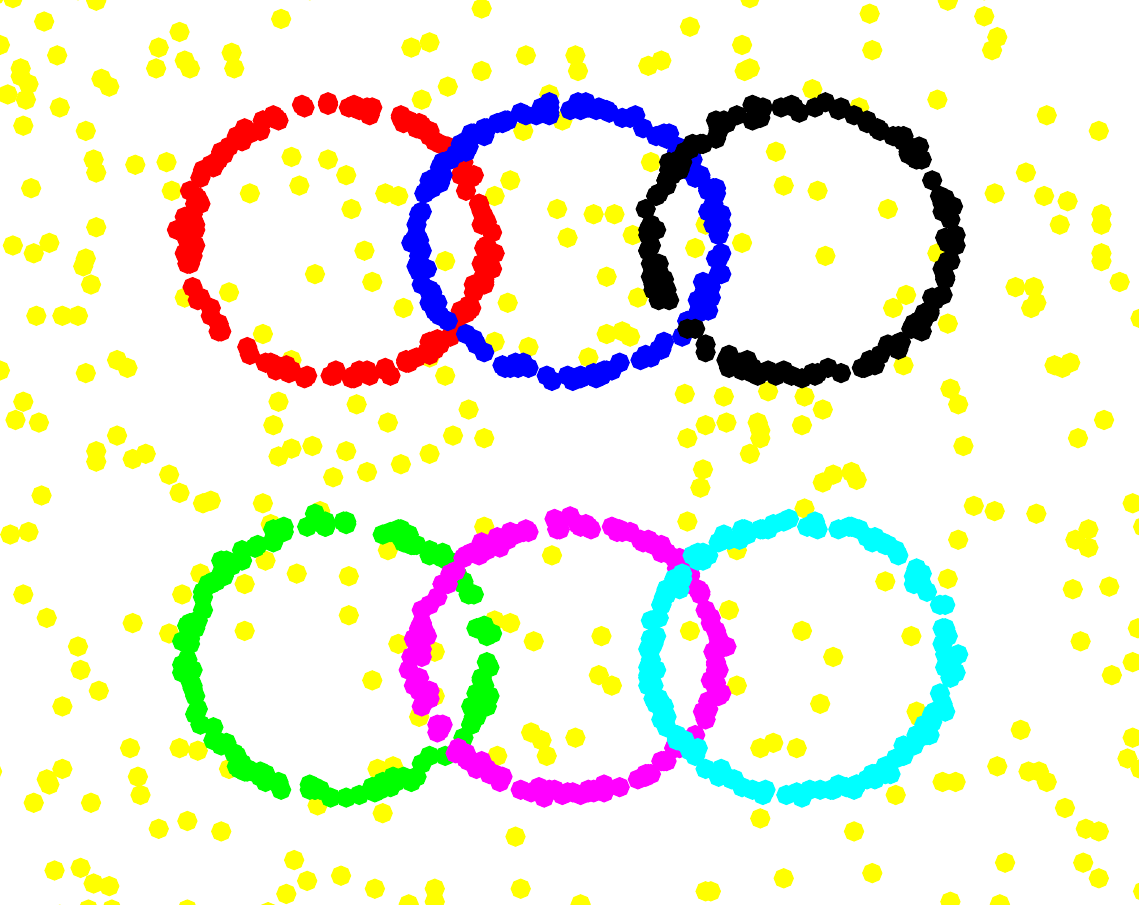}\vspace{0.5ex}
					\includegraphics[width=0.95\textwidth]{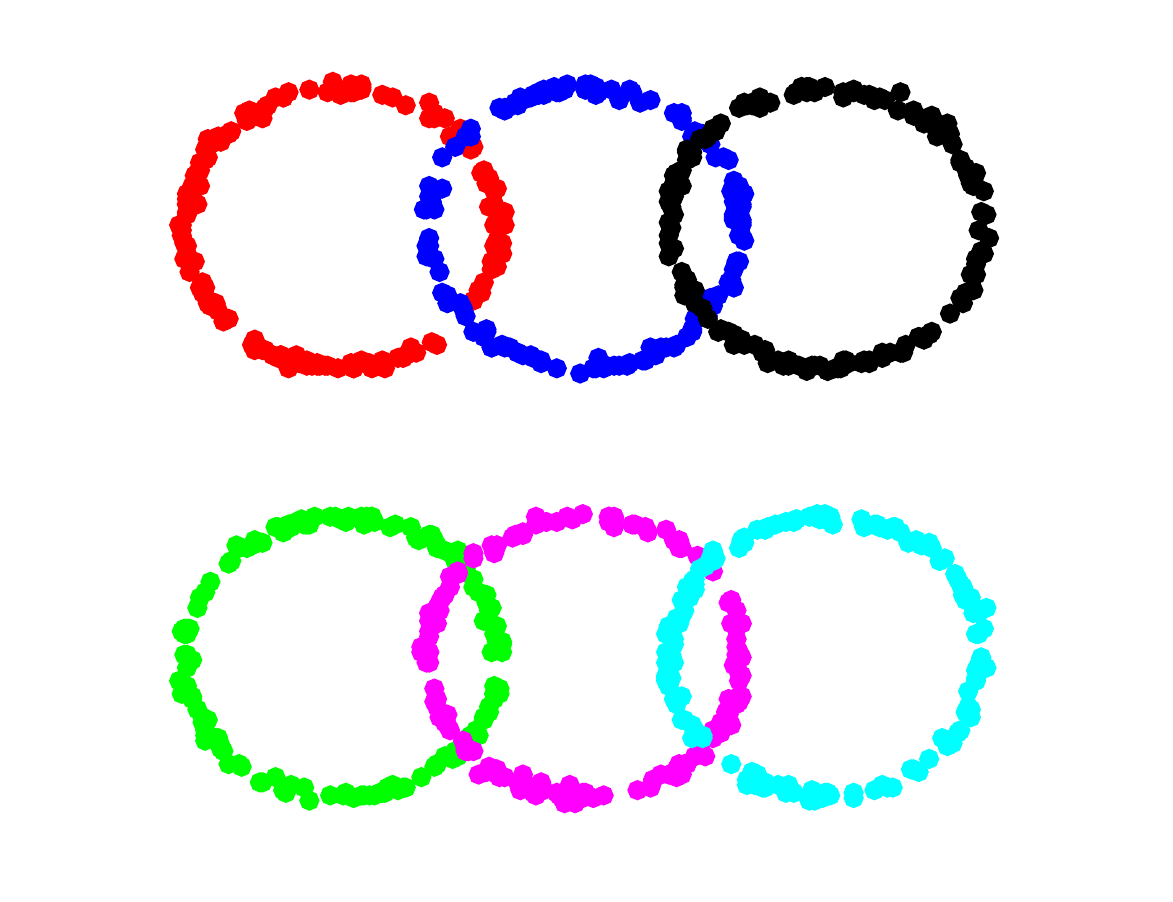}
			\end{minipage}}
			\vspace{-1.5ex}
			\caption{Examples for circle fitting. The $1^{st}$ and $2^{nd}$ rows are the Ground-truth fitting results and the results obtained by HRMP, respectively. The gross outliers are marked in the yellow color. The inliers belonging to different model instances are marked in the other different colors.}
			\label{circle}
		\end{center}
		\vspace{-4ex}
	\end{figure}

	In this subsection, we also test the performance of all the competing methods for circle fitting on the synthetic data. From Fig. \ref{circle} and Table \ref{tableCircle}, we can see that HRMP achieves the lowest total average and total median fitting errors and the fastest total average CPU time among all the competing methods. For the 4-circle data, AKSWH, RansaCov, TSMP, DGSAC and MSHF succeed in estimating the number of the circles. For the 5-circle data, the 6-circle data and the 7-circle data, with the increase of the number of the circles and the outlier rate, the performance of J-Linkage, KF, RPA and DPA gradually deteriorates because some outliers are incorrectly identified as inliers, especially when the data are contaminated with a large number of outliers. J-Linkage and KF cannot effectively segment all the circles due to their sensitivity to pseudo-outliers. Similarly, the computational time of T-Linkage and RPA is very high (T-Linkage and RPA are respectively about 360 and 610 times slower than the proposed HRMP) for the circle data because the agglomerative clustering algorithm used by T-Linkage needs to calculate the similarities of all points and RPA needs to perform principal component analysis and symmetric matrix decomposition. On the contrary, AKSWH, RansaCov, TSMP, HVF and MSHF correctly fit the number of circles within a reasonable time. Moreover, MSHF also obtains the best standard deviation of fitting errors, the second lowest total average and total median fitting errors among all the competing methods (see the bottom of Table \ref{tableCircle}), since it can effectively find the representative modes on hypergraphs.

	\vspace{-2ex}
	\subsection{Multi-Homography Segmentation}
	\label{homographyBasedSegmentation}
	In this subsection, we evaluate the performance of all the competing methods for multi-homography segmentation by using all the 19 real image pairs with single-structure and multi-structural data from the AdelaideRMF dataset (see Table \ref{tableHomography} for the quantitative comparison results). 
	
	\begin{table*}[!ht]
		\vspace{-6.5ex}
		\caption{Quantitative comparison results of multi-homography segmentation on nineteen image pairs. \# indicates the true number of model instances included in data. Mean and Time represent the mean fitting error and the mean CPU time, respectively. The best results are boldfaced.}
		\label{tableHomography}
		\vspace{-5.5ex}
		\begin{center}
			\scalebox{0.8}{\tabcolsep0.017in
				\begin{tabular}{cccrrrrrrrrrrrrrrrrrrrrrrrr}
					\toprule
					\multicolumn{1}{c}{\multirow{2}[4]{*}{Data}} & \multicolumn{1}{c}{\multirow{2}[4]{*}{\shortstack{Outlier\\rate \\(\%)}}} & \multicolumn{1}{c}{\multirow{2}[4]{*}{\shortstack{Gross\\outlier\\rate (\%)}}} & \multicolumn{2}{c}{J-Linkage} & \multicolumn{2}{c}{KF} & \multicolumn{2}{c}{AKSWH} & \multicolumn{2}{c}{T-Linkage} & \multicolumn{2}{c}{{RPA}} & \multicolumn{2}{c}{{DPA}} & \multicolumn{2}{c}{RansaCov} & \multicolumn{2}{c}{TSMP} & \multicolumn{2}{c}{{HVF}} & \multicolumn{2}{c}{{DGSAC}} & \multicolumn{2}{c}{MSHF} & \multicolumn{2}{c}{HRMP} \\
					\cmidrule{4-27}          &       &       & \multicolumn{1}{c}{Mean} & \multicolumn{1}{c}{Time} & \multicolumn{1}{c}{Mean} & \multicolumn{1}{c}{Time} & \multicolumn{1}{c}{Mean} & \multicolumn{1}{c}{Time} & \multicolumn{1}{c}{Mean} & \multicolumn{1}{c}{Time} & \multicolumn{1}{c}{{Mean}} & \multicolumn{1}{c}{{Time}} & \multicolumn{1}{c}{{Mean}} & \multicolumn{1}{c}{{Time}} & \multicolumn{1}{c}{Mean} & \multicolumn{1}{c}{Time} & \multicolumn{1}{c}{Mean} & \multicolumn{1}{c}{Time} & \multicolumn{1}{c}{{Mean}} & \multicolumn{1}{c}{{Time}} & \multicolumn{1}{c}{{Mean}} & \multicolumn{1}{c}{{Time}} & \multicolumn{1}{c}{Mean} & \multicolumn{1}{c}{Time} & \multicolumn{1}{c}{Mean} & \multicolumn{1}{c}{Time} \\
					\midrule
					\multicolumn{1}{l}{{Barrsmith (\#2)}} & {90.46} & {68.88} & {17.87} & {2.70} & {29.66} & {3.14} & {9.91} & {1.77} & {\textbf{4.38}} & {92.56} & {20.60} & {51.07} & {16.17} & {48.32} & {14.68} & {55.64} & {17.19} & {1.55} & {24.51} & {2.55} & {29.88} & {28.94} & {10.09} & {0.98} & {12.64} & {\textbf{0.86}} \\
					\multicolumn{1}{l}{{Bonhall (\#6)}} & {94.29} & {6.18} & {47.37} & {75.70} & {51.51} & {60.55} & {23.80} & {\textbf{6.48}} & {43.11} & {2157.22} & {40.23} & {413.17} & {45.50} & {247.15} & {37.47} & {661.92} & {45.69} & {9.83} & {35.91} & {64.58} & {23.92} & {236.48} & {\textbf{17.19}} & {36.07} & {20.99} & {11.21} \\
					\multicolumn{1}{l}{Bonython (\#1)} & 73.74  & 73.74  & 1.04  & 1.98  & 23.63  & 2.59  & 3.78  & 1.90  & 44.30  & 59.67  & {2.07} & {9.42} & {15.26} & {37.73} & 0.05  & 60.87  & 10.14  & 1.18  & {\textbf{0.00}} & {2.02} & {5.66} & {22.67} & 2.52  & 1.98  & \textbf{0.00} & \textbf{0.62} \\
					\multicolumn{1}{l}{Elderhalla (\#2)} & 82.24  & 60.75  & 2.90  & 2.51  & 22.57  & 2.27  & 1.64  & 2.41  & 1.31  & 75.08  & {1.40} & {17.85} & {20.98} & {43.12} & 1.40  & 109.03  & 3.13  & 1.61  & {1.73} & {2.35} & {1.54} & {26.61} & 1.26  & 4.24  & \textbf{0.51} & \textbf{0.71} \\
					\multicolumn{1}{l}{{Elderhallb (\#3)}} & {89.02} & {47.84} & {12.94} & {3.57} & {32.69} & {2.62} & {16.29} & {2.14} & {67.27} & {105.09} & {15.84} & {29.88} & {28.34} & {50.37} & {30.94} & {59.11} & {31.18} & {3.76} & {20.53} & {2.48} & {15.18} & {37.02} & {\textbf{8.82}} & {4.62} & {15.57} & {\textbf{1.30}} \\
					\multicolumn{1}{l}{Hartley (\#2)} & 89.69  & 61.56  & 8.67  & 4.45  & 15.37  & 4.94  & 12.51  & \textbf{2.04} & 22.86  & 171.53  & {14.41} & {31.37} & {16.01} & {67.47} & 11.94  & 143.72  & 15.14  & 5.17  & {4.29} & {2.50} & {2.22} & {38.47} & 6.69  & 10.24  & \textbf{1.94} & 2.82  \\
					\multicolumn{1}{l}{Johnsona (\#4)} & 84.45  & 20.91  & 24.53  & 8.92  & 56.35  & 18.56  & 33.43  & \textbf{2.74} & 39.46  & 249.47  & {6.46} & {60.29} & {32.99} & {76.66} & 42.18  & 212.45  & 20.85  & 13.17  & {30.93} & {4.26} & {\textbf{3.43}} & {46.12} & 8.24  & 3.29  & 16.80  & 3.34  \\
					\multicolumn{1}{l}{{Johnsonb (\#7)}} & {97.69} & {12.02} & {26.01} & {28.85} & {51.28} & {28.35} & {49.18} & {\textbf{4.12}} & {32.61} & {866.00} & {35.03} & {221.57} & {28.92} & {148.39} & {46.07} & {73.38} & {23.21} & {6.25} & {42.05} & {5.65} & {\textbf{16.36}} & {123.18} & {23.72} & {10.01} & {19.05} & {9.89} \\
					\multicolumn{1}{l}{{Ladysymon (\#2)}} & {78.06} & {32.49} & {10.66} & {4.41} & {24.67} & {2.34} & {18.33} & {2.48} & {9.38} & {96.99} & {\textbf{5.29}} & {19.21} & {25.99} & {46.11} & {24.45} & {120.15} & {41.01} & {4.61} & {22.11} & {2.51} & {17.05} & {22.33} & {9.07} & {4.48} & {19.49} & {\textbf{1.44}} \\
					\multicolumn{1}{l}{Library (\#2)} & 78.60  & 55.35  & 7.64  & 2.43  & 25.00  & 2.04  & 3.73  & 1.95  & 5.85  & 72.99  & {3.11} & {17.94} & {27.94} & {42.56} & 9.67  & 78.22  & 14.39  & 2.24  & {5.28} & {2.32} & {8.79} & {25.85} & \textbf{2.36} & 2.43  & 5.52  & \textbf{0.76} \\
					\multicolumn{1}{l}{{Napiera (\#2)}} & {90.07} & {62.91} & {11.23} & {3.58} & {32.19} & {3.52} & {25.72} & {\textbf{2.02}} & {\textbf{6.58}} & {146.10} & {12.33} & {25.55} & {14.42} & {61.65} & {11.68} & {59.33} & {28.77} & {5.18} & {10.62} & {2.57} & {11.95} & {42.78} & {9.25} & {2.39} & {10.96} & {3.49} \\
					\multicolumn{1}{l}{Napierb (\#3)} & 86.10  & 39.38  & \textbf{12.24} & 3.77  & 33.38  & 2.75  & 37.13  & 2.84  & 15.49  & 100.88  & {18.99} & {29.08} & {25.60} & {48.49} & 22.78  & 115.39  & 20.08  & 4.26  & {34.43} & {2.42} & {19.85} & {34.18} & 19.45  & 3.00  & 16.50  & \textbf{0.85} \\
					\multicolumn{1}{l}{Neem (\#3)} & 82.16  & 36.51  & 20.96  & 3.73  & 34.43  & 3.47  & 11.78  & 2.42  & 12.65  & 89.41  & {13.57} & {28.54} & {38.26} & {46.79} & 37.65  & 122.42  & 12.43  & 3.05  & {23.35} & {2.62} & {5.93} & {25.17} & 5.23  & 12.62  & \textbf{3.35} & \textbf{0.90} \\
					\multicolumn{1}{l}{{Nese (\#2)}} & {69.69} & {33.46} & {1.20} & {4.80} & {35.85} & {3.39} & {6.18} & {2.74} & {47.88} & {110.60} & {14.11} & {20.46} & {2.86} & {49.40} & {34.48} & {60.85} & {5.19} & {3.03} & {\textbf{0.46}} & {3.73} & {2.56} & {22.22} & {16.72} & {4.74} & {1.72} & {\textbf{1.08}} \\
					\multicolumn{1}{l}{{Oldclassics. (\#2)}} & {81.27} & {32.45} & {19.97} & {12.03} & {15.04} & {5.94} & {4.99} & {\textbf{3.77}} & {32.64} & {267.84} & {2.53} & {34.26} & {3.49} & {80.77} & {19.42} & {75.64} & {31.05} & {4.37} & {17.91} & {6.72} & {2.48} & {43.58} & {\textbf{2.04}} & {34.04} & {2.11} & {4.77} \\
					\multicolumn{1}{l}{Physics (\#1)} & 45.28  & 45.28  & 5.44  & 1.10  & 12.62  & 0.75  & 28.64  & 2.18  & 15.44  & 15.67  & {4.27} & {4.90} & {8.85} & {16.55} & 3.30  & 44.33  & 8.83  & 1.23  & {0.10} & {2.05} & {2.83} & {7.16} & 0.93  & 6.26  & \textbf{0.00} & \textbf{0.62} \\
					\multicolumn{1}{l}{Sene (\#2)} & 81.60  & 47.20  & 1.95  & 4.02  & 12.03  & 2.94  & 7.58  & 2.76  & 0.68  & 101.42  & {2.37} & {19.87} & {9.60} & {51.18} & 18.60  & 118.60  & 0.97  & 2.89  & {0.85} & {2.68} & {2.80} & {25.82} & 0.60  & 9.70  & \textbf{0.30} & \textbf{0.89} \\
					\multicolumn{1}{l}{{Unihouse (\#5)}} & {95.83} & {16.55} & {22.57} & {215.76} & {49.15} & {144.76} & {12.28} & {\textbf{7.36}} & {16.03} & {7693.00} & {10.62} & {1223.35} & {26.55} & {572.69} & {42.24} & {1186.51} & {27.95} & {13.41} & {24.70} & {225.80} & {\textbf{5.76}} & {1021.85} & {12.17} & {23.67} & {17.62} & {21.11} \\
					\multicolumn{1}{l}{Unionhouse (\#1)} & 76.51  & 76.51  & 3.12  & 4.01  & 21.59  & 4.32  & 1.99  & \textbf{1.90} & 7.73  & 171.35  & {1.56} & {16.73} & {17.76} & {71.70} & \textbf{0.31} & 69.71  & 23.52  & 3.57  & {0.81} & {2.28} & {5.54} & {48.19} & 0.71  & 6.29  & 1.59  & 2.93  \\
					\midrule
					{Std.} & {} & {} & {11.60} & {-} & {13.61} & {-} & {13.42} & {-} & {18.94} & {-} & {11.04} & {-} & {11.38} & {-} & {15.41} & {-} & {12.29} & {-} & {14.13} & {-} & {8.40} & {-} & {\textbf{6.96}} & {-} & {8.04} & {-} \\
					{Total median} & {} & {} & {11.23} & {-} & {29.66} & {-} & {12.28} & {-} & {15.49} & {-} & {10.62} & {-} & {20.98} & {-} & {19.42} & {-} & {20.08} & {-} & {17.91} & {-} & {5.76} & {-} & {8.24} & {-} & {\textbf{5.52}} & {-} \\
					{Total average} & {} & {} & {13.59} & {20.44} & {30.47} & {15.75} & {16.26} & {\textbf{2.95}} & {22.40} & {665.41} & {11.83} & {119.71} & {21.34} & {95.11} & {21.54} & {180.38} & {20.04} & {4.76} & {15.82} & {18.00} & {9.67} & {98.87} & {\textbf{8.27}} & {9.53} & {8.77} & {3.66} \\
					\bottomrule
				\end{tabular}
			}
		\end{center}
		\vspace{-1ex}
	\end{table*}
	
	\begin{table*}[!ht]
		\vspace{-2.5ex}
		\caption{Quantitative comparison results of two-view motion segmentation on nineteen image pairs. \# indicates the true number of model instances included in data. Mean and Time represent the mean fitting error and the mean CPU time, respectively. The best results are boldfaced.}
		\label{tableFundamental}
		\vspace{-5.5ex}
		\begin{center}
			\scalebox{0.8}{\tabcolsep0.0215in
				\begin{tabular}{cccrrrrrrrrrrrrrrrrrrrrrrrr}
					\toprule
					\multicolumn{1}{c}{\multirow{2}[4]{*}{Data}} & \multicolumn{1}{c}{\multirow{2}[4]{*}{\shortstack{Outlier\\rate \\(\%)}}} & \multicolumn{1}{c}{\multirow{2}[4]{*}{\shortstack{Gross\\outlier\\rate (\%)}}} & \multicolumn{2}{c}{J-Linkage} & \multicolumn{2}{c}{KF} & \multicolumn{2}{c}{AKSWH} & \multicolumn{2}{c}{T-Linkage} & \multicolumn{2}{c}{{RPA}} & \multicolumn{2}{c}{{DPA}} & \multicolumn{2}{c}{RansaCov} & \multicolumn{2}{c}{TSMP} & \multicolumn{2}{c}{{HVF}} & \multicolumn{2}{c}{{DGSAC}} & \multicolumn{2}{c}{MSHF} & \multicolumn{2}{c}{HRMP} \\
					\cmidrule{4-27}          &       &       & \multicolumn{1}{c}{Mean} & \multicolumn{1}{c}{Time} & \multicolumn{1}{c}{Mean} & \multicolumn{1}{c}{Time} & \multicolumn{1}{c}{Mean} & \multicolumn{1}{c}{Time} & \multicolumn{1}{c}{Mean} & \multicolumn{1}{c}{Time} & \multicolumn{1}{c}{{Mean}} & \multicolumn{1}{c}{{Time}} & \multicolumn{1}{c}{{Mean}} & \multicolumn{1}{c}{{Time}} & \multicolumn{1}{c}{Mean} & \multicolumn{1}{c}{Time} & \multicolumn{1}{c}{Mean} & \multicolumn{1}{c}{Time} & \multicolumn{1}{c}{{Mean}} & \multicolumn{1}{c}{{Time}} & \multicolumn{1}{c}{{Mean}} & \multicolumn{1}{c}{{Time}} & \multicolumn{1}{c}{Mean} & \multicolumn{1}{c}{Time} & \multicolumn{1}{c}{Mean} & \multicolumn{1}{c}{Time} \\
					\midrule
					\multicolumn{1}{l}{Biscuit (\#1)} & 55.76  & 55.76  & 19.34  & 6.33  & 0.72  & 3.97  & 16.77  & 2.39  & 22.79  & 202.93  & {7.65} & {10.57} & {6.96} & {22.61} & 3.01  & 700.39  & 5.49  & 6.01  & {4.92} & {2.94} & {1.88} & {178.73} & 5.39  & 6.04  & \textbf{0.13} & \textbf{1.88} \\
					\multicolumn{1}{l}{Biscuitbook (\#2)} & 75.95  & 47.51  & 12.49  & 9.63  & 19.94  & 4.85  & 9.97  & 2.41  & 10.91  & 253.57  & {5.81} & {17.59} & {10.09} & {24.28} & 9.77  & 695.87  & 1.00  & 7.45  & {9.33} & {3.31} & {1.35} & {170.87} & 2.55  & 6.28  & \textbf{0.09} & \textbf{1.89} \\
					\multicolumn{1}{l}{Biscuitbookbox (\#3)} & 84.17  & 37.45  & 12.29  & 6.12  & 20.89  & 3.71  & 25.74  & 1.90  & 13.33  & 143.33  & {9.30} & {18.10} & {13.18} & {16.01} & 16.01  & 674.11  & 8.49  & 7.23  & {37.05} & {2.23} & {10.12} & {108.67} & \textbf{2.65} & 5.35  & 13.64  & \textbf{1.69} \\
					\multicolumn{1}{l}{{Boardgame (\#3)}} & {89.61} & {40.50} & {\textbf{11.43}} & {6.02} & {34.96} & {3.16} & {21.13} & {2.25} & {24.36} & {149.14} & {14.21} & {17.66} & {23.38} & {16.87} & {27.11} & {24.89} & {34.85} & {4.88} & {36.95} & {\textbf{2.22}} & {17.31} & {139.87} & {19.14} & {9.03} & {17.63} & {2.31} \\
					\multicolumn{1}{l}{{Book (\#1)}} & {43.85} & {43.85} & {14.11} & {2.05} & {4.97} & {1.55} & {36.16} & {1.62} & {22.54} & {61.55} & {4.43} & {5.22} & {5.84} & {9.67} & {3.30} & {17.13} & {15.24} & {2.76} & {16.76} & {1.99} & {1.28} & {60.06} & {24.22} & {7.57} & {\textbf{0.59}} & {\textbf{0.68}} \\
					\multicolumn{1}{l}{{Breadcartoychips (\#4)}} & {90.30} & {34.60} & {24.20} & {4.53} & {30.26} & {8.32} & {25.32} & {1.67} & {16.75} & {109.03} & {14.63} & {19.72} & {18.87} & {14.05} & {19.96} & {19.18} & {23.68} & {3.48} & {31.60} & {2.36} & {\textbf{12.83}} & {67.72} & {28.87} & {11.64} & {19.74} & {\textbf{0.90}} \\
					\multicolumn{1}{l}{{Breadcube (\#2)}} & {73.97} & {31.82} & {14.21} & {4.85} & {11.85} & {2.47} & {17.81} & {1.64} & {10.99} & {113.11} & {6.61} & {11.33} & {7.21} & {14.18} & {16.35} & {22.89} & {12.36} & {5.71} & {20.21} & {2.26} & {\textbf{2.23}} & {77.10} & {9.79} & {11.62} & {7.08} & {\textbf{1.09}} \\
					\multicolumn{1}{l}{Breadcubechips (\#3)} & 85.22  & 35.22  & \textbf{10.48} & 4.73  & 24.83  & 3.05  & 22.74  & 1.71  & 18.04  & 111.84  & {14.35} & {15.22} & {20.70} & {13.91} & 27.13  & 608.14  & 16.30  & 4.25  & {25.13} & {2.10} & {16.09} & {57.96} & 10.54  & 4.35  & 14.39  & \textbf{0.54} \\
					\multicolumn{1}{l}{{Breadtoy (\#2)}} & {79.86} & {36.81} & {11.22} & {5.21} & {10.00} & {3.28} & {34.10} & {\textbf{1.85}} & {\textbf{7.30}} & {155.29} & {27.91} & {13.58} & {19.35} & {17.58} & {22.99} & {29.51} & {16.04} & {5.90} & {21.33} & {2.26} & {20.80} & {120.75} & {26.08} & {14.72} & {11.80} & {2.24} \\
					\multicolumn{1}{l}{Breadtoycar (\#3)} & 79.52  & 33.73  & 18.60  & 2.67  & 27.99  & 2.67  & 21.34  & 1.56  & 18.41  & 53.71  & {14.15} & {12.73} & {22.07} & {8.63} & 18.23  & 566.45  & \textbf{3.90} & 2.43  & {26.40} & {1.94} & {8.55} & {28.73} & 14.13  & 4.00  & 5.30  & \textbf{0.49} \\
					\multicolumn{1}{l}{{Carchipscube (\#3)}} & {88.48} & {36.36} & {13.90} & {2.14} & {\textbf{9.51}} & {1.91} & {37.99} & {1.47} & {22.50} & {50.37} & {20.37} & {11.75} & {17.93} & {7.95} & {24.88} & {13.77} & {26.65} & {2.69} & {36.89} & {1.64} & {16.79} & {40.01} & {17.38} & {6.31} & {22.56} & {\textbf{0.74}} \\
					\multicolumn{1}{l}{Cube (\#1)} & 67.88  & 67.88  & 24.34  & 6.28  & 8.78  & 3.30  & 9.53  & \textbf{1.86} & 23.69  & 176.16  & {8.47} & {71.05} & {10.03} & {20.31} & 5.76  & 658.34  & 12.73  & 5.23  & {19.80} & {2.45} & {2.91} & {143.77} & 2.28  & 2.51  & \textbf{1.63} & 2.04  \\
					\multicolumn{1}{l}{Cubebreadtoyc. (\#4)} & 88.38  & 26.91  & 22.32  & 8.07  & 32.29  & 12.63  & 18.69  & \textbf{1.52} & 17.93  & 210.80  & {9.55} & {97.28} & {20.32} & {21.84} & 20.61  & 662.95  & 9.04  & 7.31  & {25.10} & {2.30} & {13.52} & {92.89} & 16.51  & 5.17  & \textbf{6.24} & 1.85  \\
					\multicolumn{1}{l}{{Cubechips (\#2)}} & {79.93} & {50.35} & {12.35} & {6.12} & {6.03} & {3.24} & {13.50} & {\textbf{1.91}} & {8.81} & {159.63} & {9.03} & {66.81} & {8.01} & {17.89} & {12.13} & {26.24} & {10.97} & {5.25} & {16.21} & {2.36} & {\textbf{2.92}} & {130.23} & {4.26} & {14.22} & {7.80} & {2.43} \\
					\multicolumn{1}{l}{Cubetoy (\#2)} & 71.08  & 39.76  & 16.86  & 4.91  & 11.63  & 3.17  & 19.21  & 2.10  & 7.99  & 120.45  & {6.28} & {48.78} & {6.11} & {14.68} & 8.03  & 639.32  & 4.81  & 3.16  & {8.20} & {2.39} & {2.77} & {67.27} & 4.42  & 5.18  & \textbf{0.13} & \textbf{0.50} \\
					\multicolumn{1}{l}{{Dinobooks (\#3)}} & {88.61} & {43.06} & {24.10} & {10.12} & {\textbf{20.38}} & {8.75} & {25.78} & {\textbf{1.67}} & {27.52} & {254.77} & {21.24} & {105.77} & {26.73} & {24.32} & {27.91} & {34.30} & {40.48} & {10.22} & {37.26} & {2.29} & {24.28} & {148.05} & {21.71} & {9.65} & {34.12} & {2.98} \\
					\multicolumn{1}{l}{Game (\#1)} & 72.96  & 72.96  & 22.13  & 4.78  & 29.70  & 2.21  & 8.87  & 1.66  & 35.09  & 108.04  & {7.39} & {42.00} & {30.61} & {14.04} & 6.52  & 584.33  & 4.01  & 1.79  & {22.61} & {2.31} & {7.04} & {58.73} & 3.52  & 3.41  & \textbf{0.13} & \textbf{0.53} \\
					\multicolumn{1}{l}{Gamebiscuit (\#2)} & 77.74  & 50.91  & 12.25  & 8.49  & 23.49  & 4.11  & 22.41  & 2.18  & 10.59  & 228.80  & {7.10} & {87.75} & {4.94} & {22.25} & 9.10  & 698.33  & 5.31  & 6.29  & {24.94} & {2.36} & {2.29} & {185.73} & 3.94  & 5.58  & \textbf{0.22} & \textbf{1.85} \\
					\multicolumn{1}{l}{{Toycubecar (\#3)}} & {93.00} & {36.00} & {19.04} & {2.94} & {23.79} & {2.74} & {42.93} & {1.76} & {17.78} & {76.61} & {\textbf{11.62}} & {36.44} & {15.45} & {10.76} & {31.97} & {17.19} & {42.66} & {4.02} & {24.80} & {1.86} & {15.90} & {59.72} & {20.76} & {9.11} & {22.85} & {\textbf{0.87}} \\
					\midrule
					{Std.} & {} & {} & {\textbf{4.94}} & {-} & {10.31} & {-} & {9.67} & {-} & {7.44} & {-} & {6.16} & {-} & {7.80} & {-} & {9.13} & {-} & {12.59} & {-} & {9.79} & {-} & {7.48} & {-} & {9.06} & {-} & {9.96} & {-} \\
					{Total median} & {} & {} & {14.21} & {-} & {20.38} & {-} & {21.34} & {-} & {17.93} & {-} & {9.30} & {-} & {15.45} & {-} & {16.35} & {-} & {12.36} & {-} & {24.80} & {-} & {8.55} & {-} & {10.54} & {-} & {\textbf{7.08}} & {-} \\
					{Total average} & {} & {} & {16.61} & {5.58} & {18.53} & {4.16} & {22.63} & {1.85} & {17.75} & {144.16} & {11.58} & {37.33} & {15.15} & {16.41} & {16.36} & {352.28} & {15.48} & {5.06} & {23.45} & {2.29} & {\textbf{9.52}} & {101.94} & {12.53} & {7.46} & {9.79} & {\textbf{1.45}} \\
					\bottomrule
				\end{tabular}
			}%
		\end{center}
		\vspace{-4ex}
	\end{table*}				
	%
	From Table \ref{tableHomography}, we can see that HRMP obtains the lowest total median fitting error, the second lowest standard deviation of fitting errors and the second lowest total average fitting error among all the competing methods. 
	This demonstrates that HRMP can effectively remove gross outliers by using the proposed hierarchical message propagation algorithm and cluster the remaining data points by using the improved affinity propagation algorithm. MSHF obtains the best standard deviation of fitting errors and the best total average fitting error, because it improves the fitting accuracy by preserving the inliers corresponding to the significant model hypotheses. However, MSHF is about 2.6 times slower than the proposed HRMP.
	\begin{figure}[!t]
		\vspace{-3.5ex}
		\begin{center}
			\subfigure[Bonython]{
				\begin{minipage}[b]{0.13\textwidth}
					\includegraphics[width=1.\textwidth]{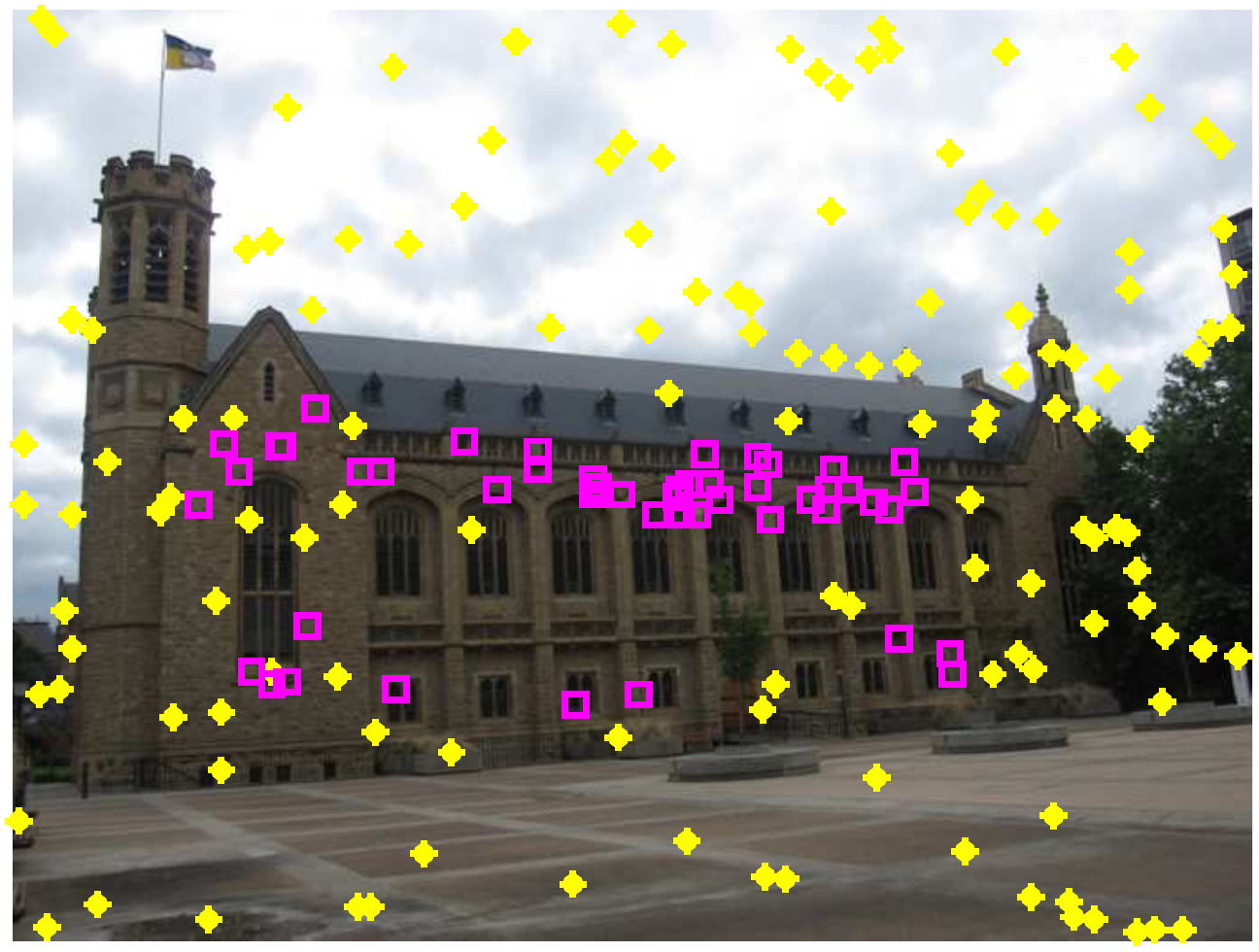}\vspace{0.5ex}
					\includegraphics[width=1.\textwidth]{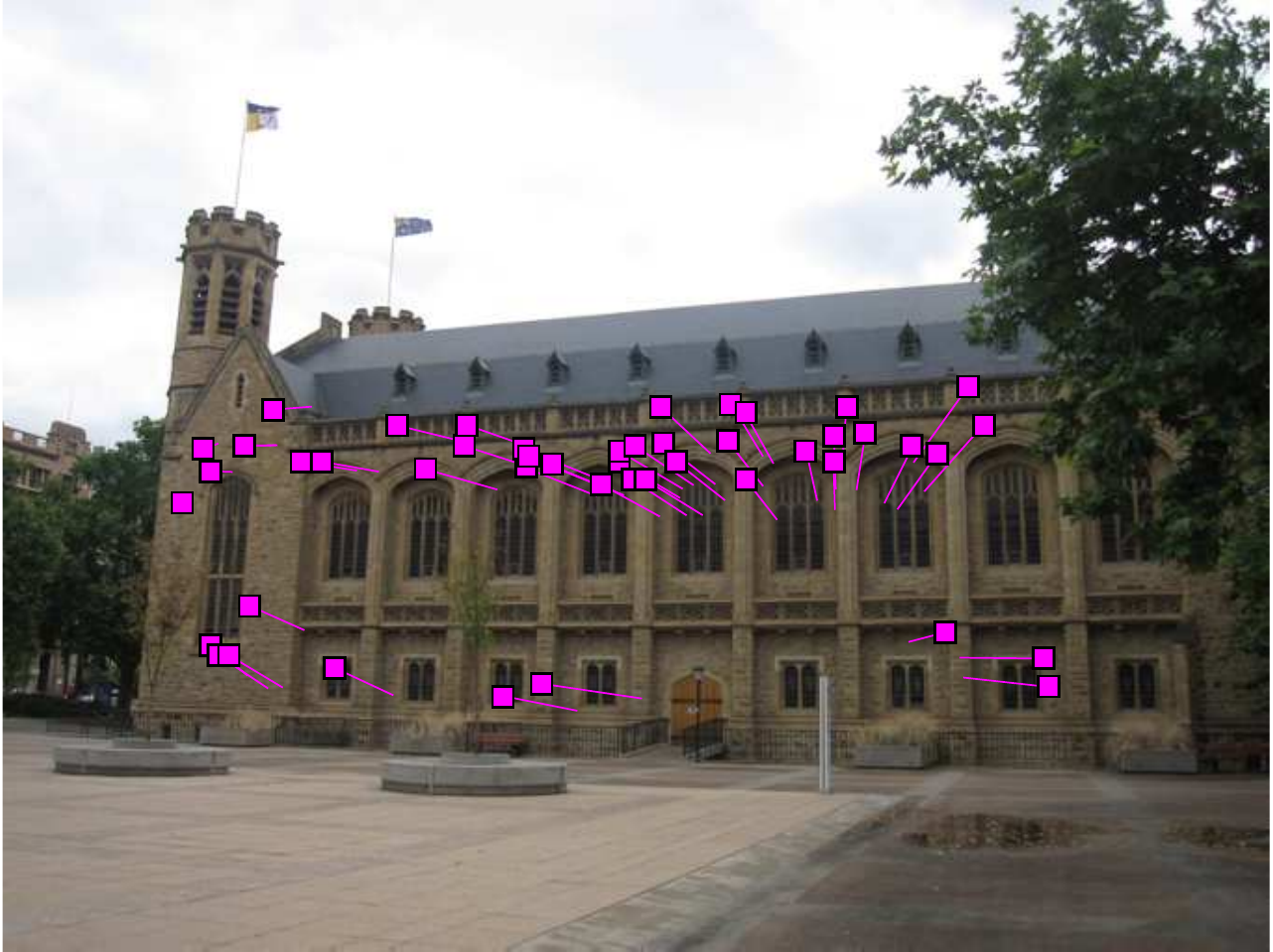}
			\end{minipage}}
			\subfigure[Elderhalla]{
				\begin{minipage}[b]{0.13\textwidth}
					\includegraphics[width=1.\textwidth]{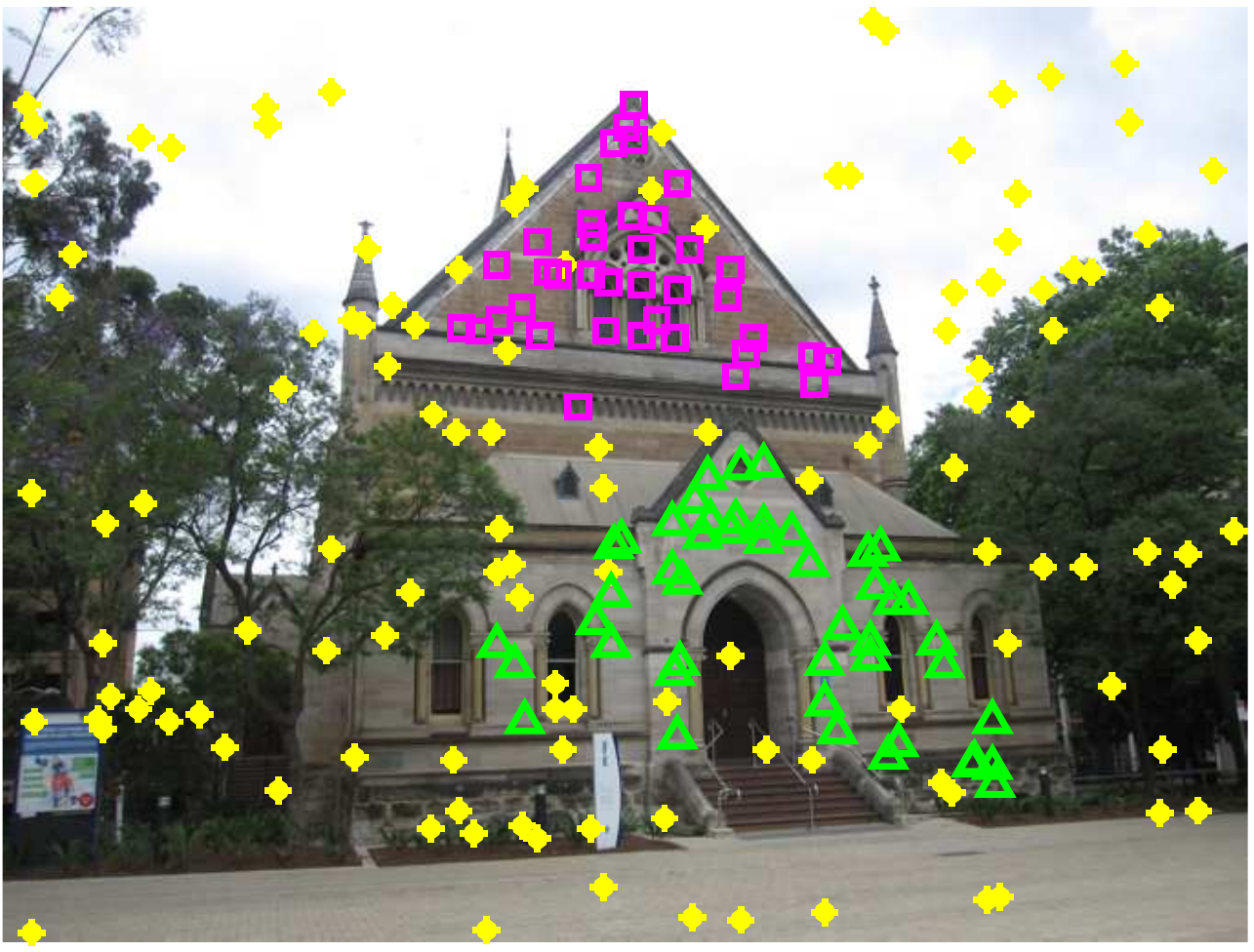}\vspace{0.5ex}
					\includegraphics[width=1.\textwidth]{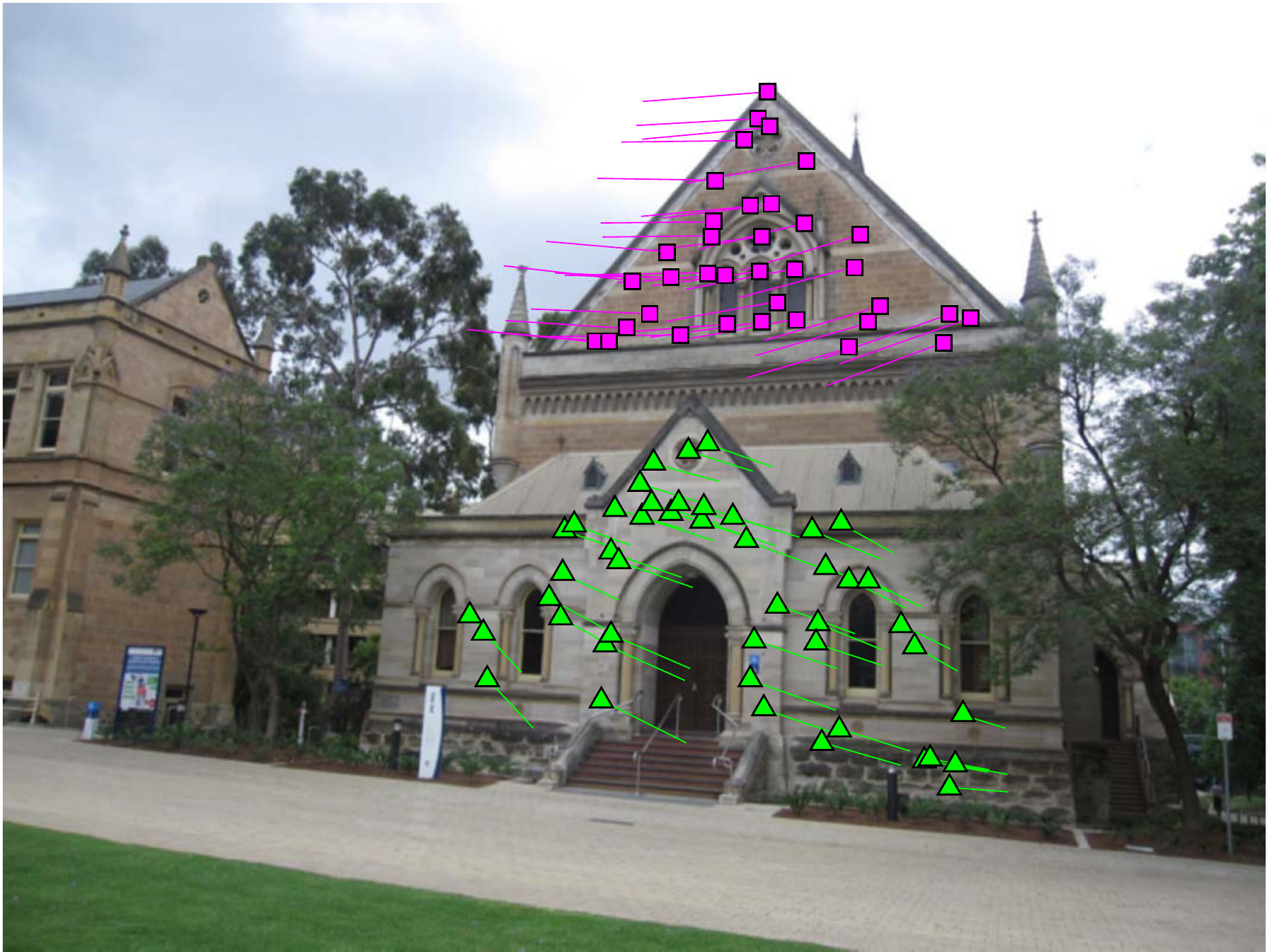}
			\end{minipage}}
			\subfigure[Neem]{
				\begin{minipage}[b]{0.13\textwidth}
					\includegraphics[width=1.\textwidth]{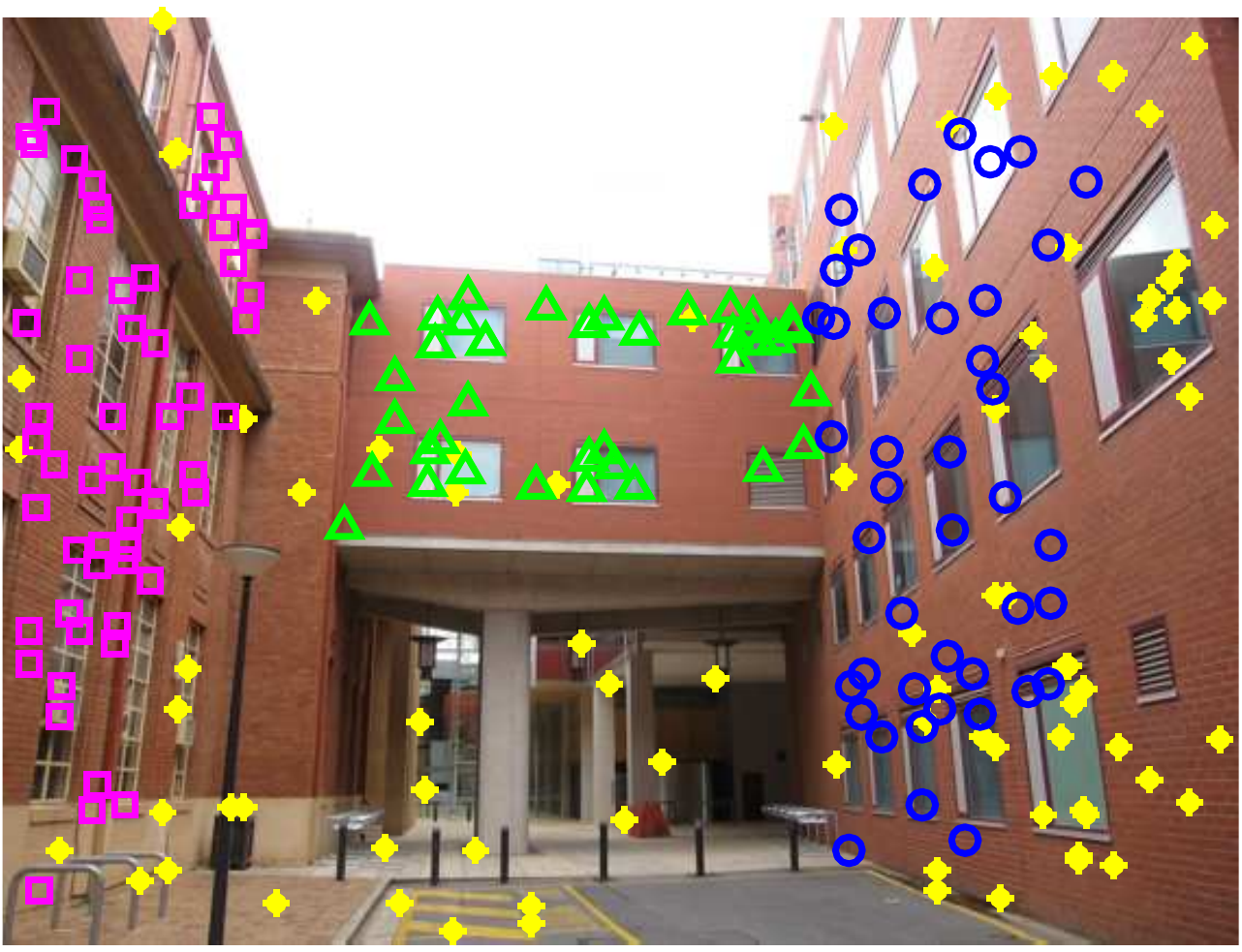}\vspace{0.5ex}
					\includegraphics[width=1.\textwidth]{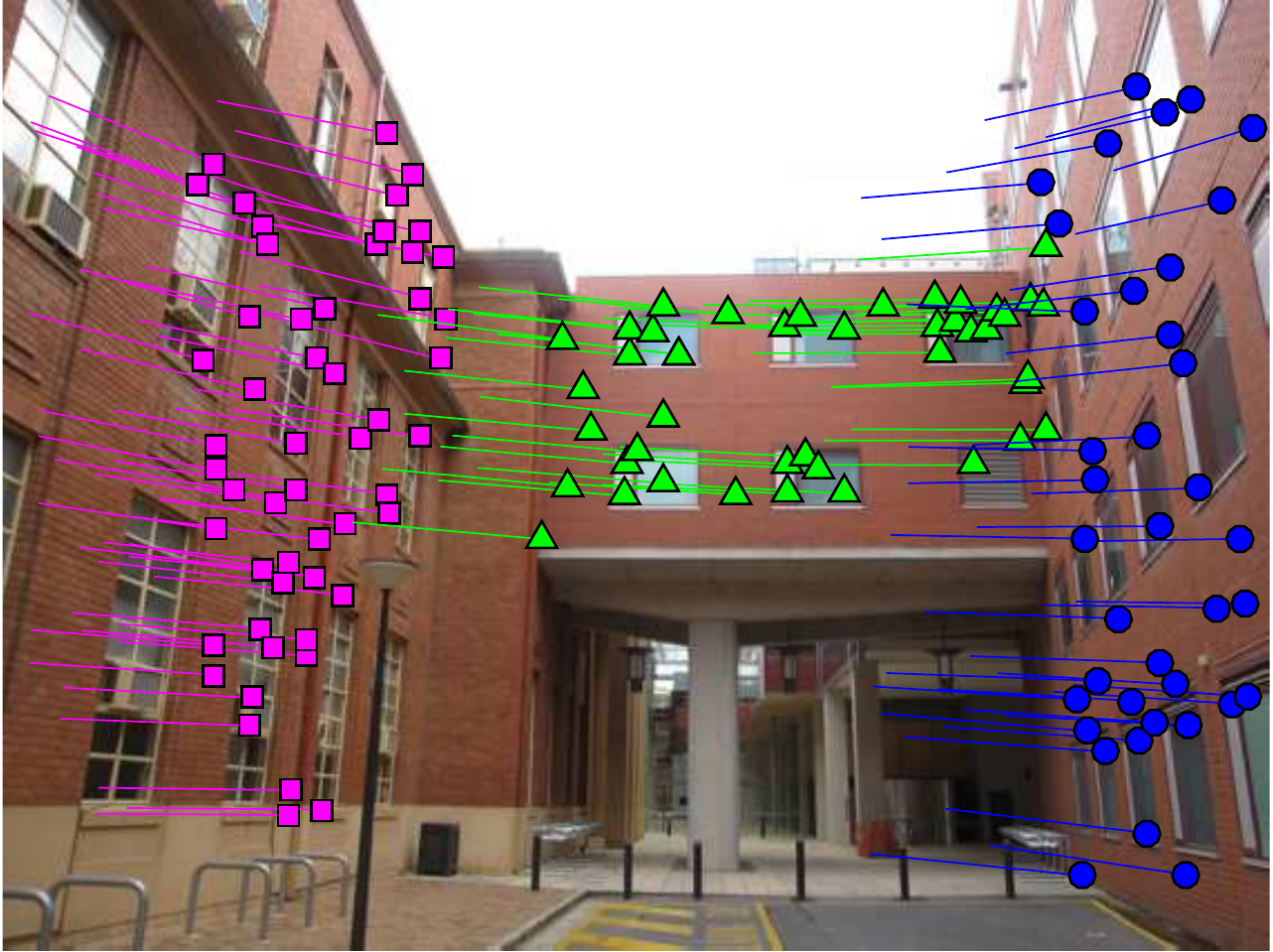}
			\end{minipage}}	
			\vspace{-1.5ex}
			\caption{Examples for multi-homography segmentation on three image pairs. The $1^{st}$ and $2^{nd}$ rows are the Ground-truth segmentation results and the segmentation results obtained by HRMP on the image pairs. The gross outliers are marked in the yellow color. The inliers belonging to different model instances are marked in other different colors, respectively.}
			\label{homo}
		\end{center}
		\vspace{-5ex}
	\end{figure}
	\begin{figure}[!t]
		\vspace{-3.5ex}
		\begin{center}
			\subfigure[Game]{
				\begin{minipage}[b]{0.13\textwidth}
					\includegraphics[width=1.\textwidth]{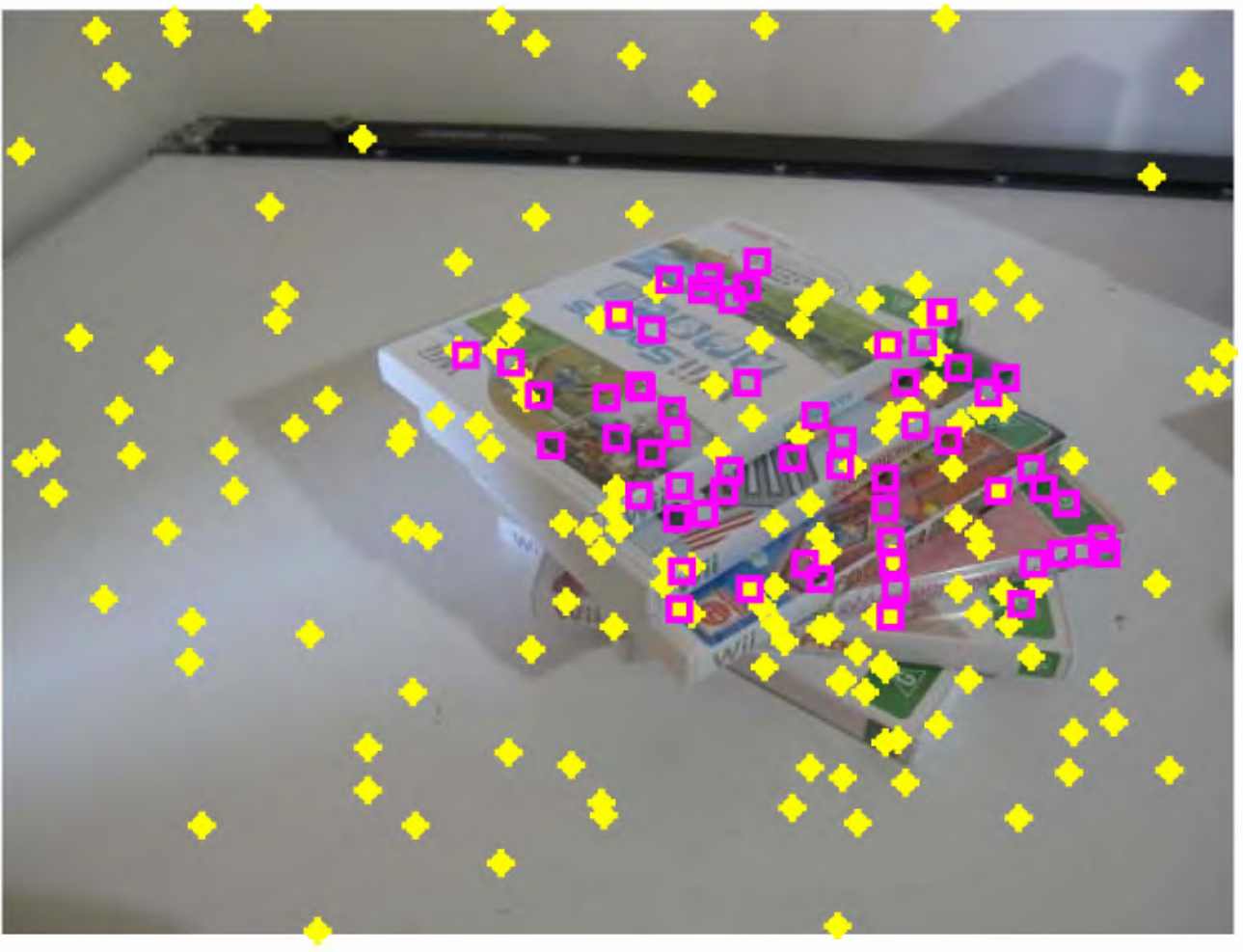}\vspace{0.5ex}
					\includegraphics[width=1.\textwidth]{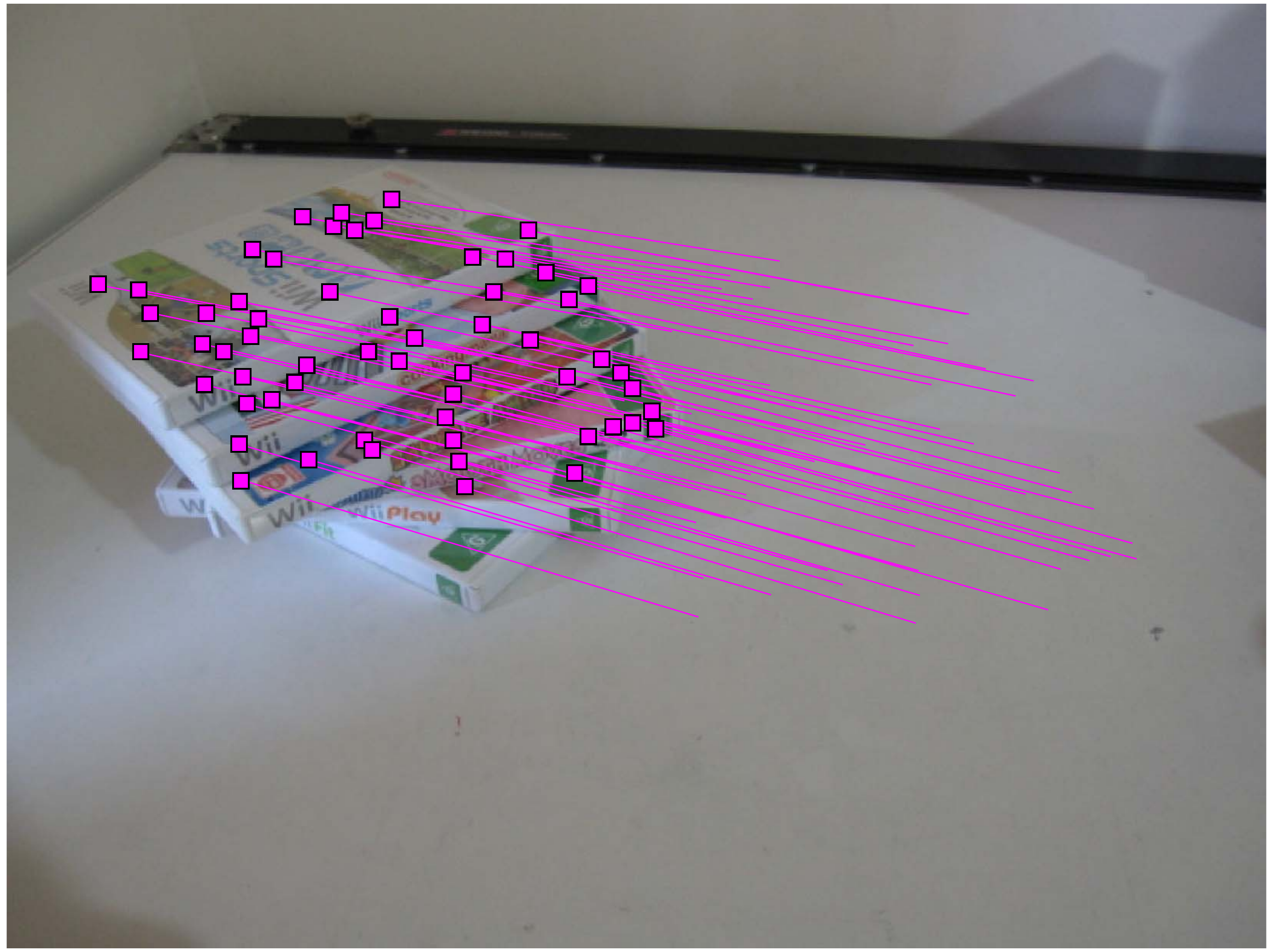}
			\end{minipage}}
			\subfigure[Biscuitbook]{
				\begin{minipage}[b]{0.13\textwidth}
					\includegraphics[width=1.\textwidth]{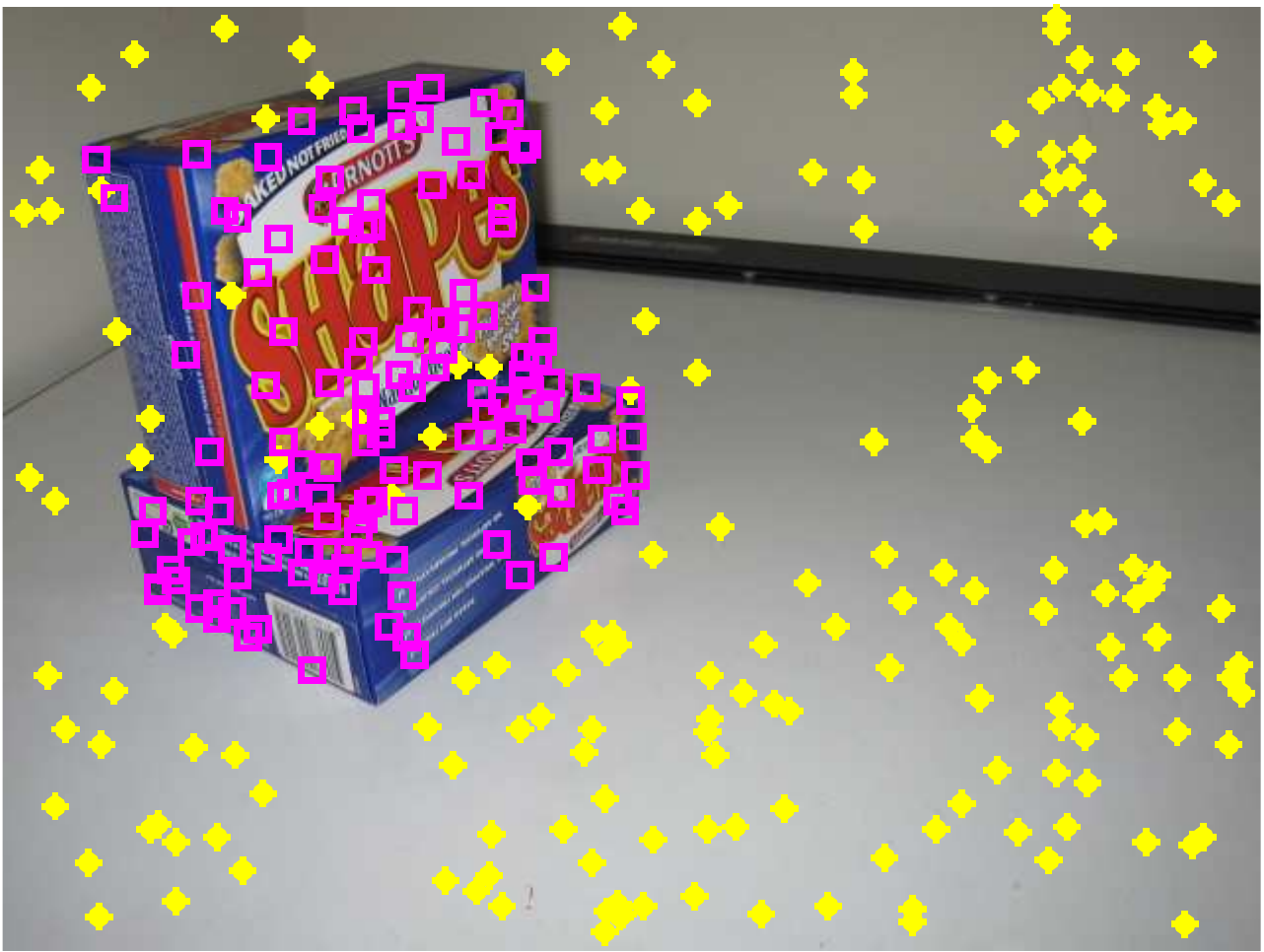}\vspace{0.5ex}
					\includegraphics[width=1.\textwidth]{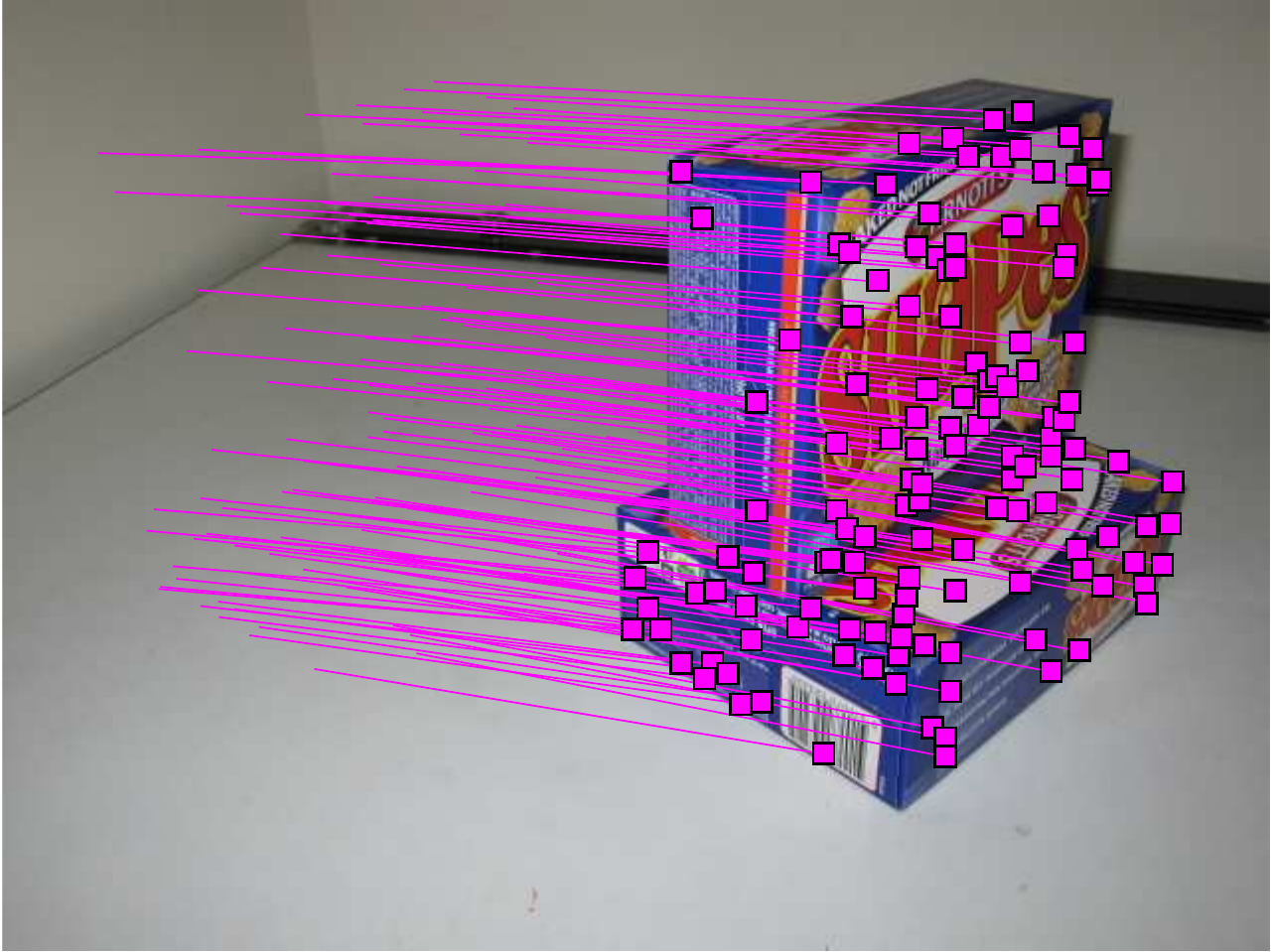}
			\end{minipage}}	
			\subfigure[Breadtoycar]{
				\begin{minipage}[b]{0.13\textwidth}
					\includegraphics[width=1.\textwidth]{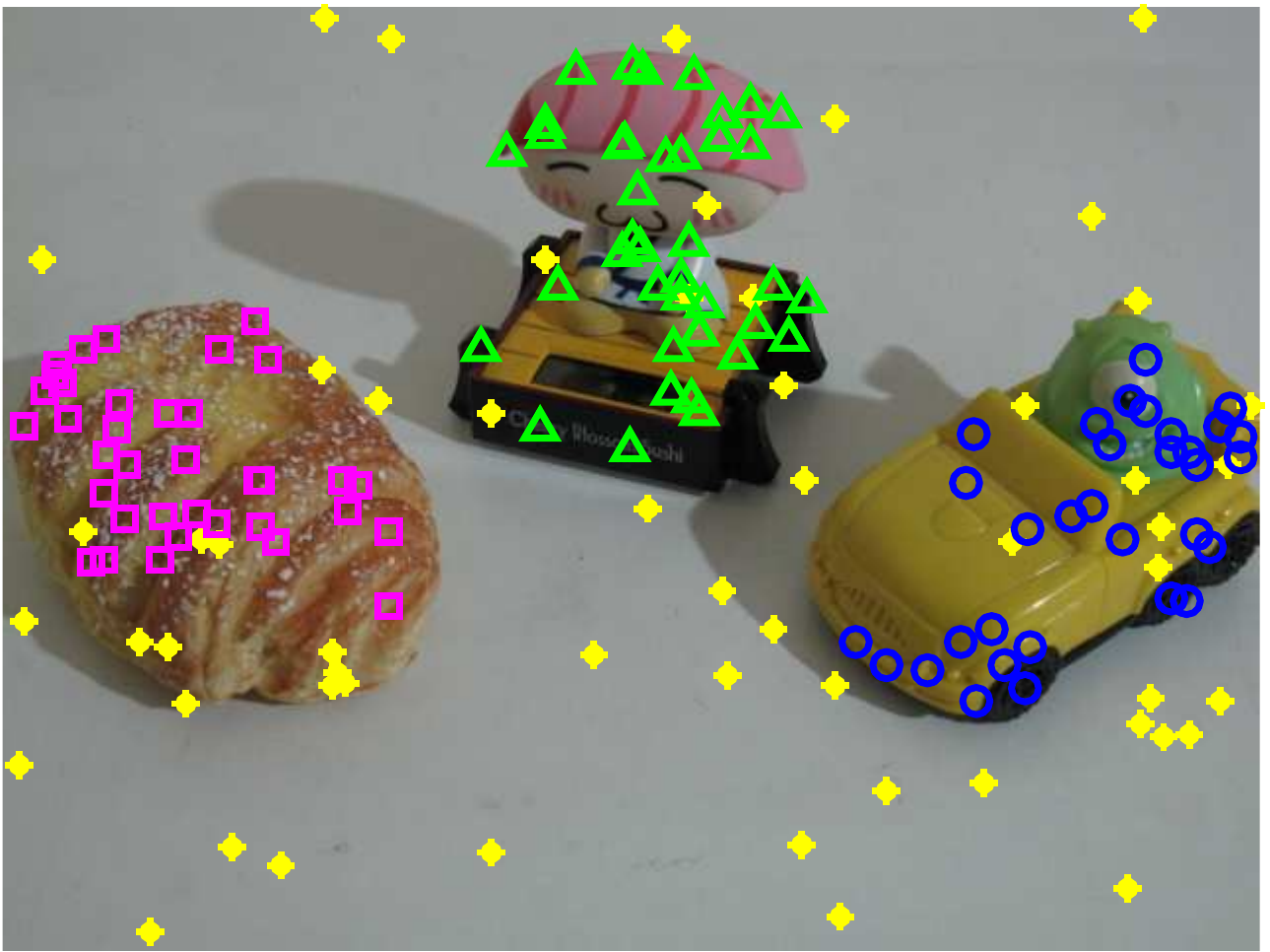}\vspace{0.5ex}
					\includegraphics[width=1.\textwidth]{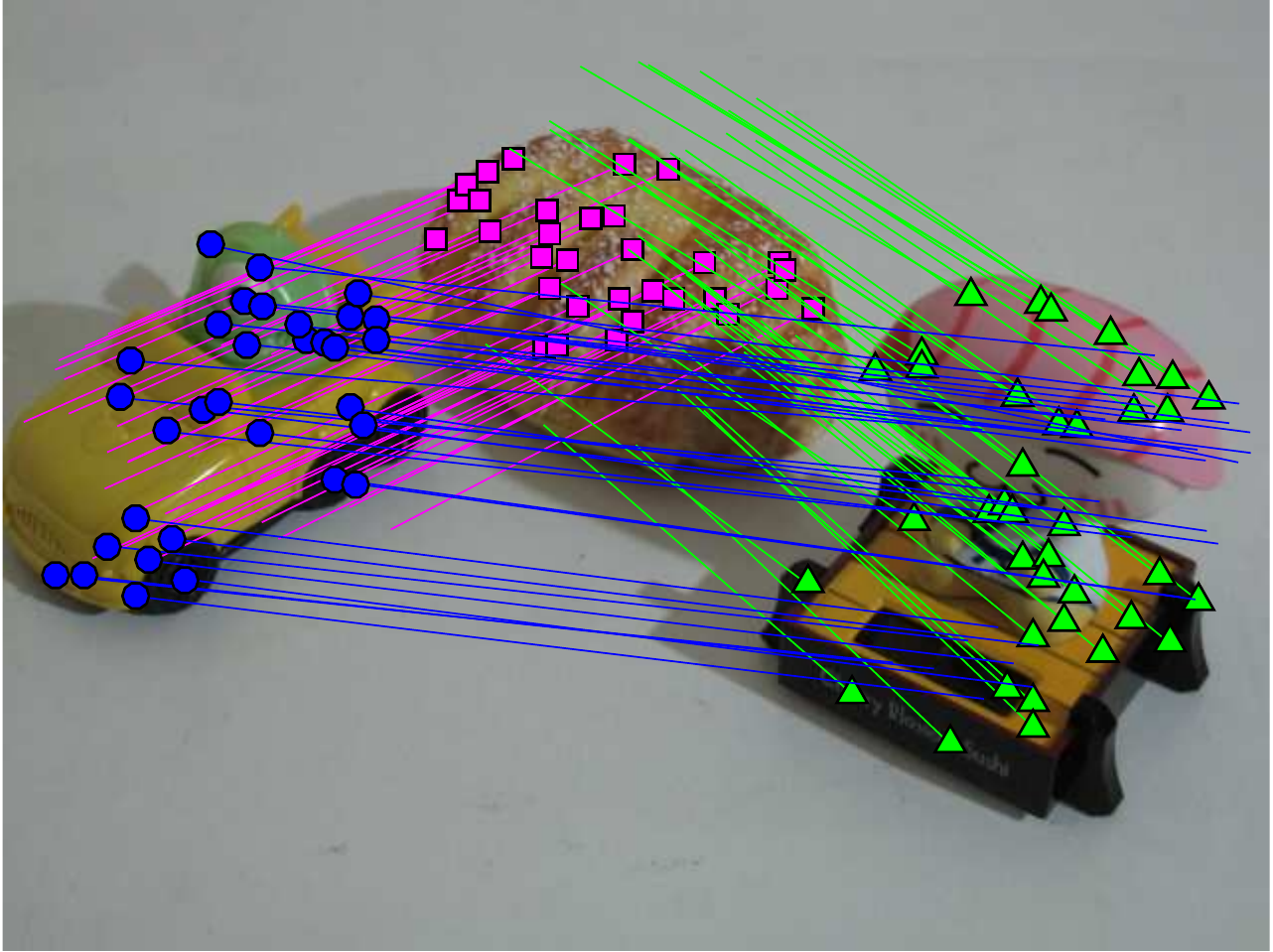}
			\end{minipage}}
			\vspace{-1.5ex}
			\caption{Examples for two-view motion segmentation on three image pairs. The $1^{st}$ and $2^{nd}$ rows are the Ground-truth segmentation results and the segmentation results obtained by HRMP on the image pairs. The gross outliers are marked in the yellow color. The inliers belonging to different model instances are marked in other different colors, respectively.}
			\label{fun}
		\end{center}
		\vspace{-4ex}
	\end{figure}
	
	\begin{figure}[!htp]
		\vspace{-3ex}
		\begin{center}\hspace{-2.3ex}
			\subfigure[Masonry nails]{
				\begin{minipage}[b]{0.125\textwidth}
					\includegraphics[width=1.\textwidth]{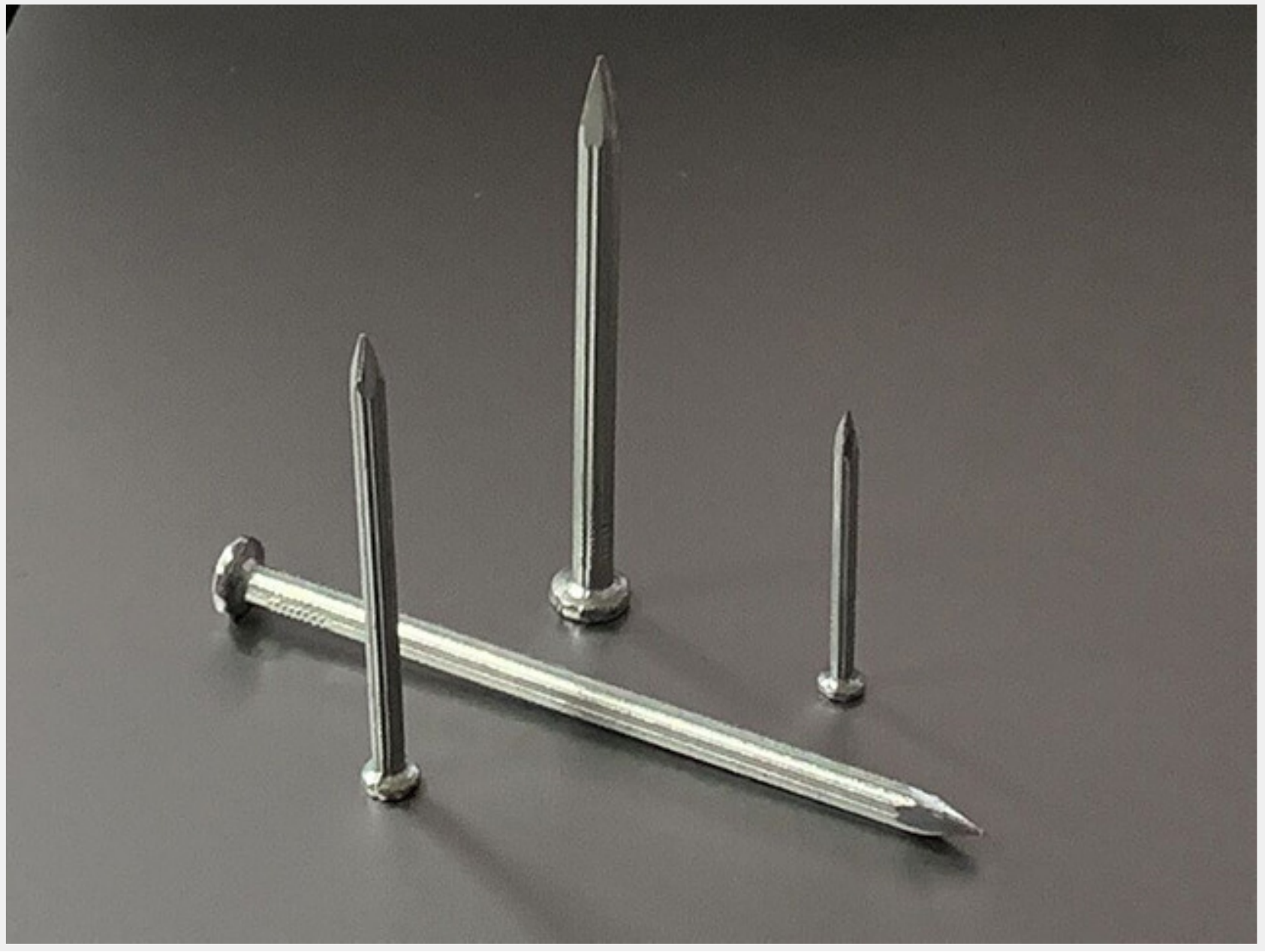}\vspace{0.2ex}
					\includegraphics[width=1.\textwidth]{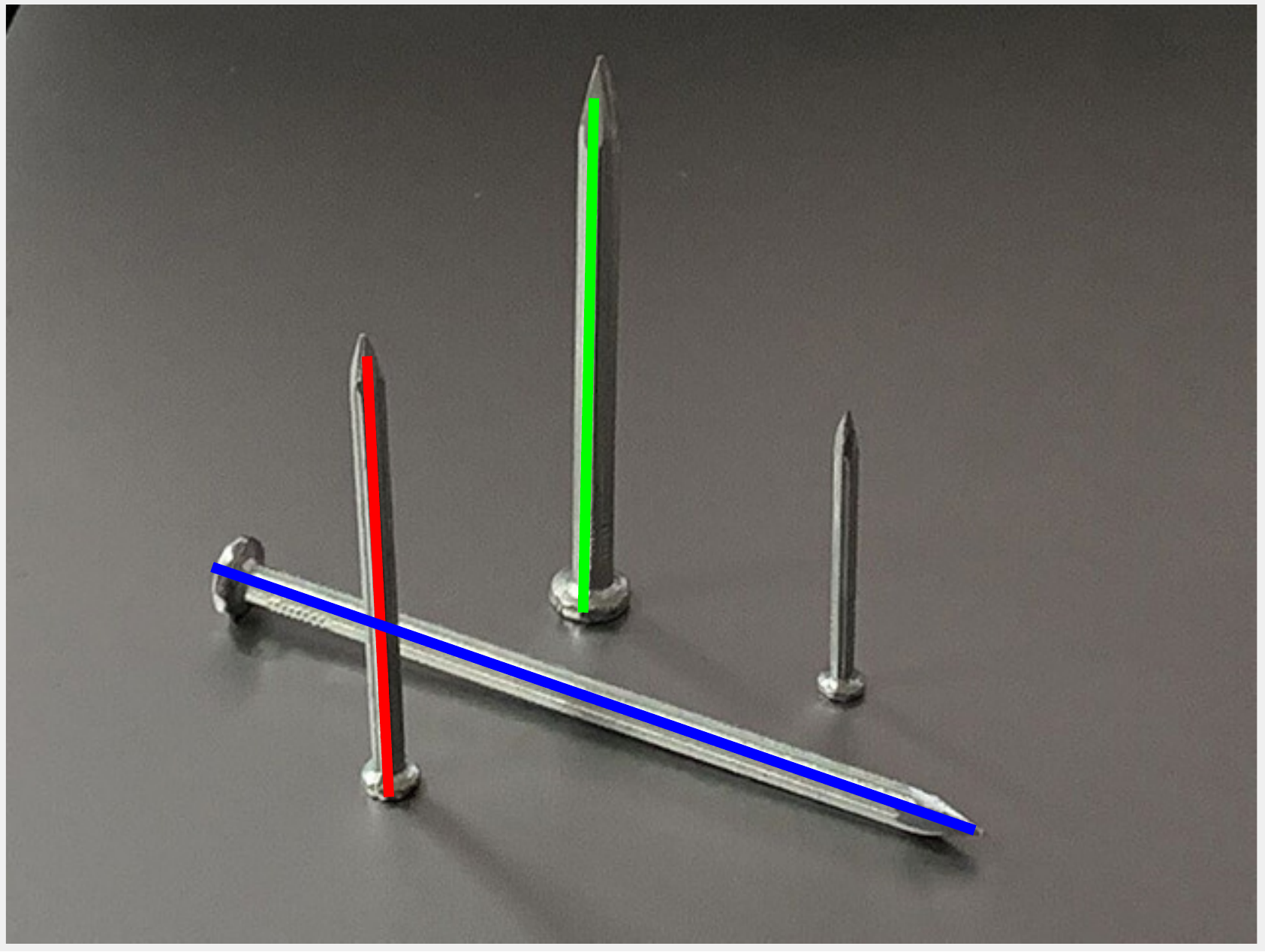}\vspace{0.2ex}
					\includegraphics[width=1.\textwidth]{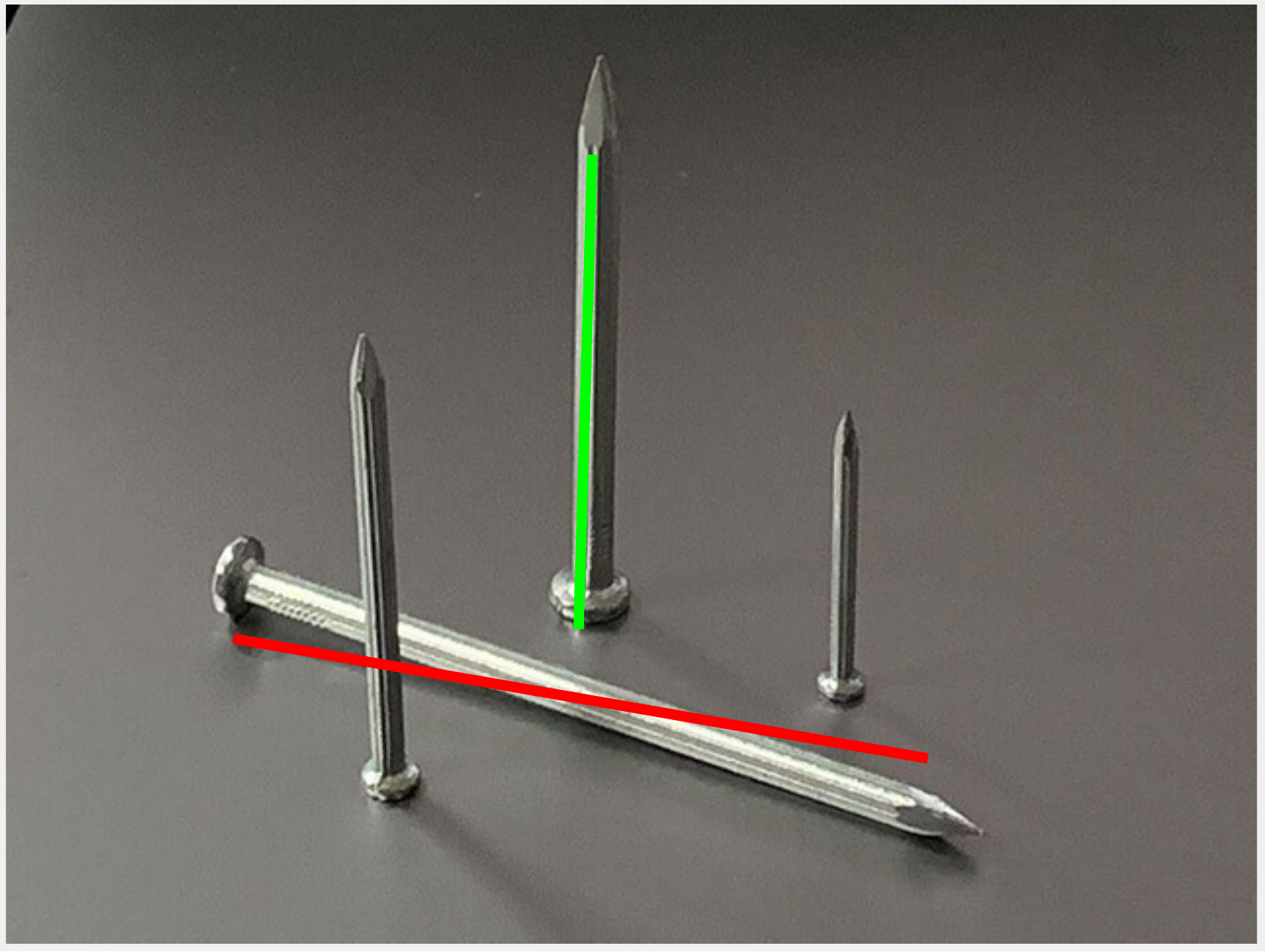}\vspace{0.2ex}
					\includegraphics[width=1.\textwidth]{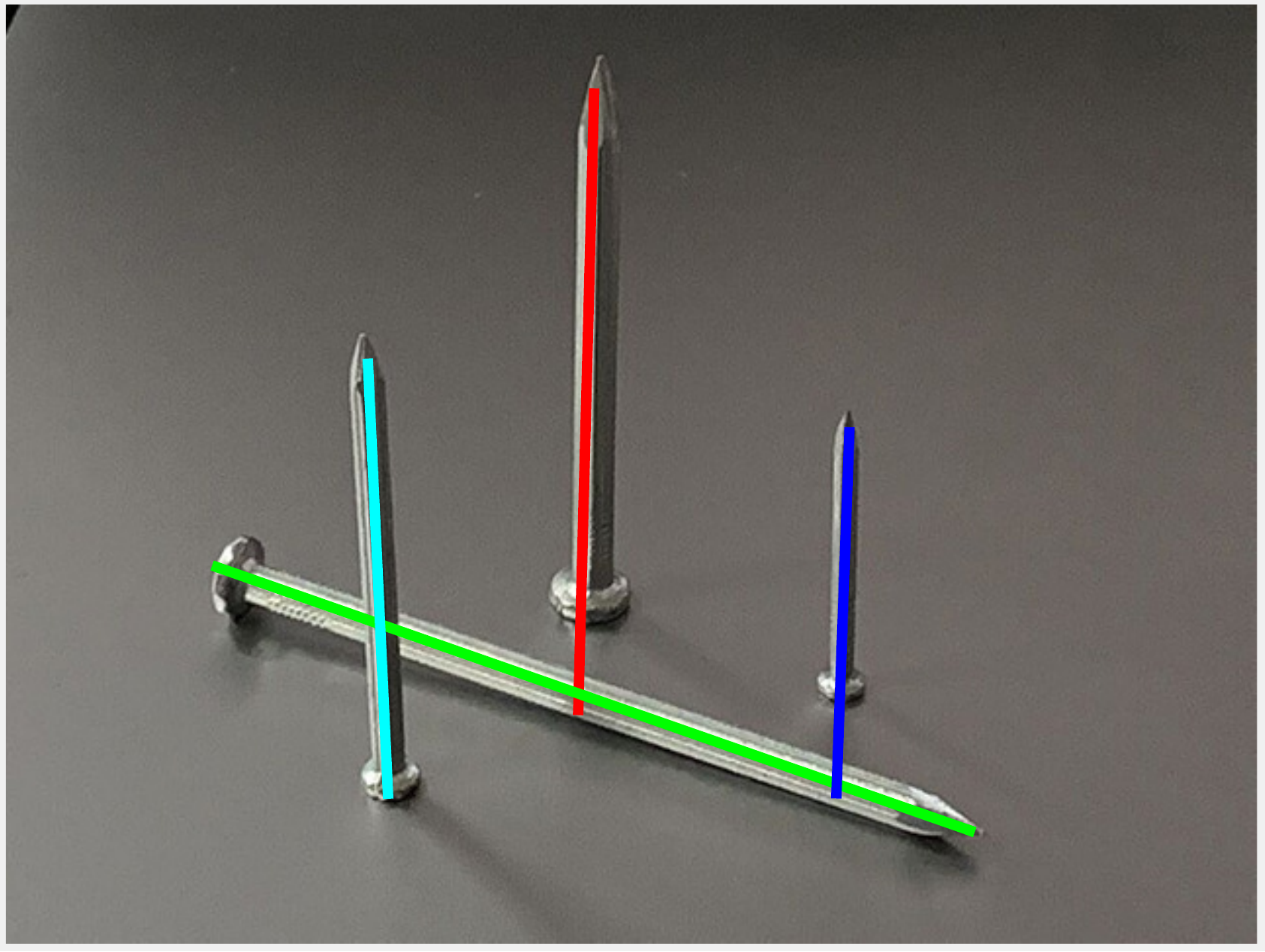}\vspace{0.2ex}
					\includegraphics[width=1.\textwidth]{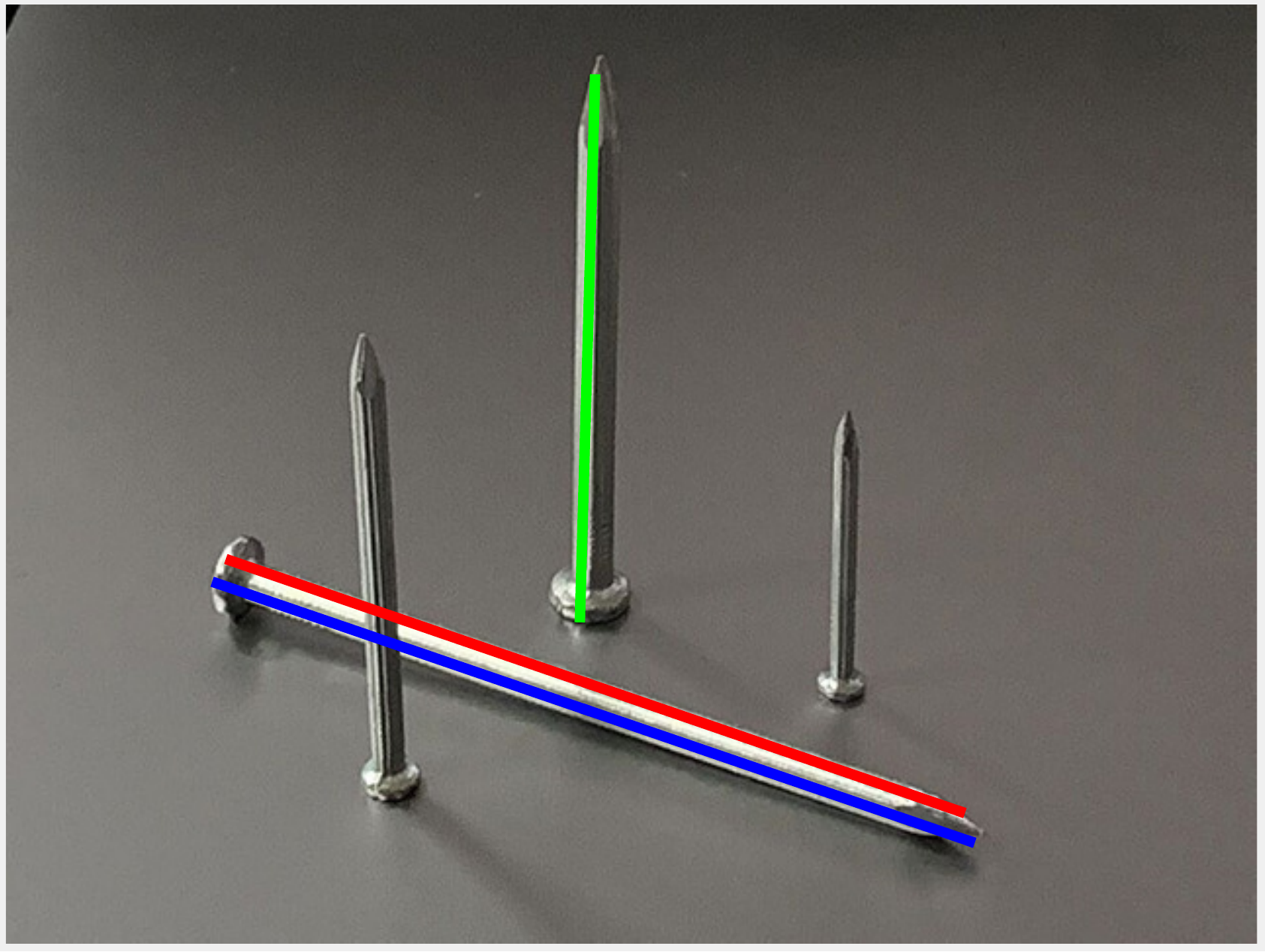}\vspace{0.2ex}
					\includegraphics[width=1.\textwidth]{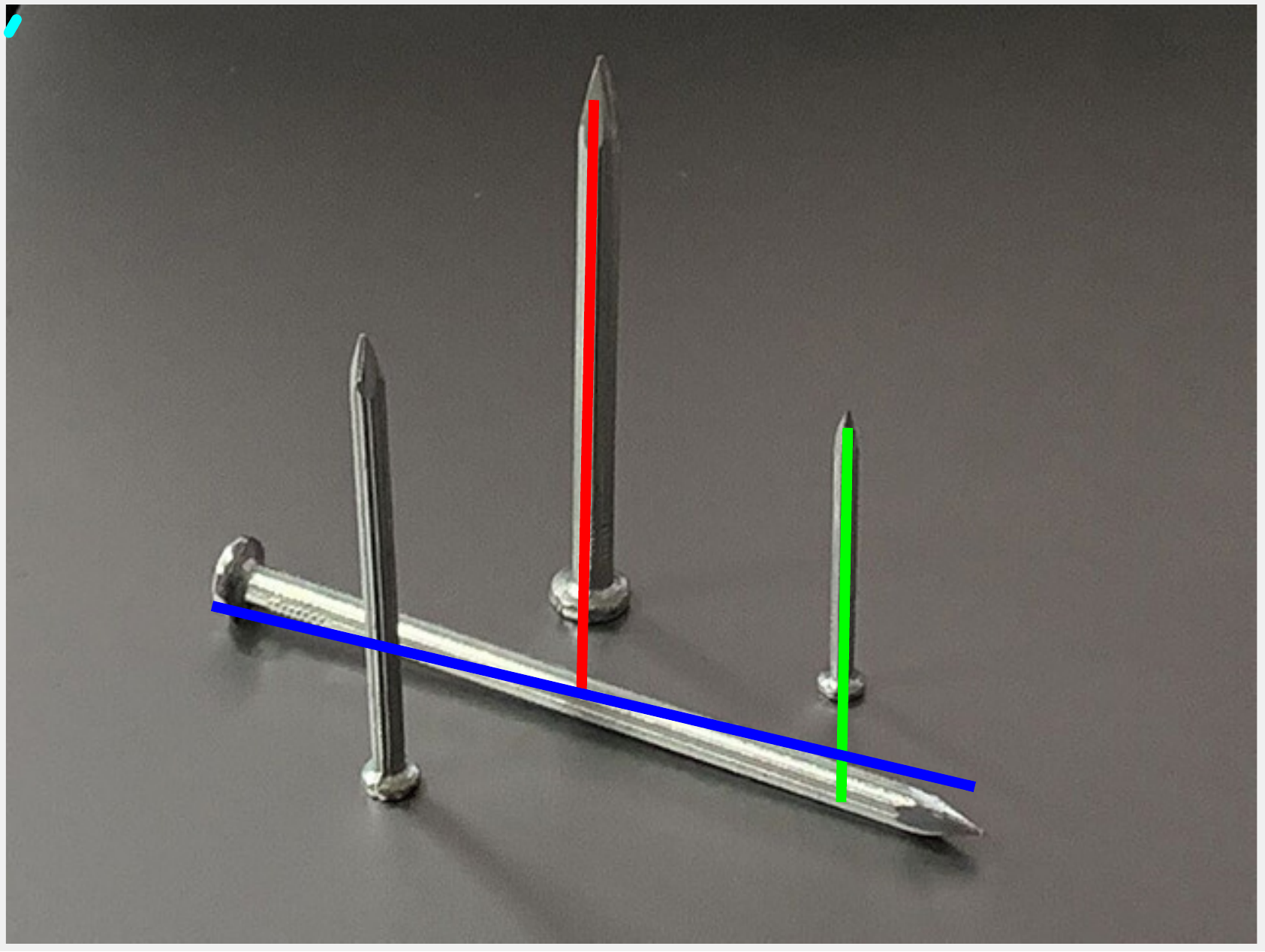}\vspace{0.2ex}
					\includegraphics[width=1.\textwidth]{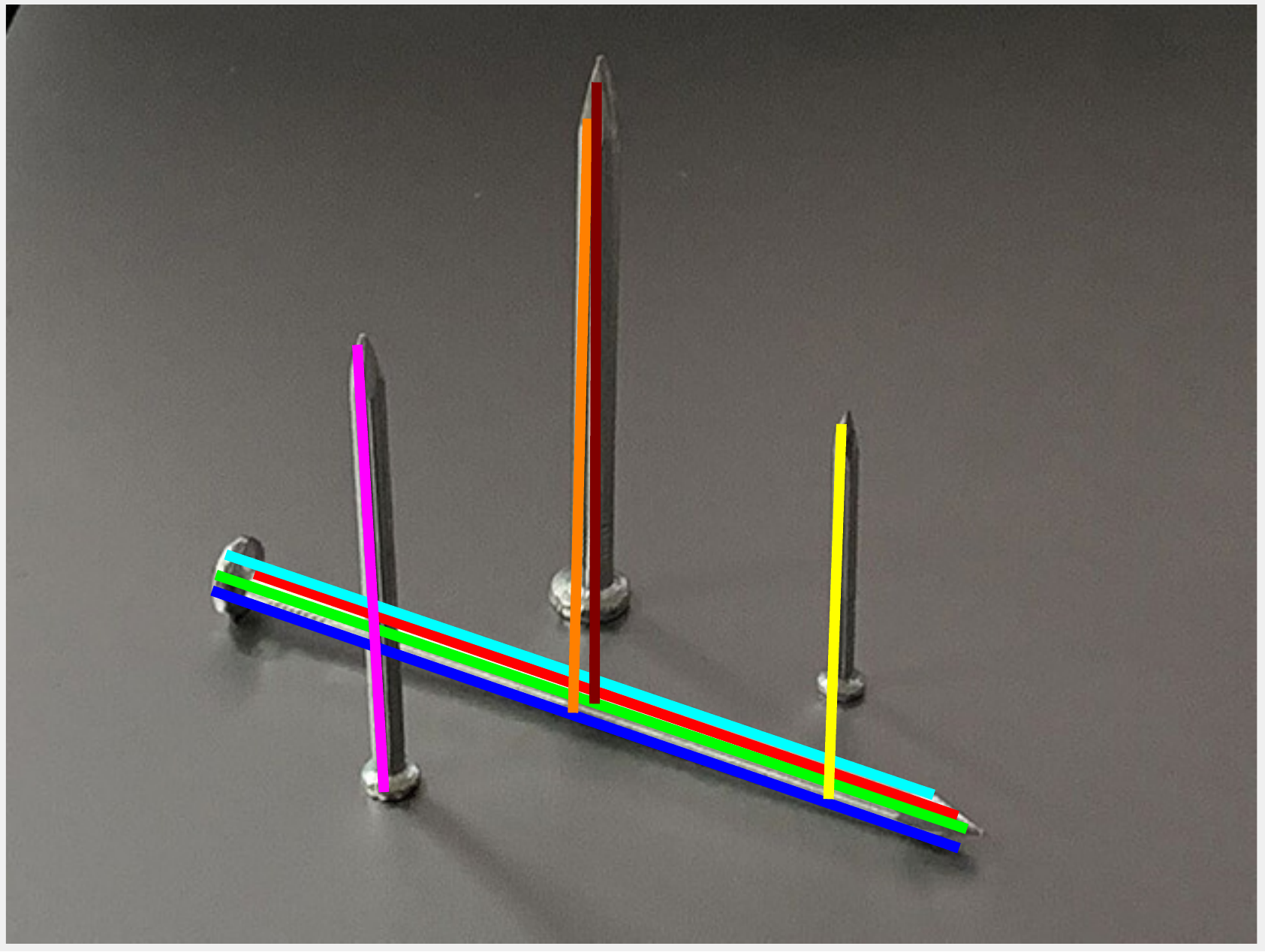}\vspace{0.2ex}
					\includegraphics[width=1.\textwidth]{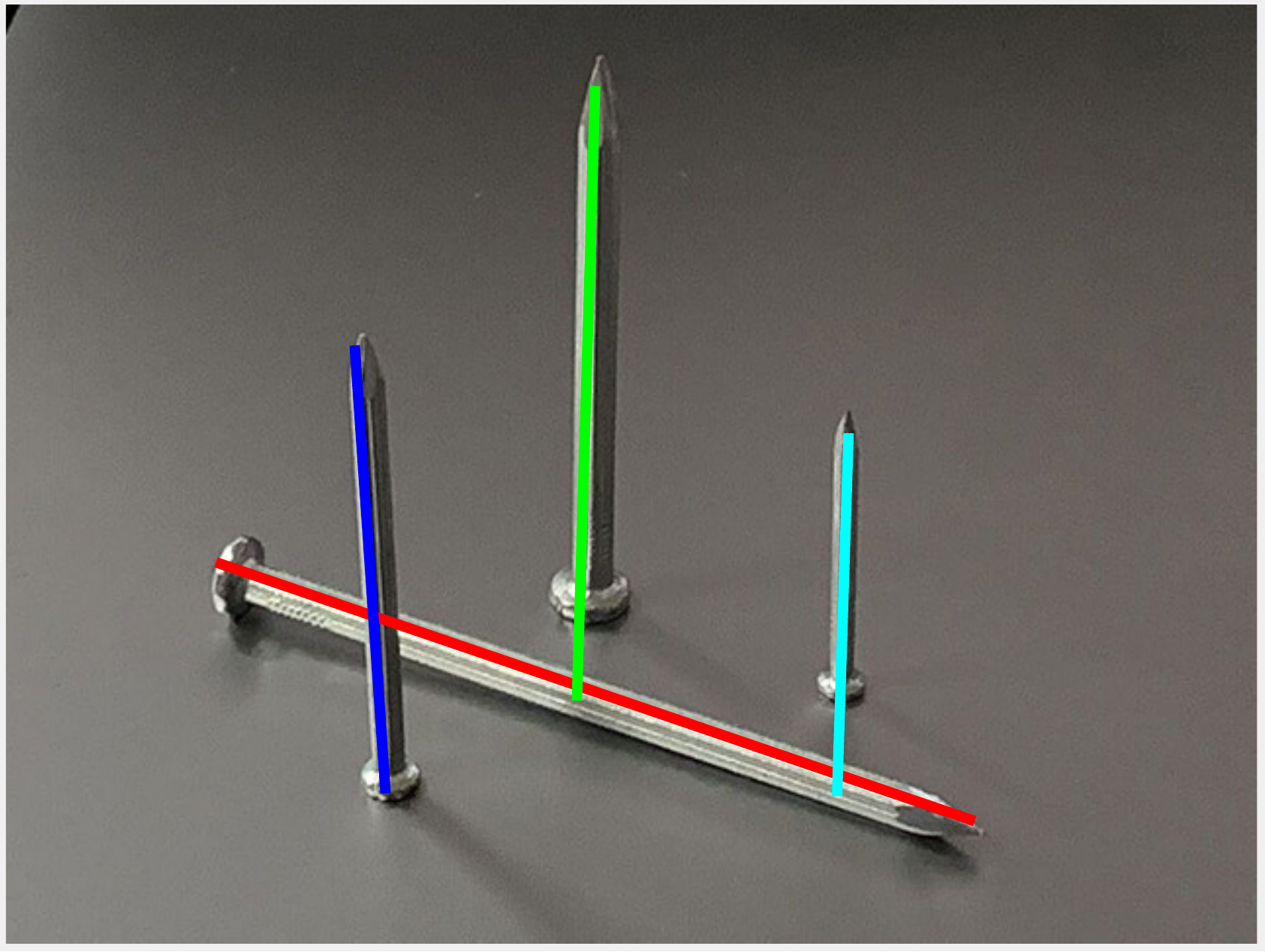}\vspace{0.2ex}
					\includegraphics[width=1.\textwidth]{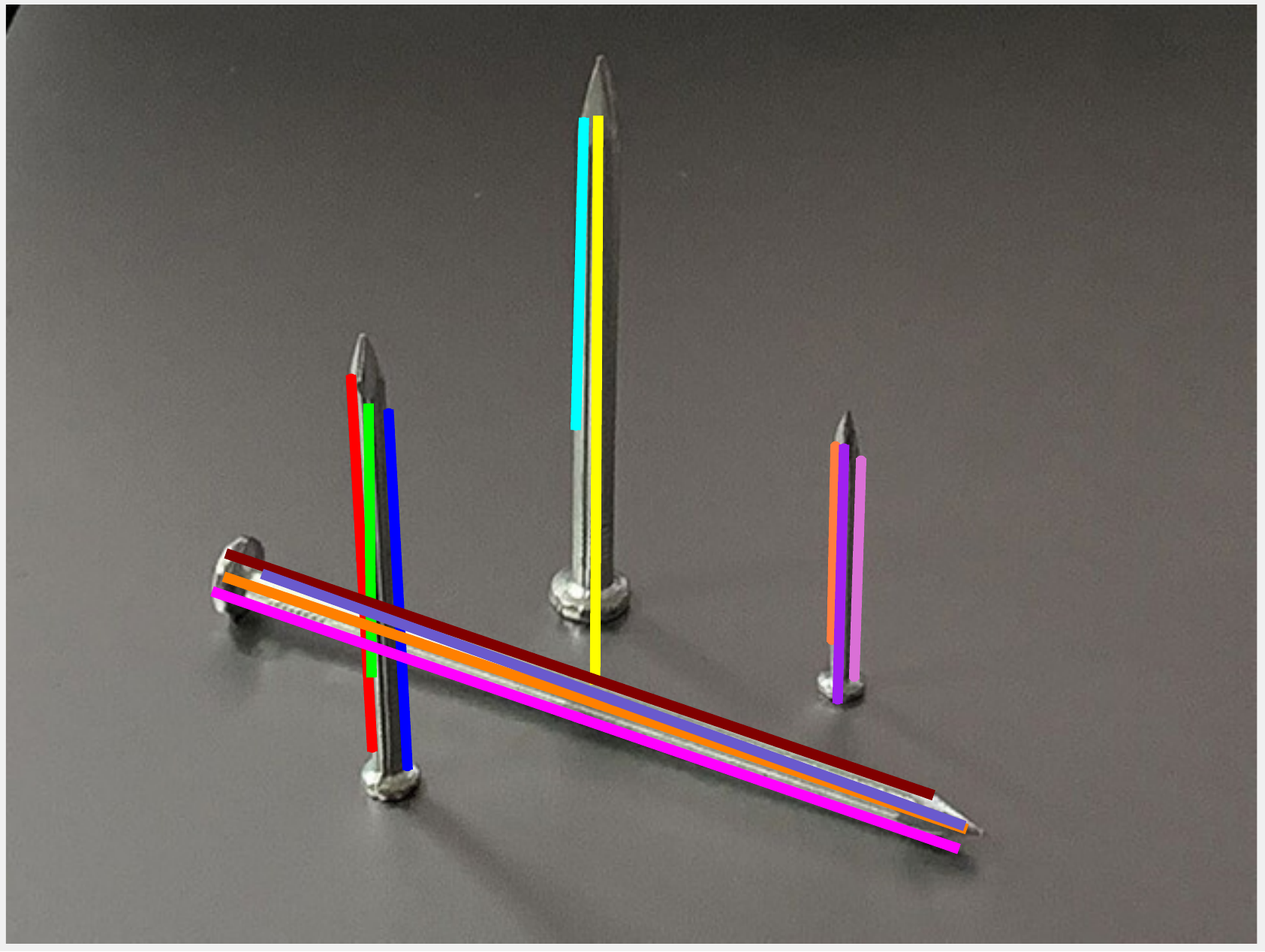}\vspace{0.2ex}
					\includegraphics[width=1.\textwidth]{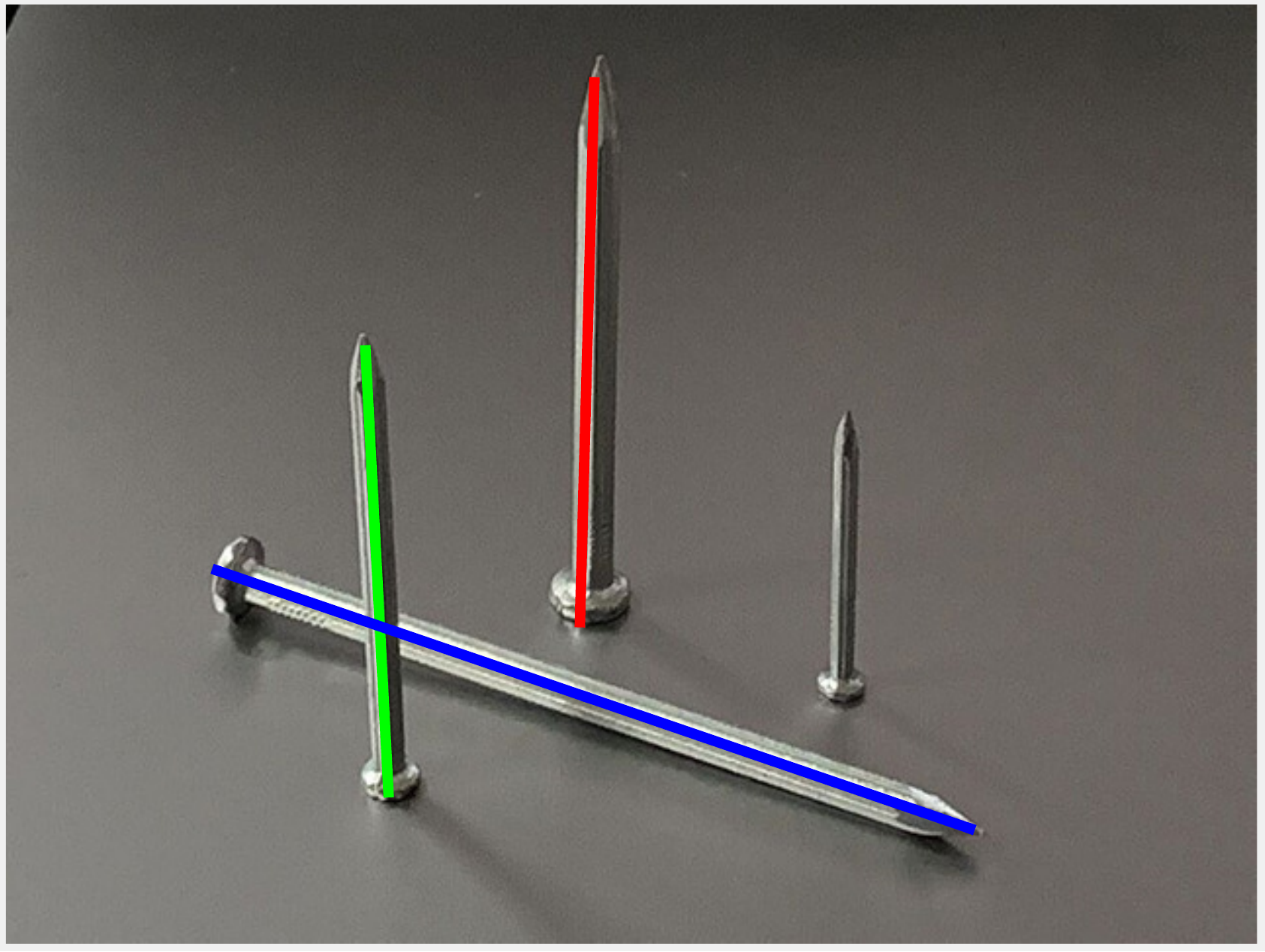}\vspace{0.2ex}
					\includegraphics[width=1.\textwidth]{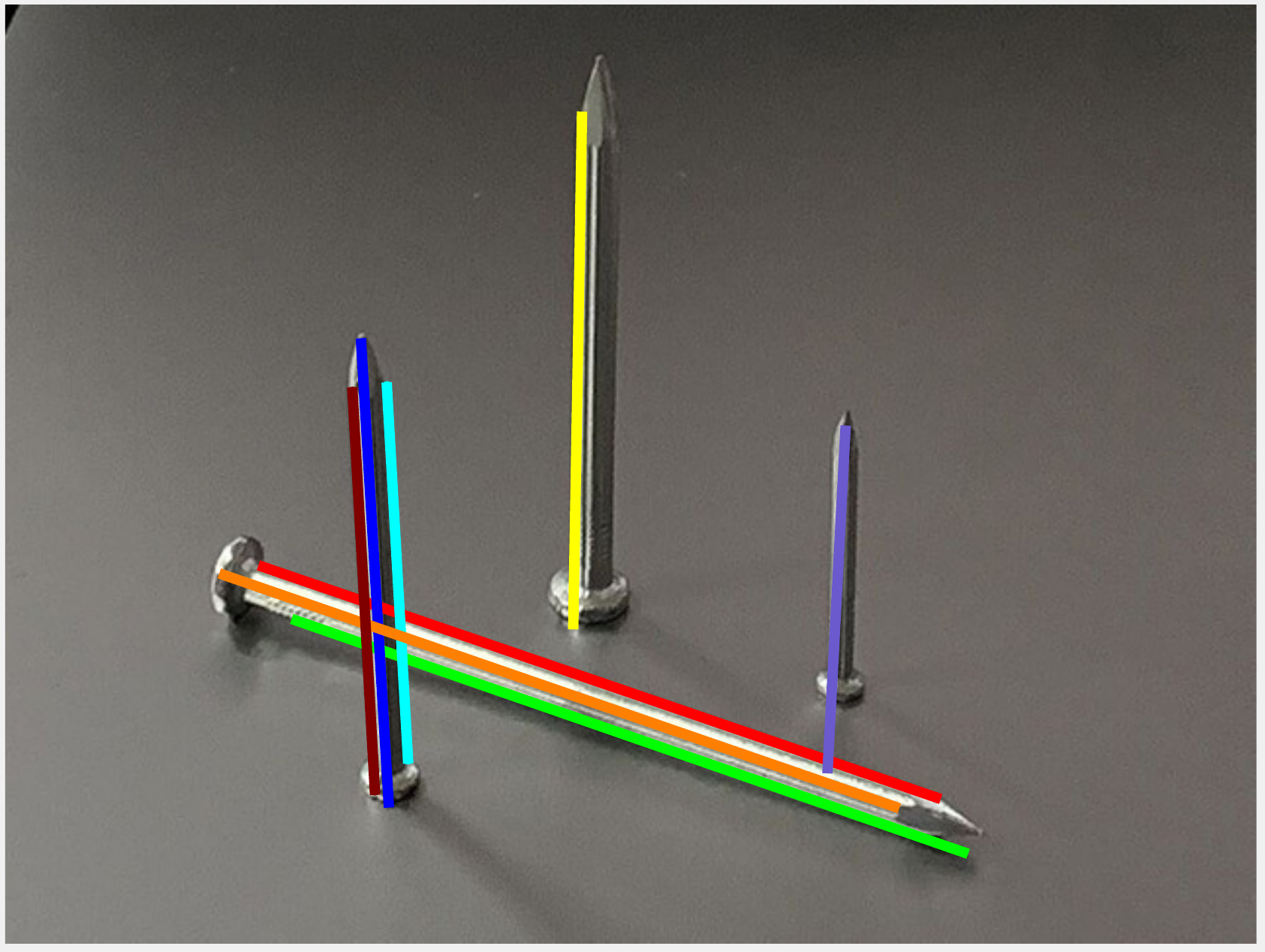}\vspace{0.2ex}
					\includegraphics[width=1.\textwidth]{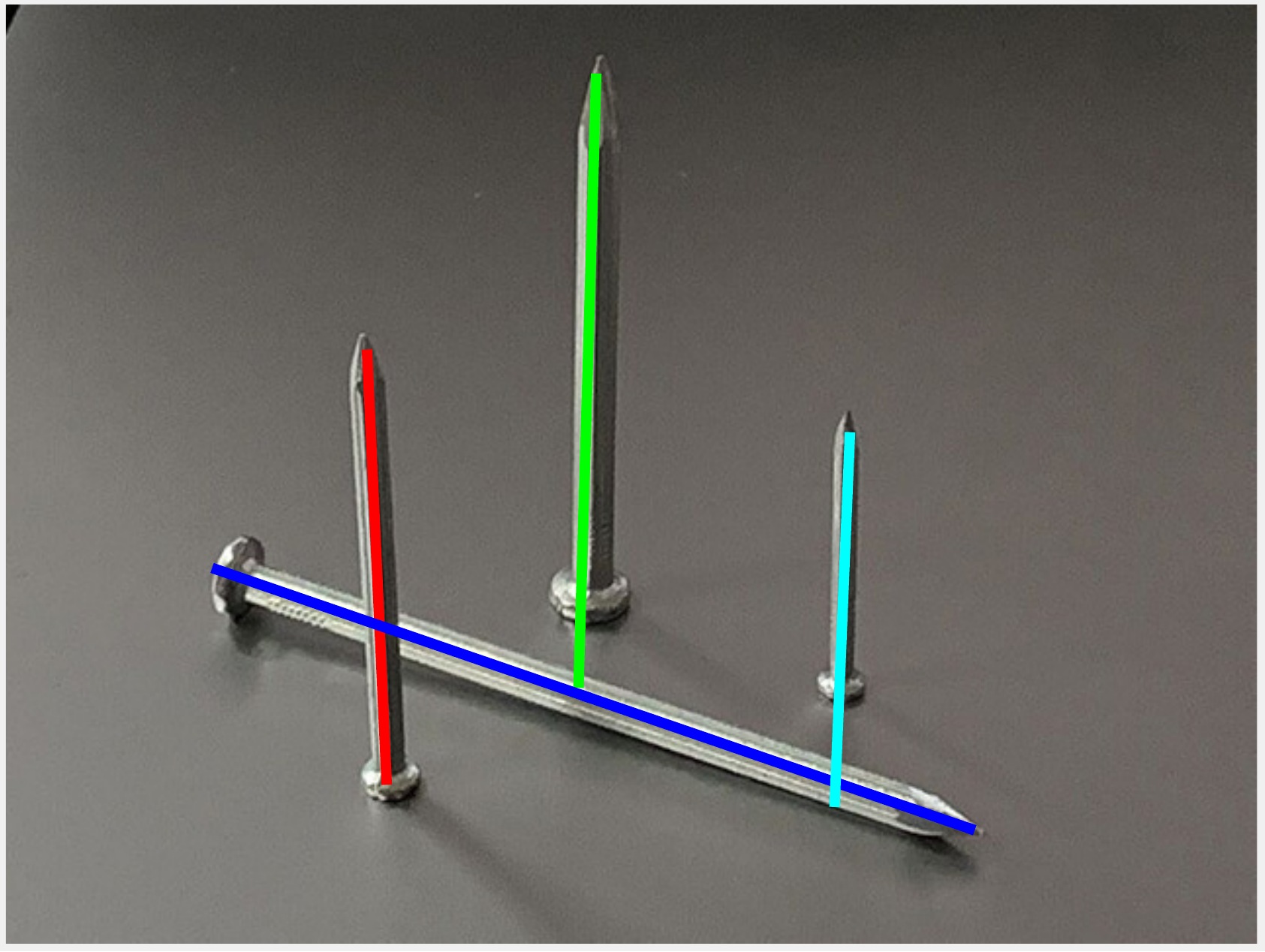}\vspace{0.2ex}
					\includegraphics[width=1.\textwidth]{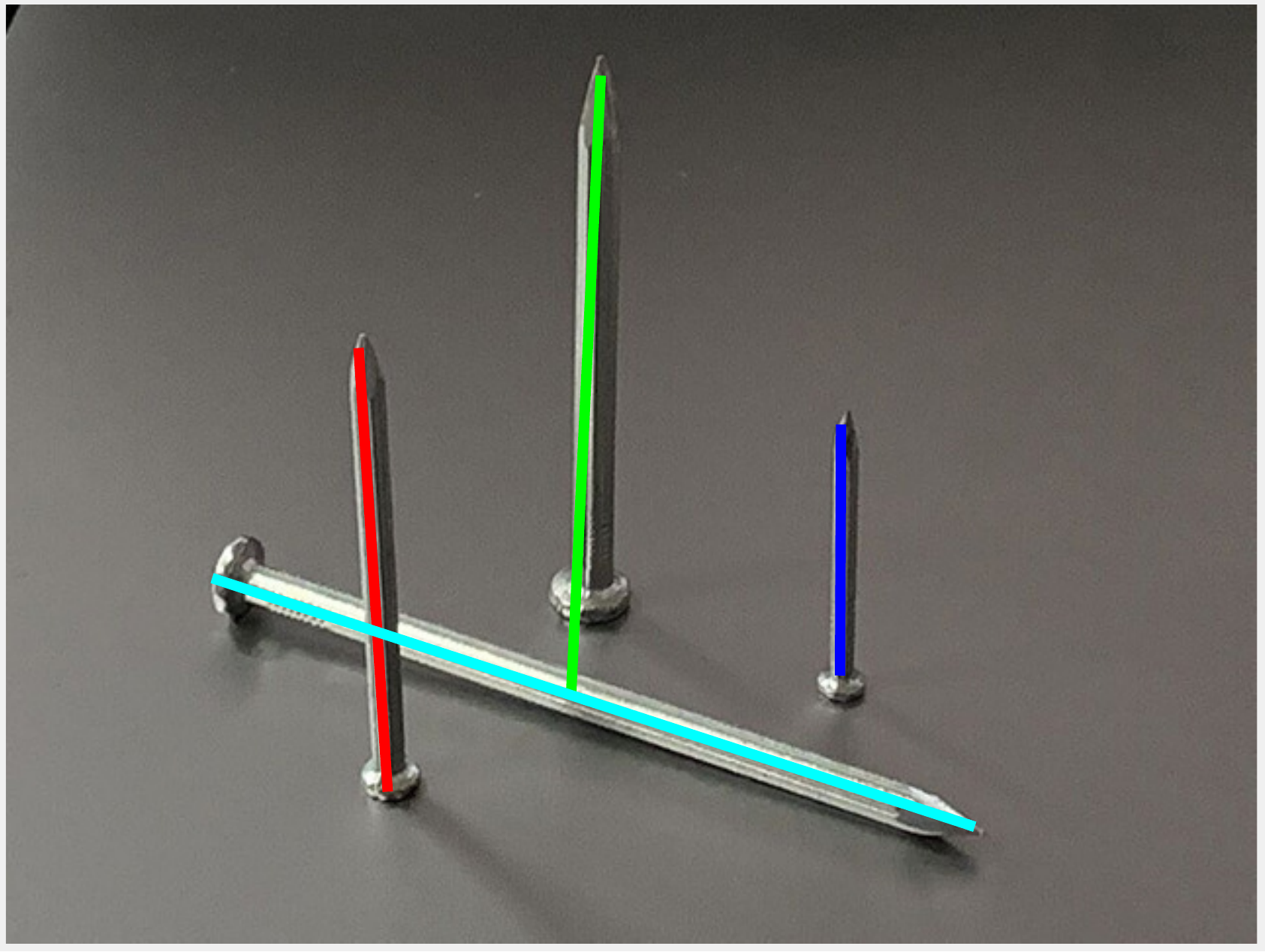}	
			\end{minipage}}\hspace{-1.5ex}
			\subfigure[Power lines]{
				\begin{minipage}[b]{0.125\textwidth}
					\includegraphics[width=1.\textwidth]{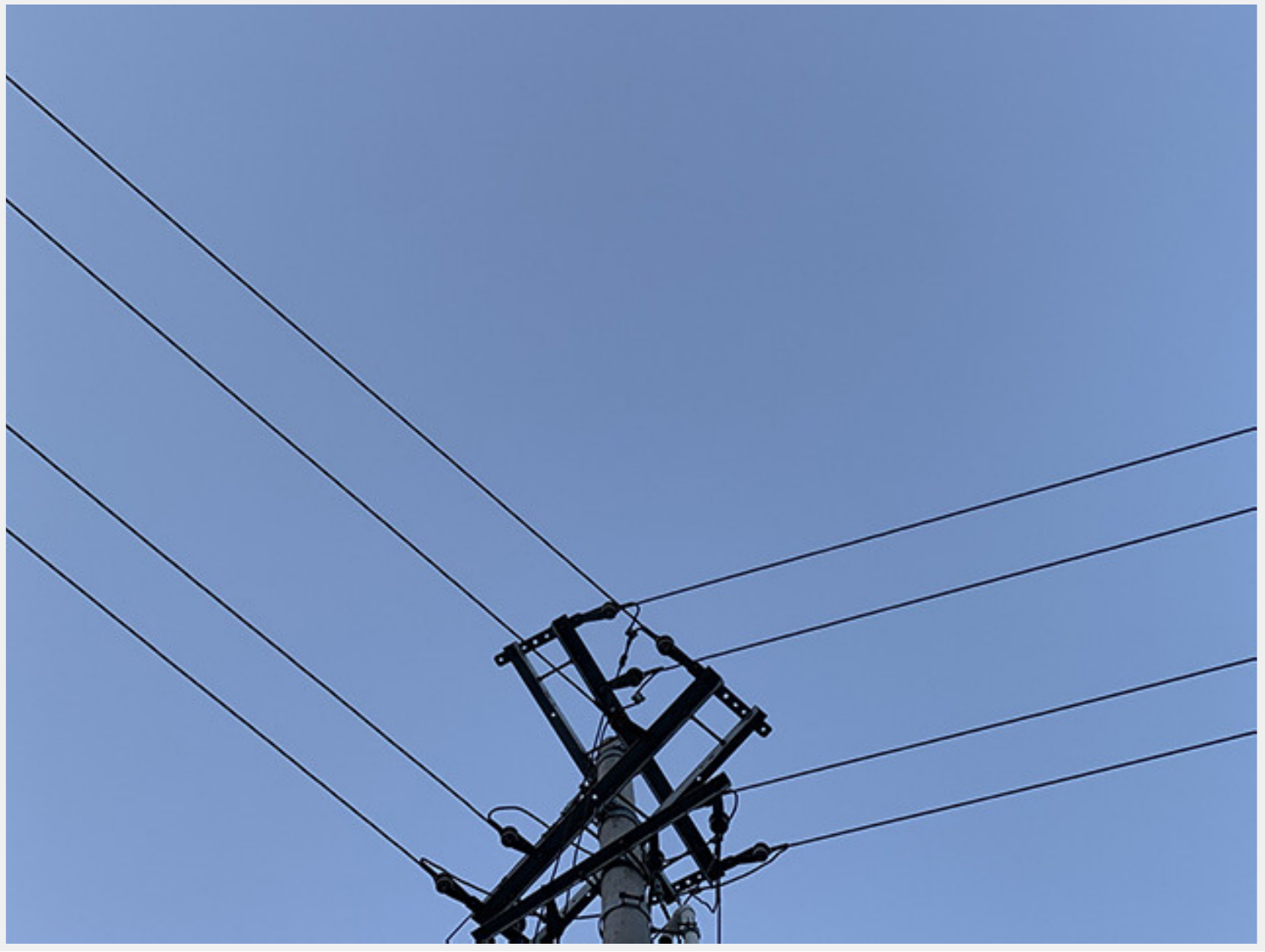}\vspace{0.2ex}
					\includegraphics[width=1.\textwidth]{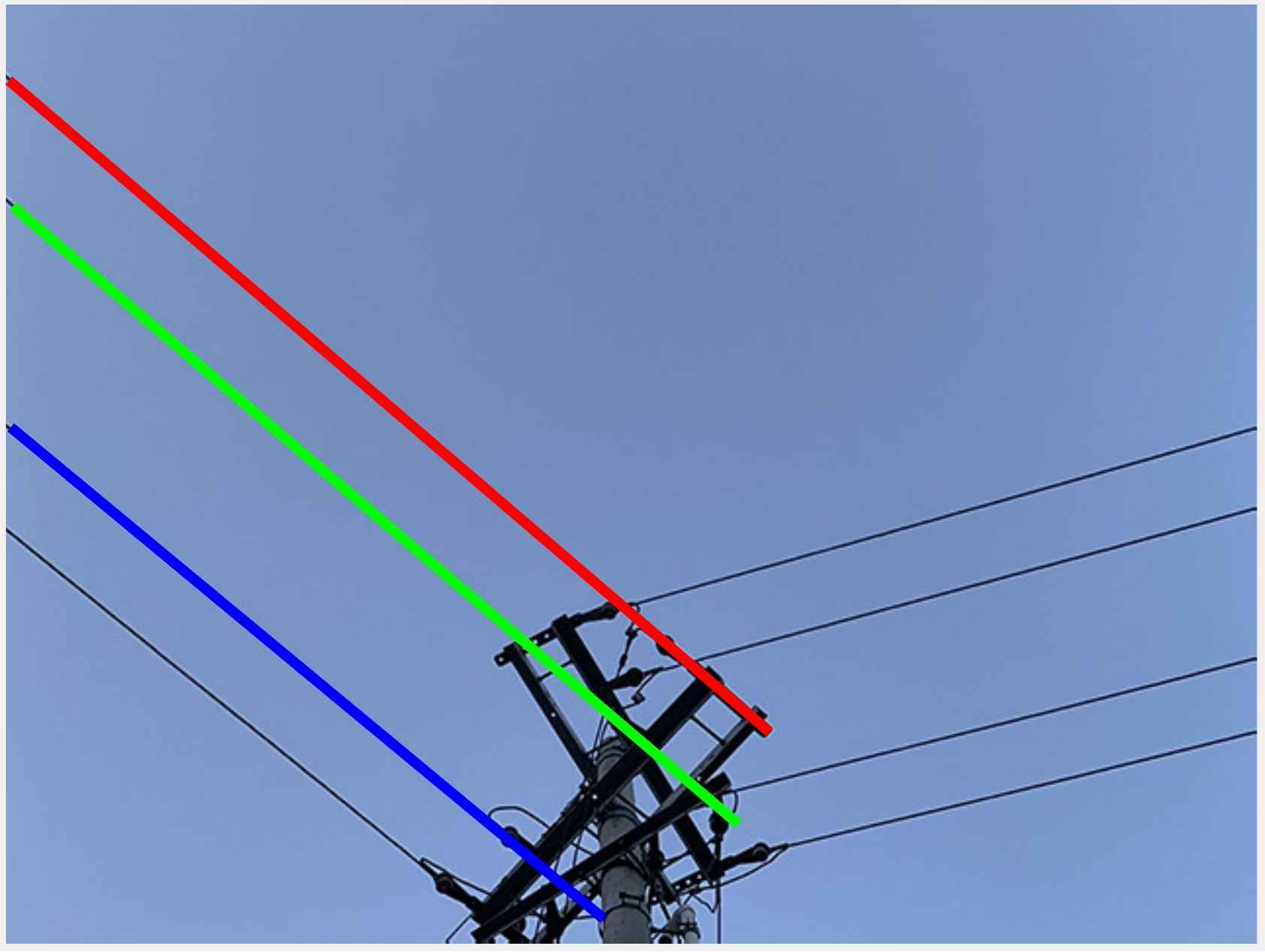}\vspace{0.2ex}
					\includegraphics[width=1.\textwidth]{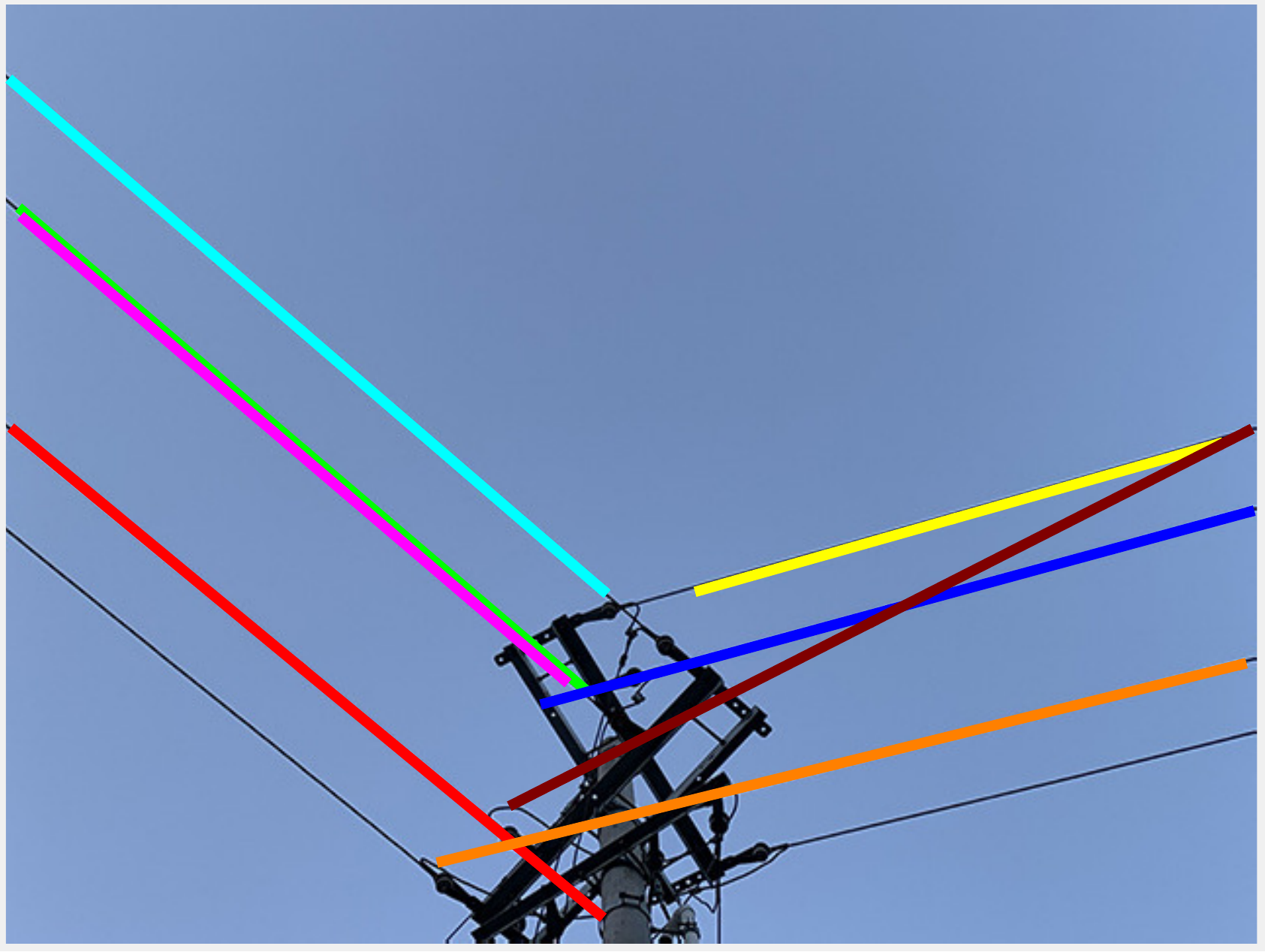}\vspace{0.2ex}
					\includegraphics[width=1.\textwidth]{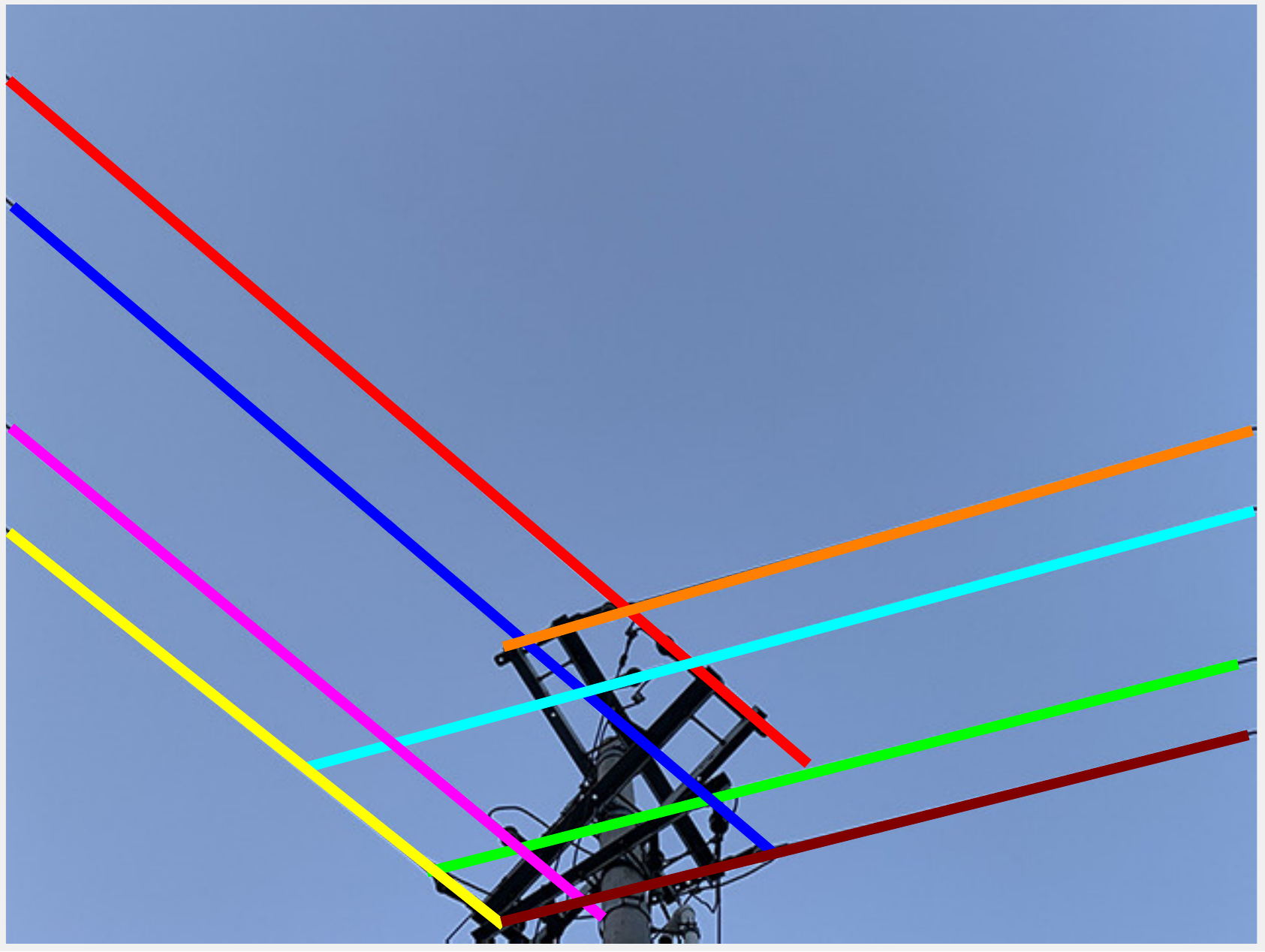}\vspace{0.2ex}
					\includegraphics[width=1.\textwidth]{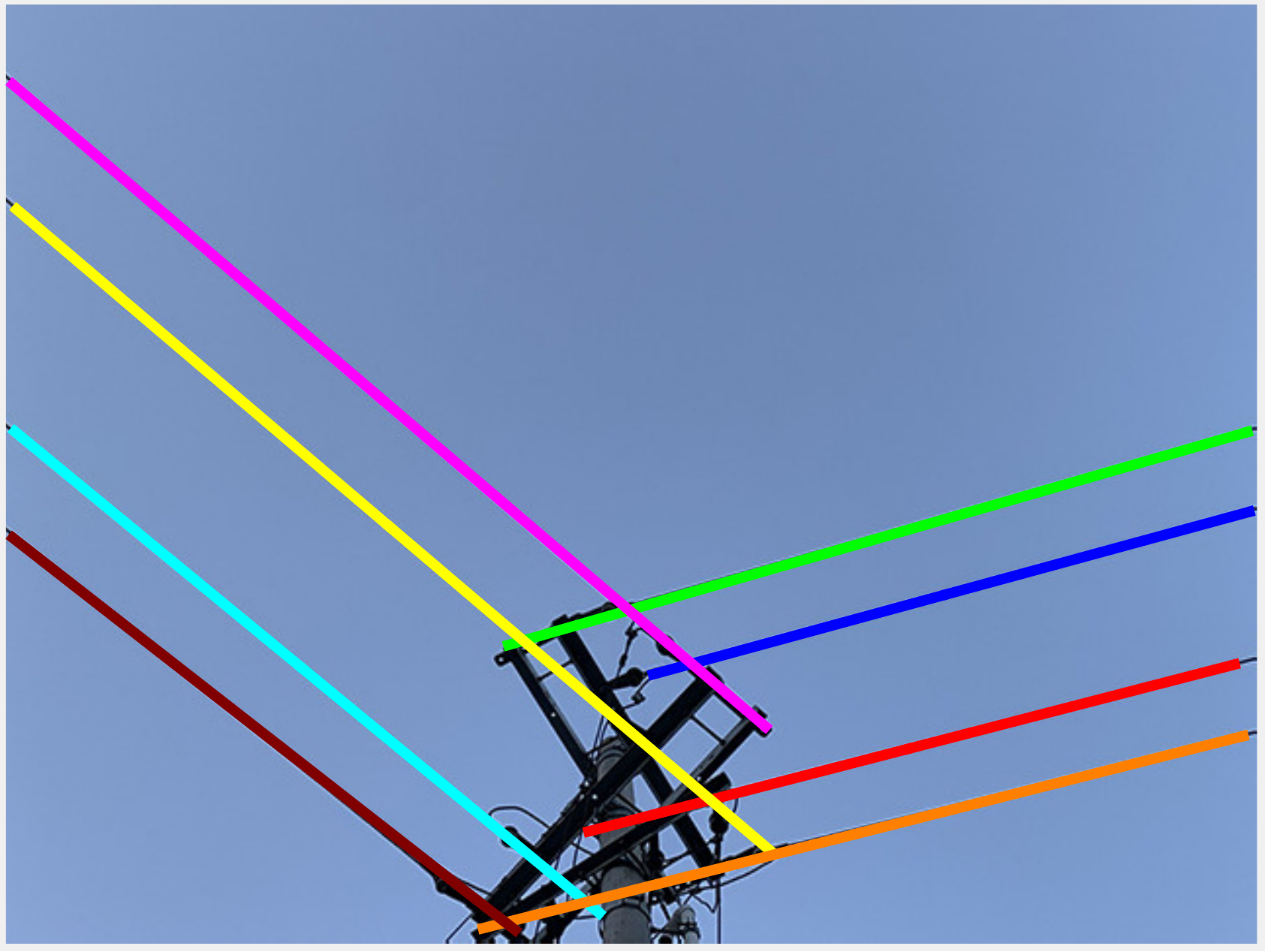}\vspace{0.2ex}
					\includegraphics[width=1.\textwidth]{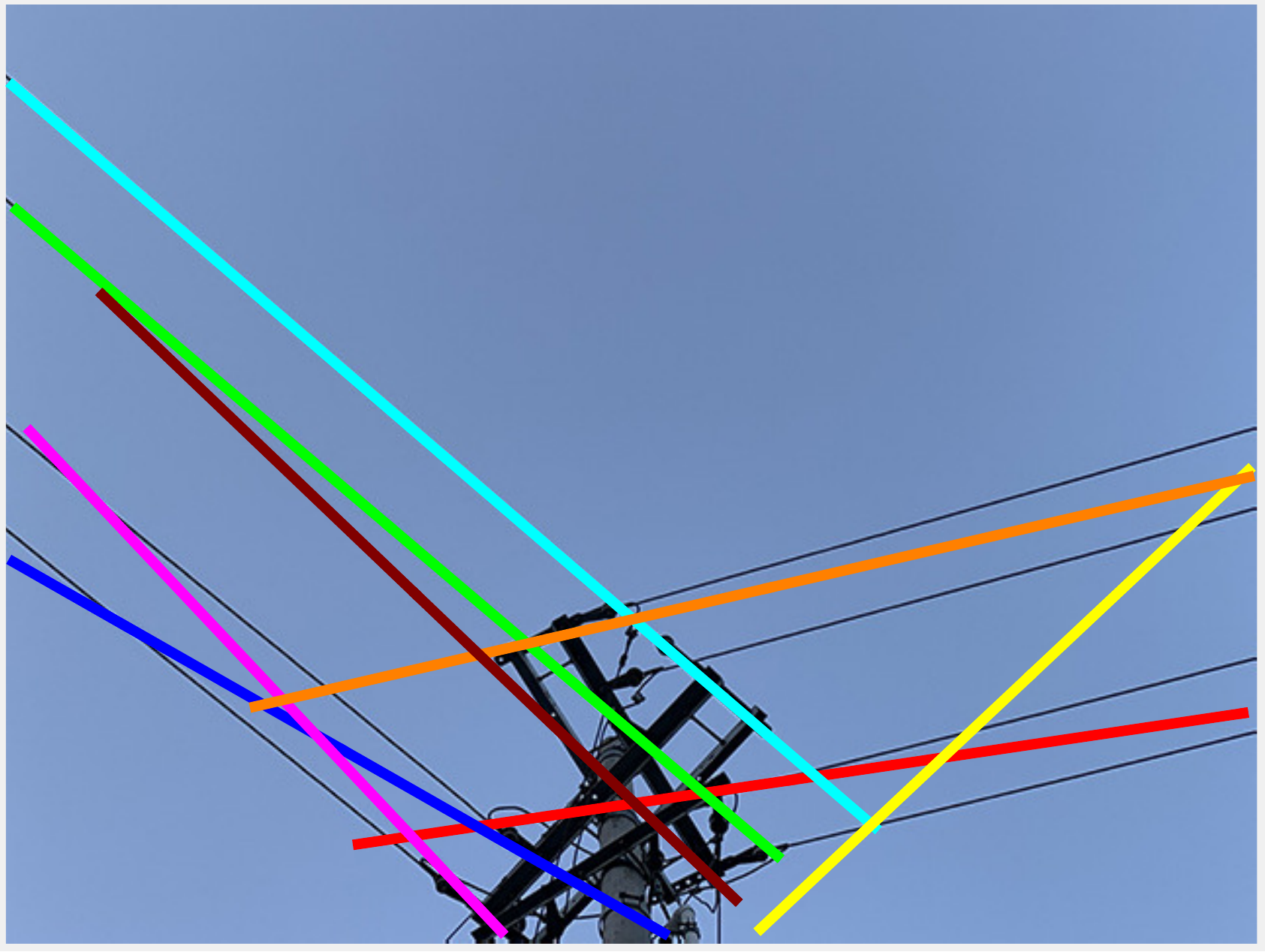}\vspace{0.2ex}
					\includegraphics[width=1.\textwidth]{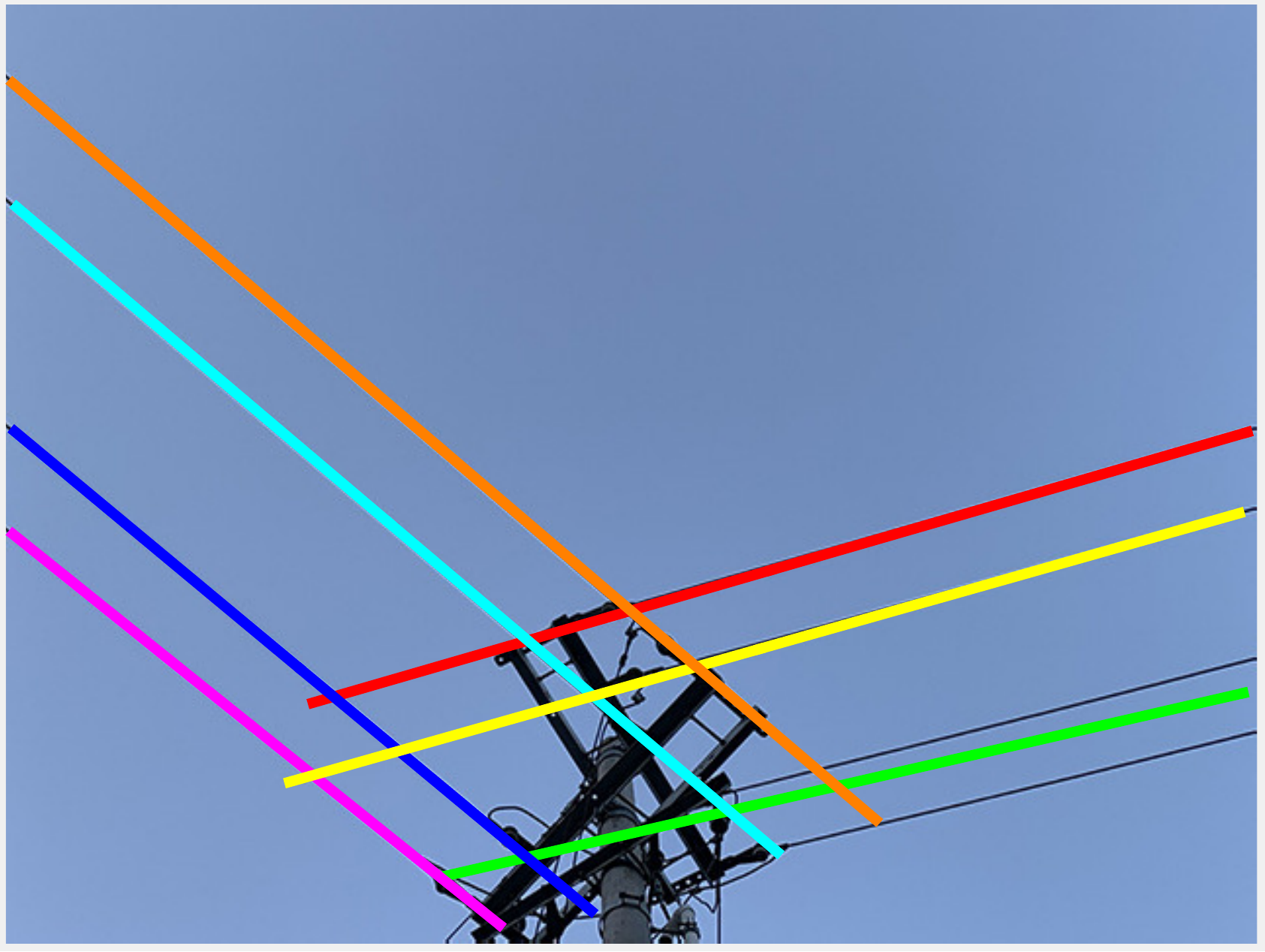}\vspace{0.2ex}
					\includegraphics[width=1.\textwidth]{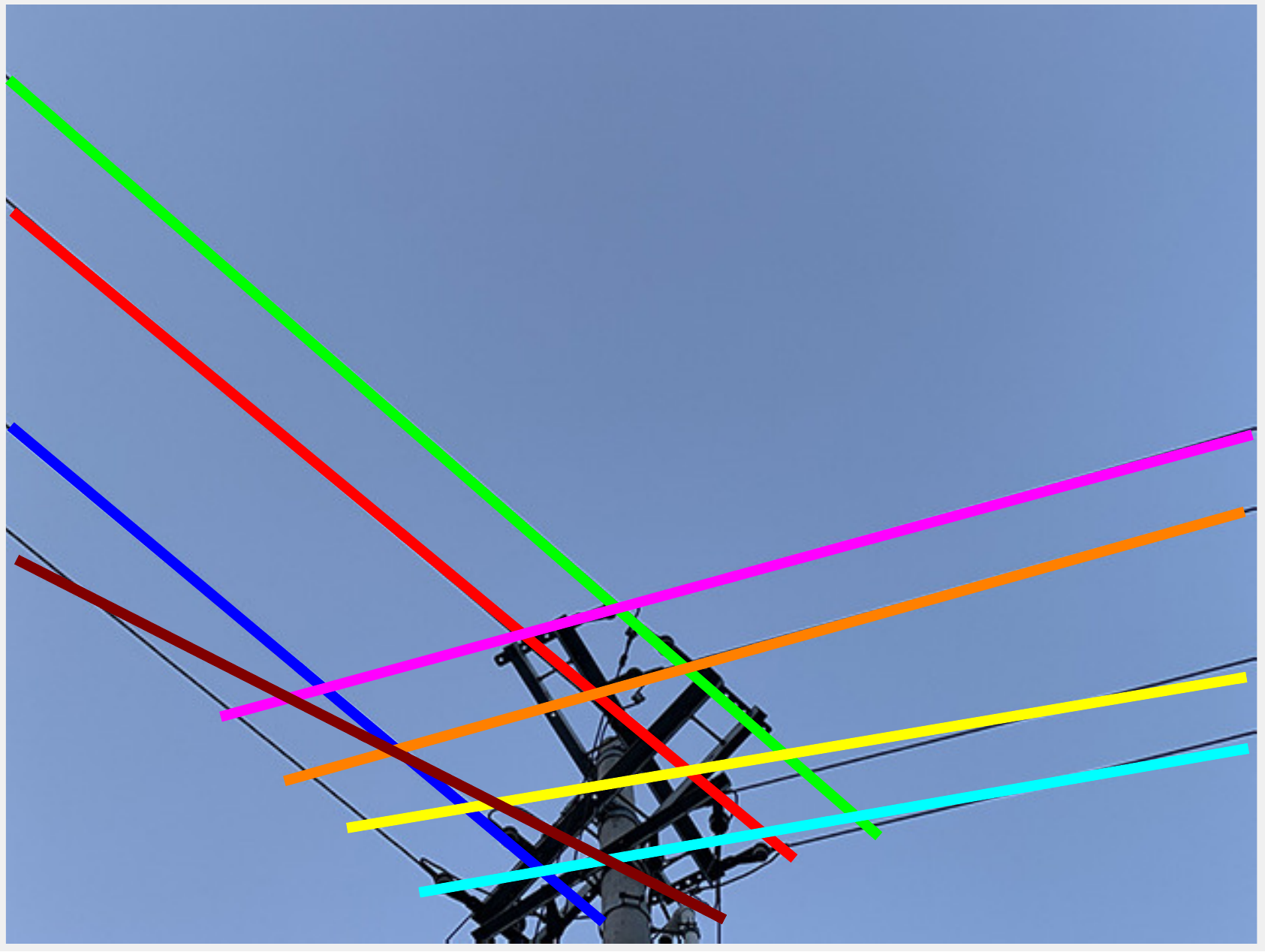}\vspace{0.2ex}
					\includegraphics[width=1.\textwidth]{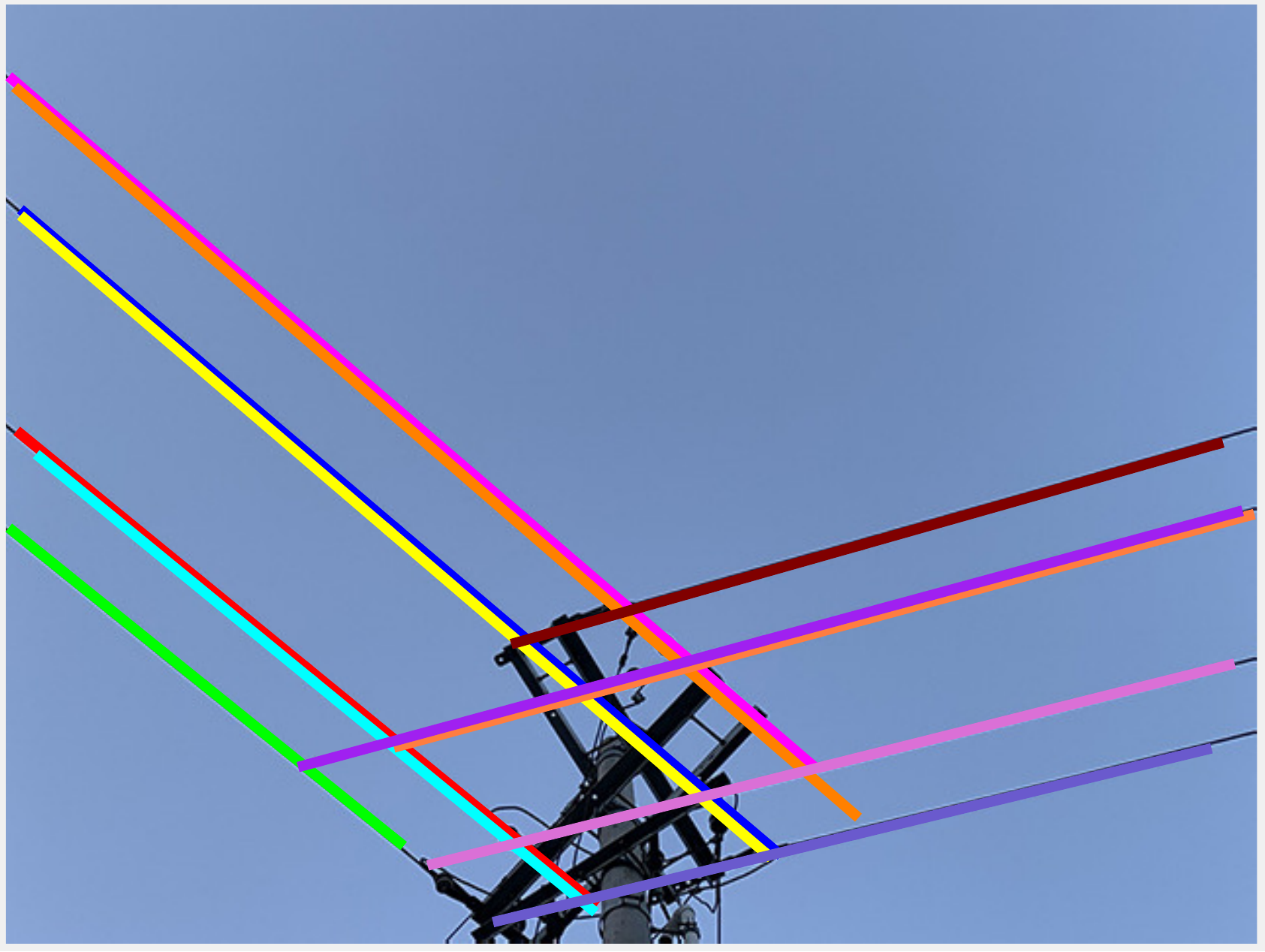}\vspace{0.2ex}
					\includegraphics[width=1.\textwidth]{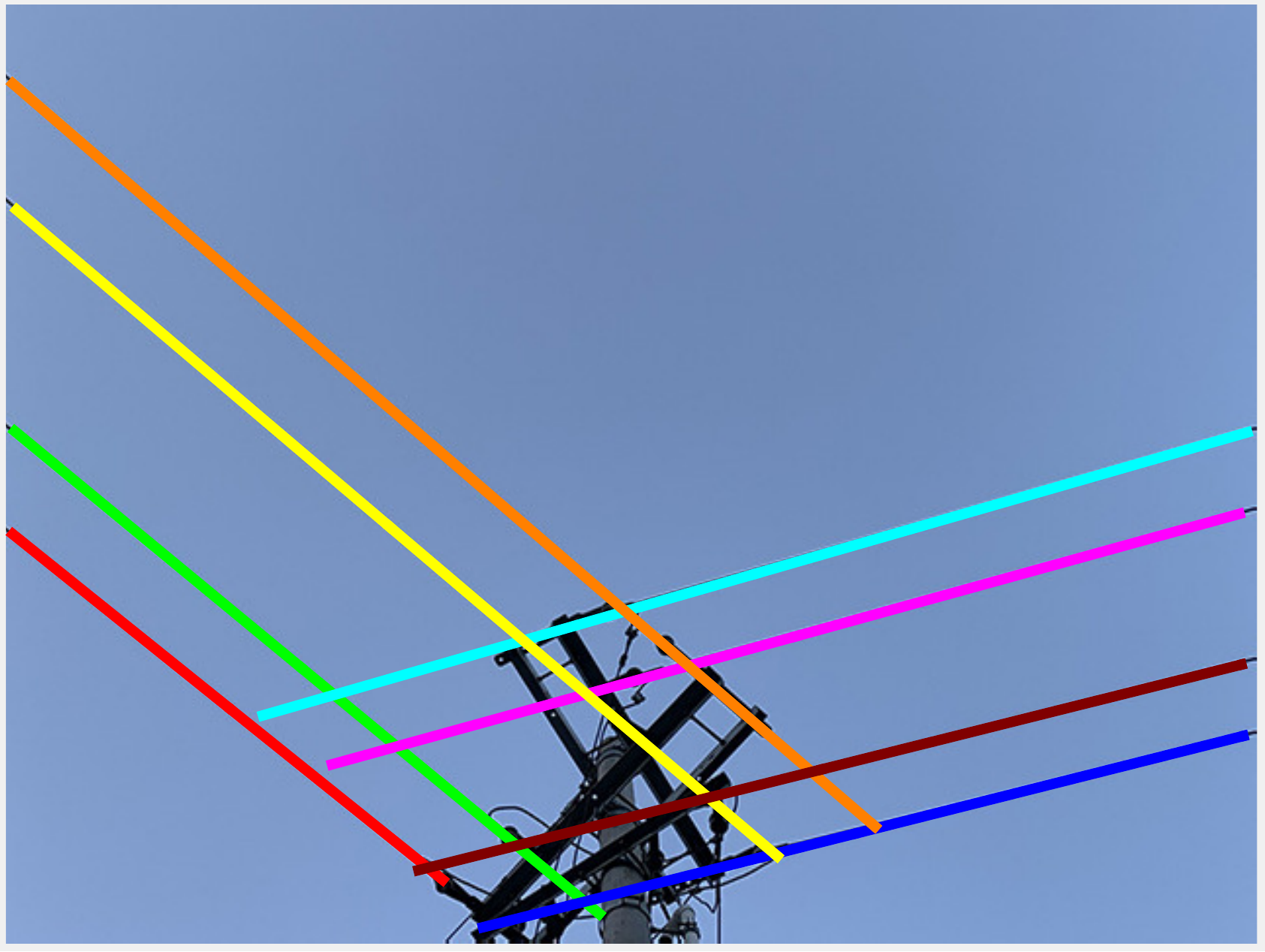}\vspace{0.2ex}
					\includegraphics[width=1.\textwidth]{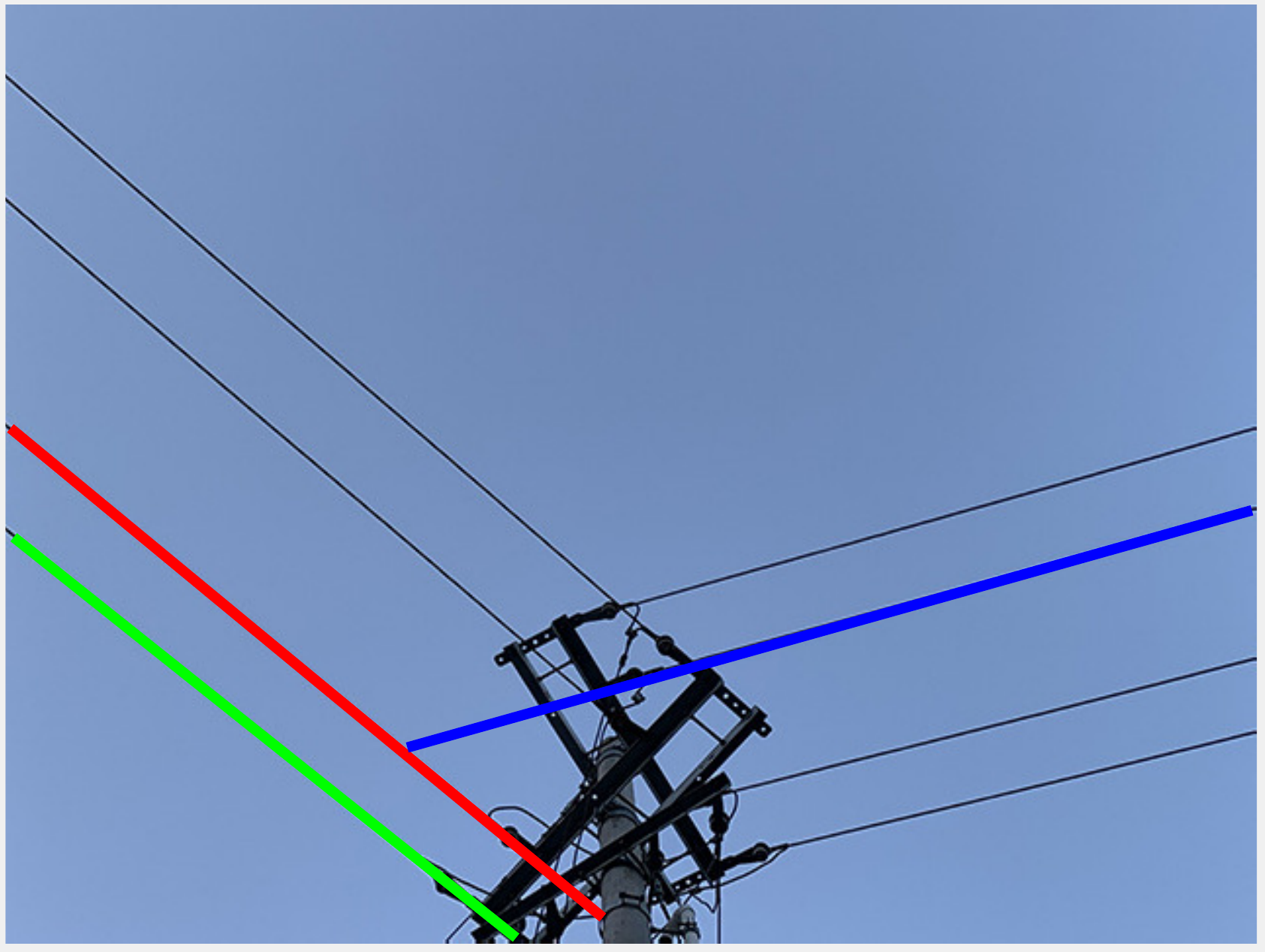}\vspace{0.2ex}
					\includegraphics[width=1.\textwidth]{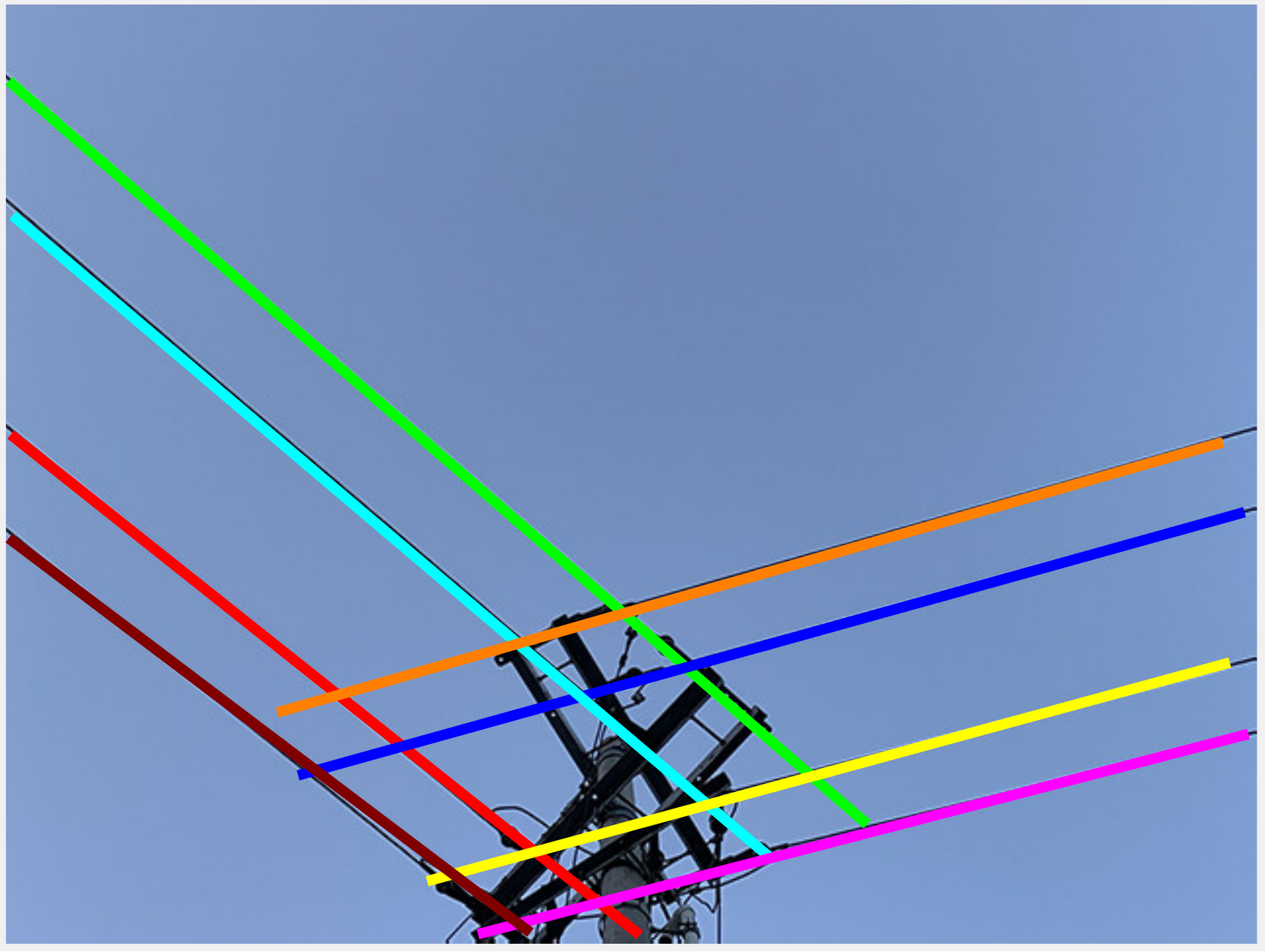}\vspace{0.2ex}
					\includegraphics[width=1.\textwidth]{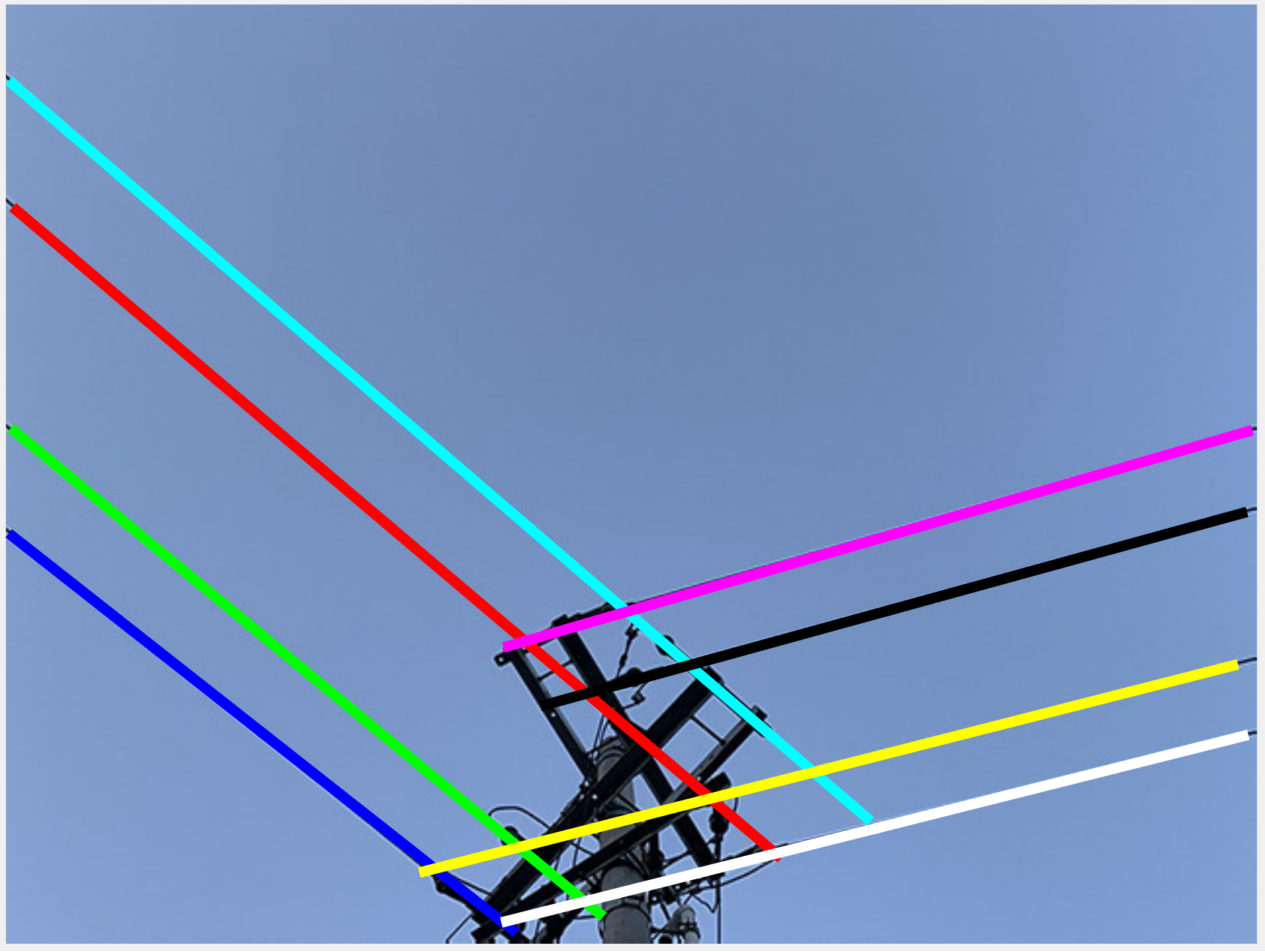}
			\end{minipage}}\hspace{-1.5ex}
			\subfigure[PVC pipes]{
				\begin{minipage}[b]{0.125\textwidth}
					\includegraphics[width=1.\textwidth]{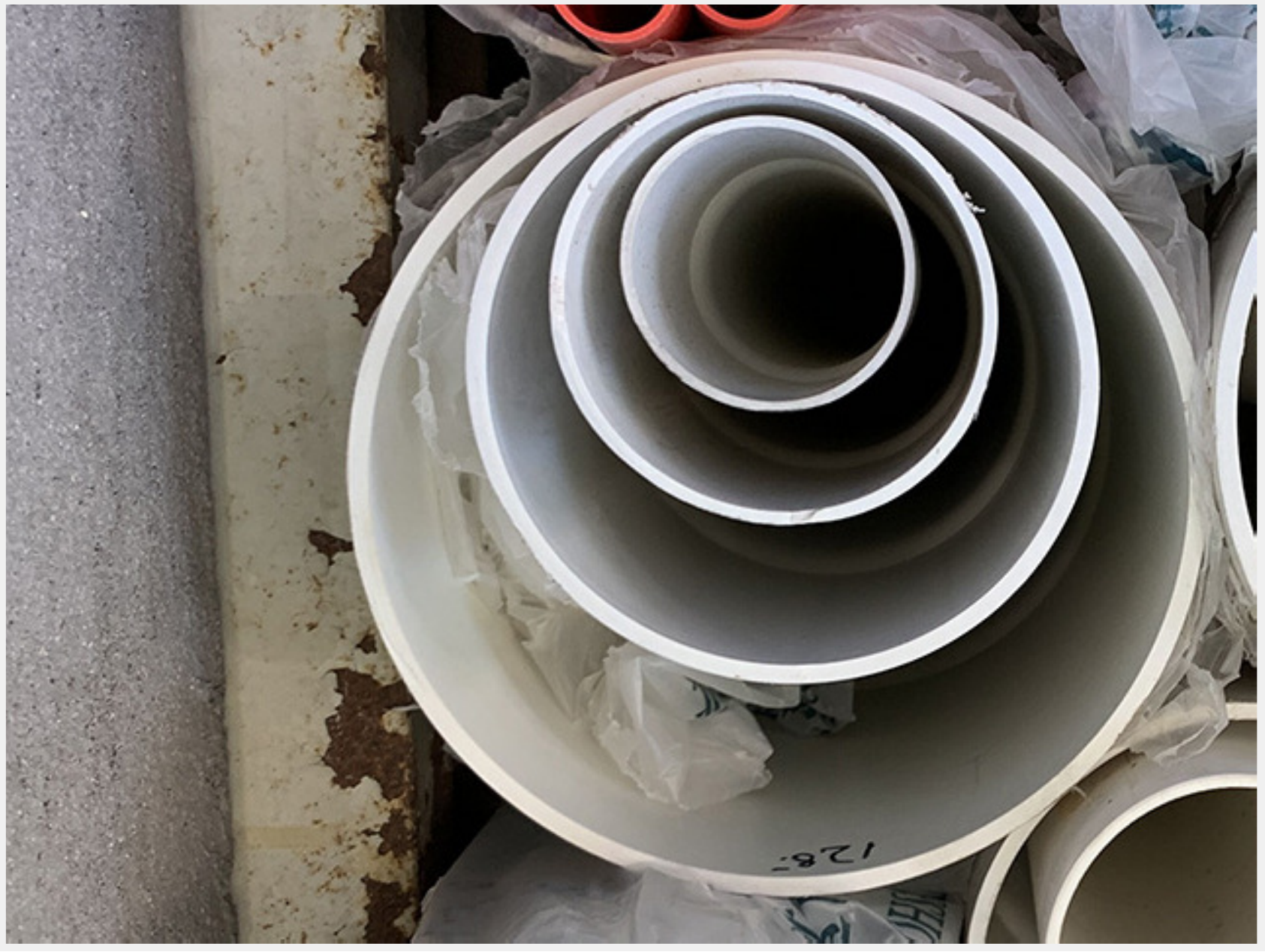}\vspace{0.2ex}
					\includegraphics[width=1.\textwidth]{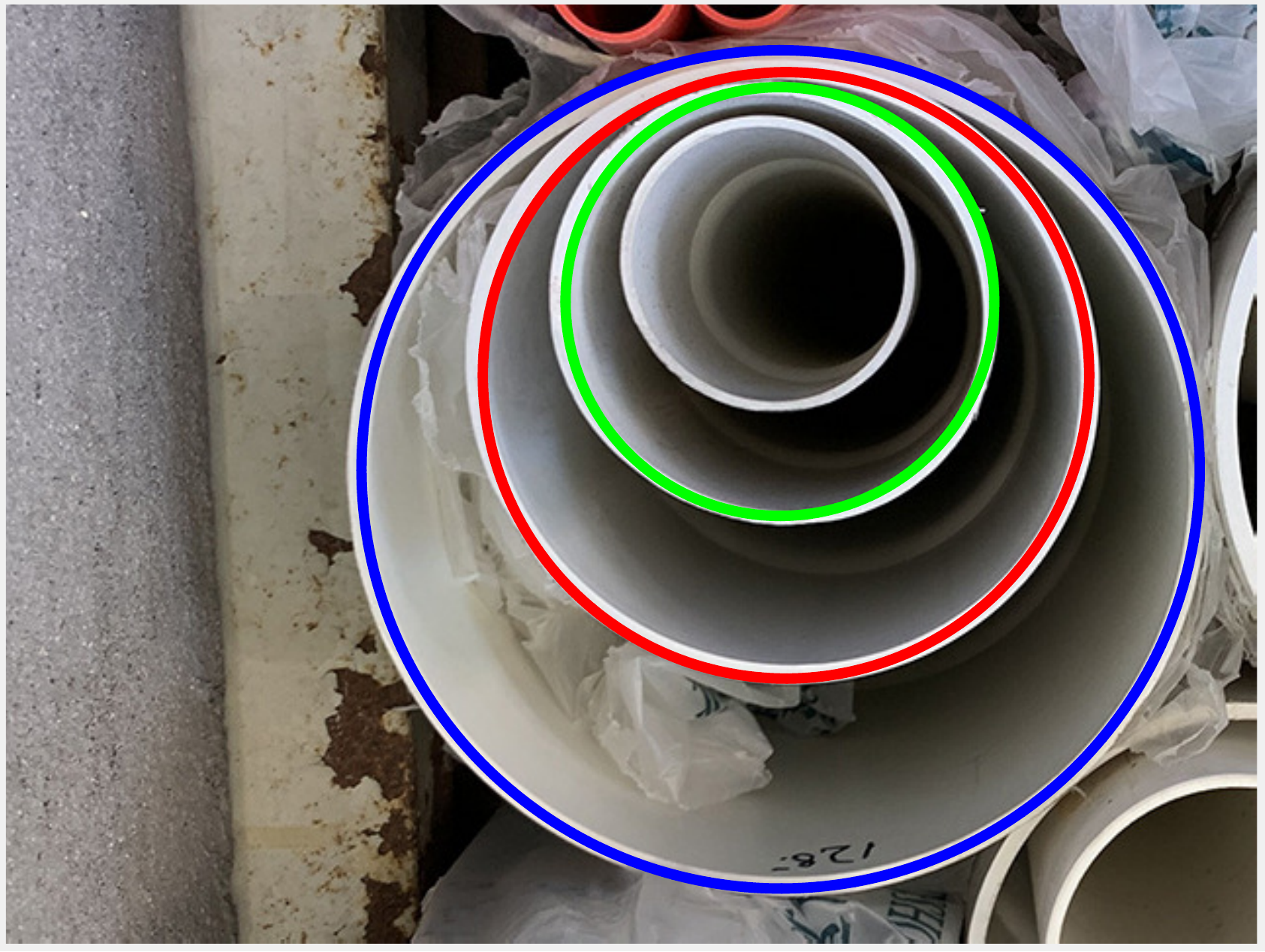}\vspace{0.2ex}
					\includegraphics[width=1.\textwidth]{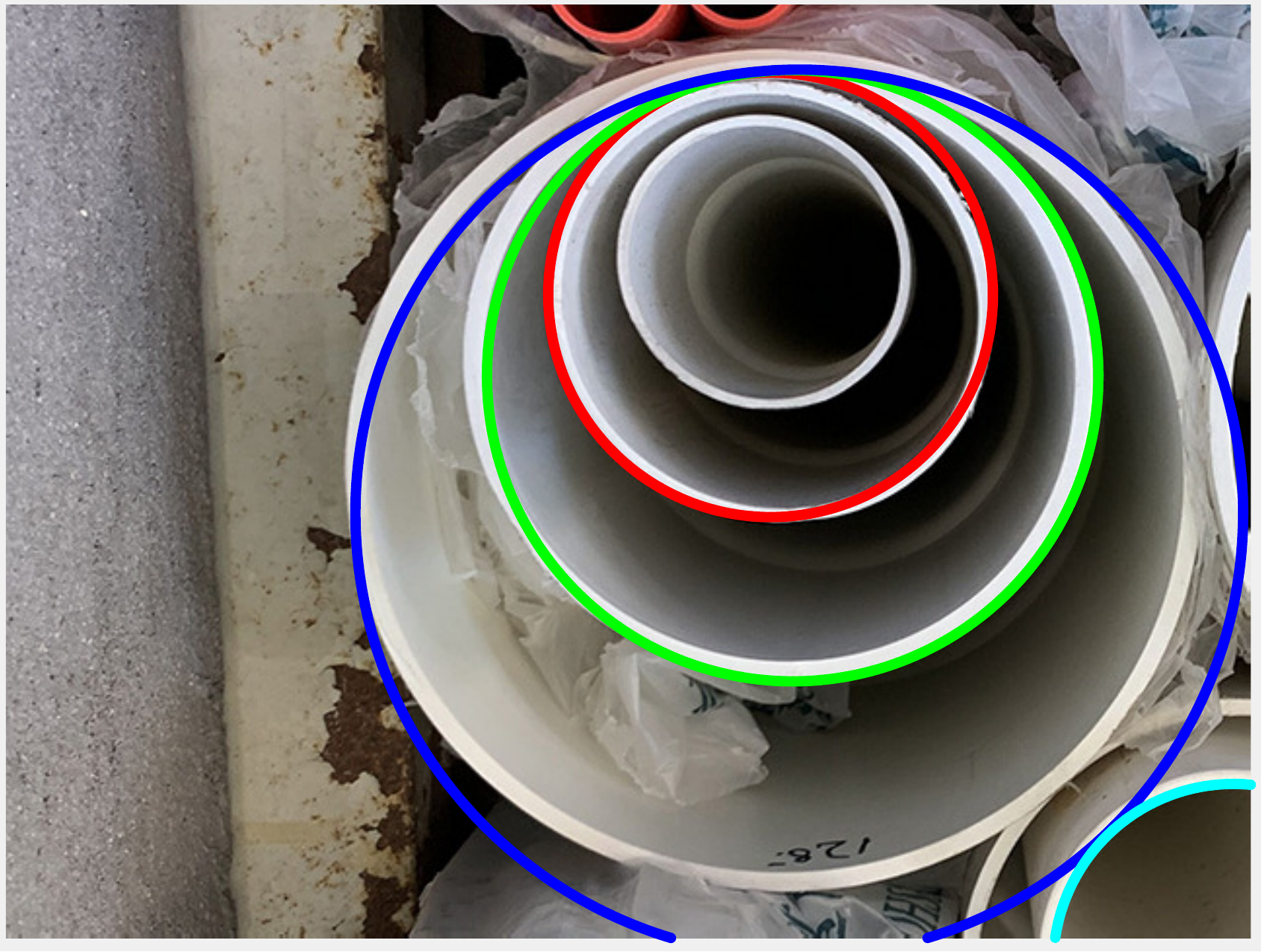}\vspace{0.2ex}
					\includegraphics[width=1.\textwidth]{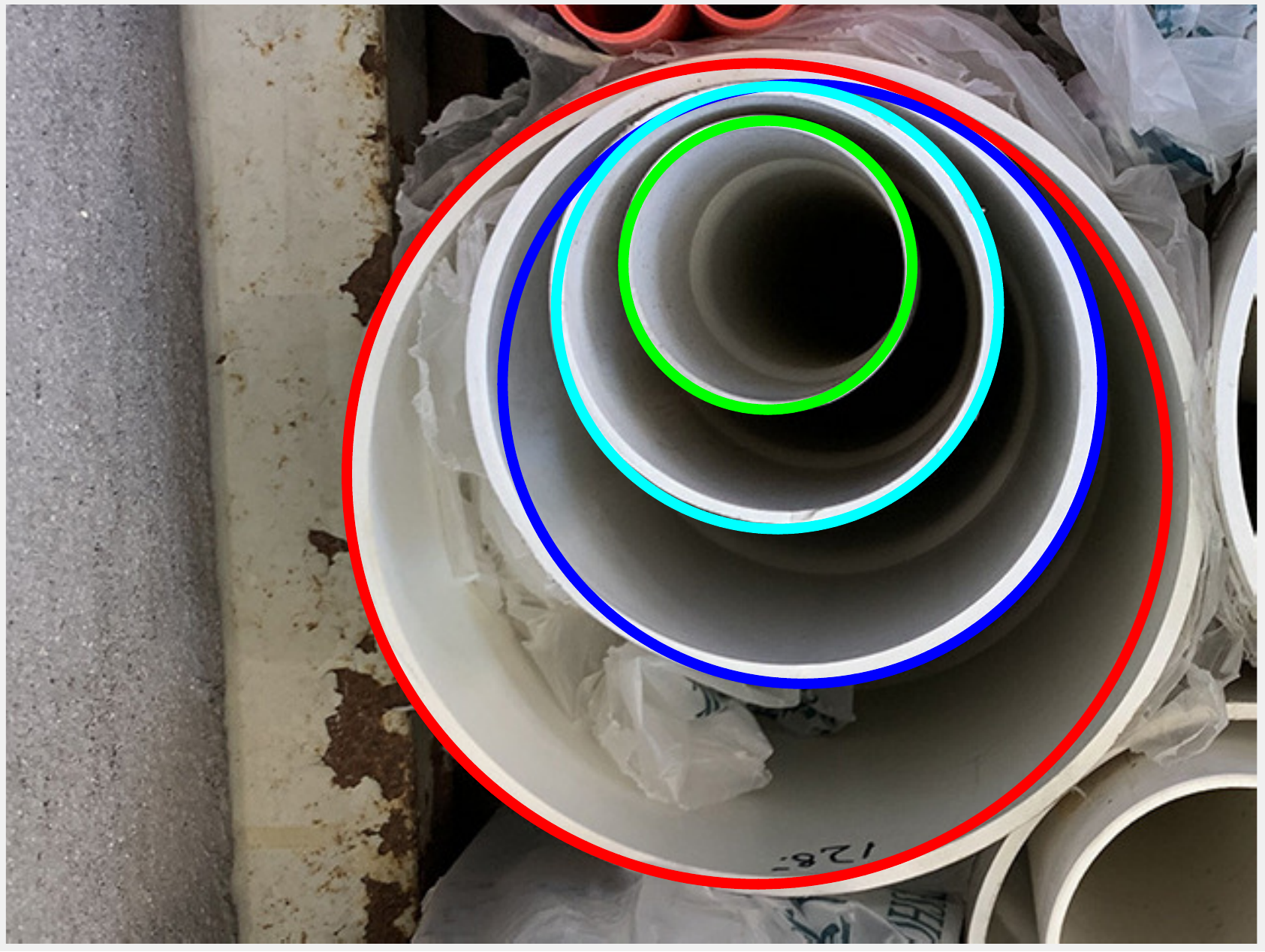}\vspace{0.2ex}
					\includegraphics[width=1.\textwidth]{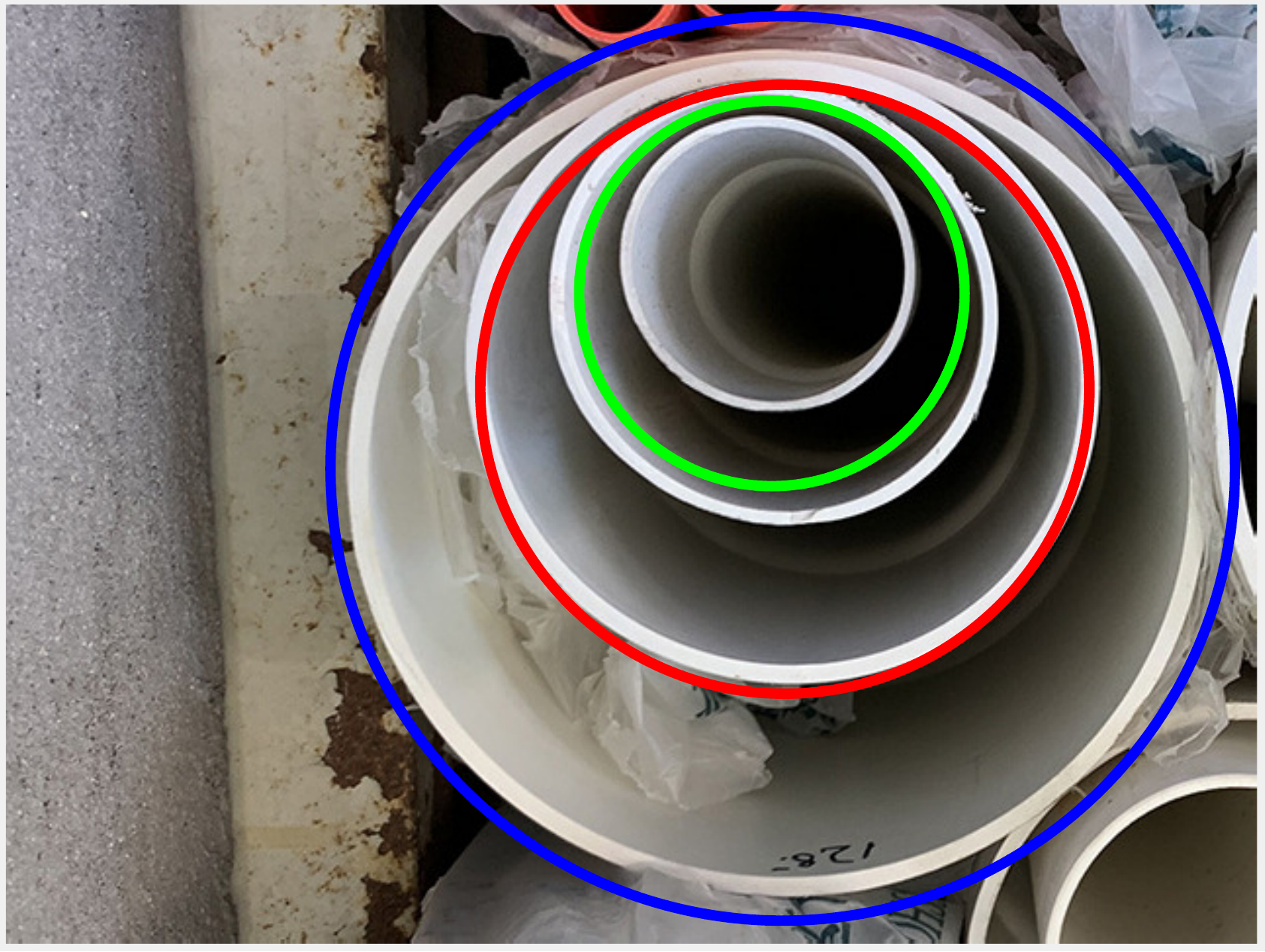}\vspace{0.2ex}
					\includegraphics[width=1.\textwidth]{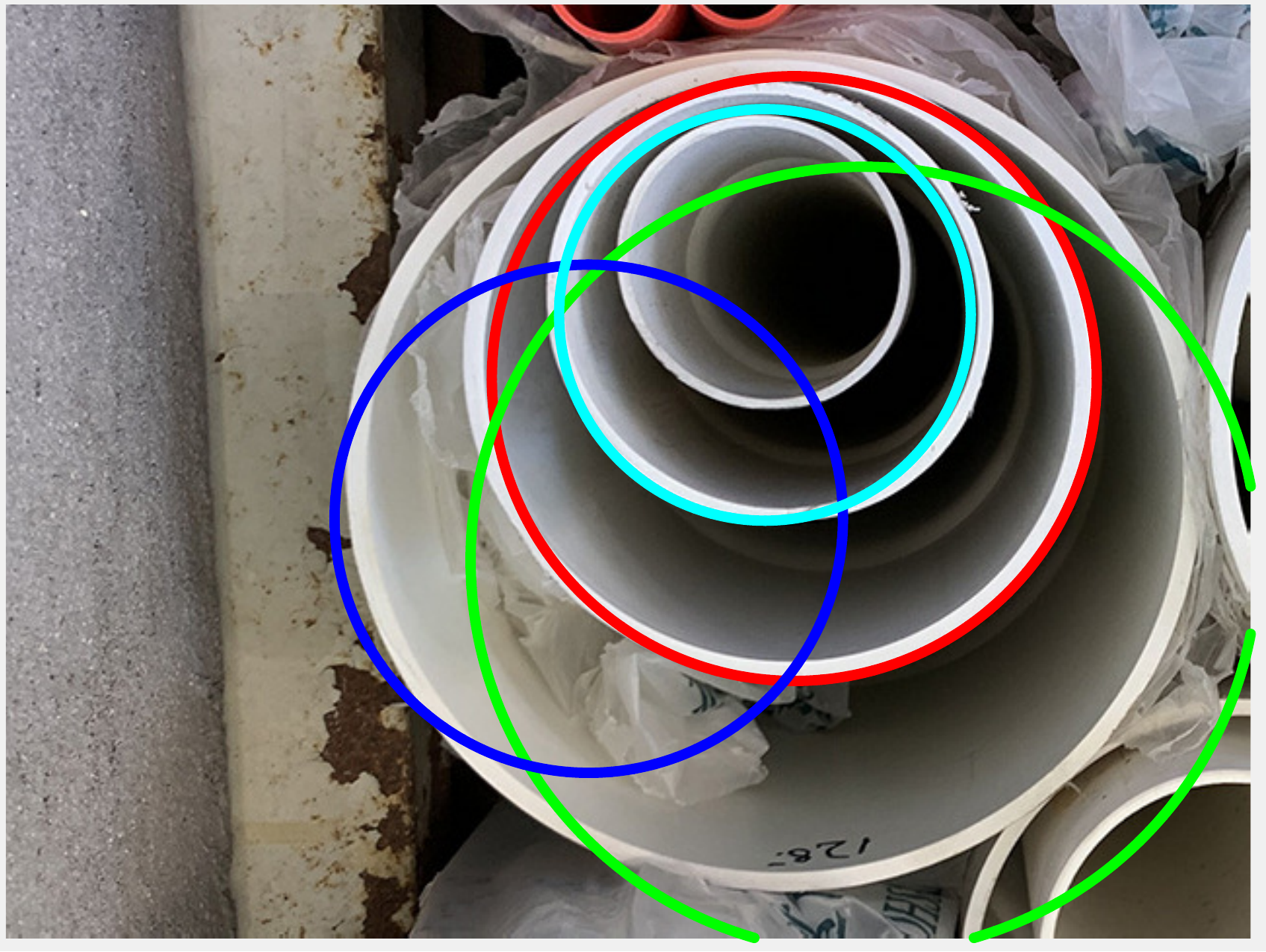}\vspace{0.2ex}
					\includegraphics[width=1.\textwidth]{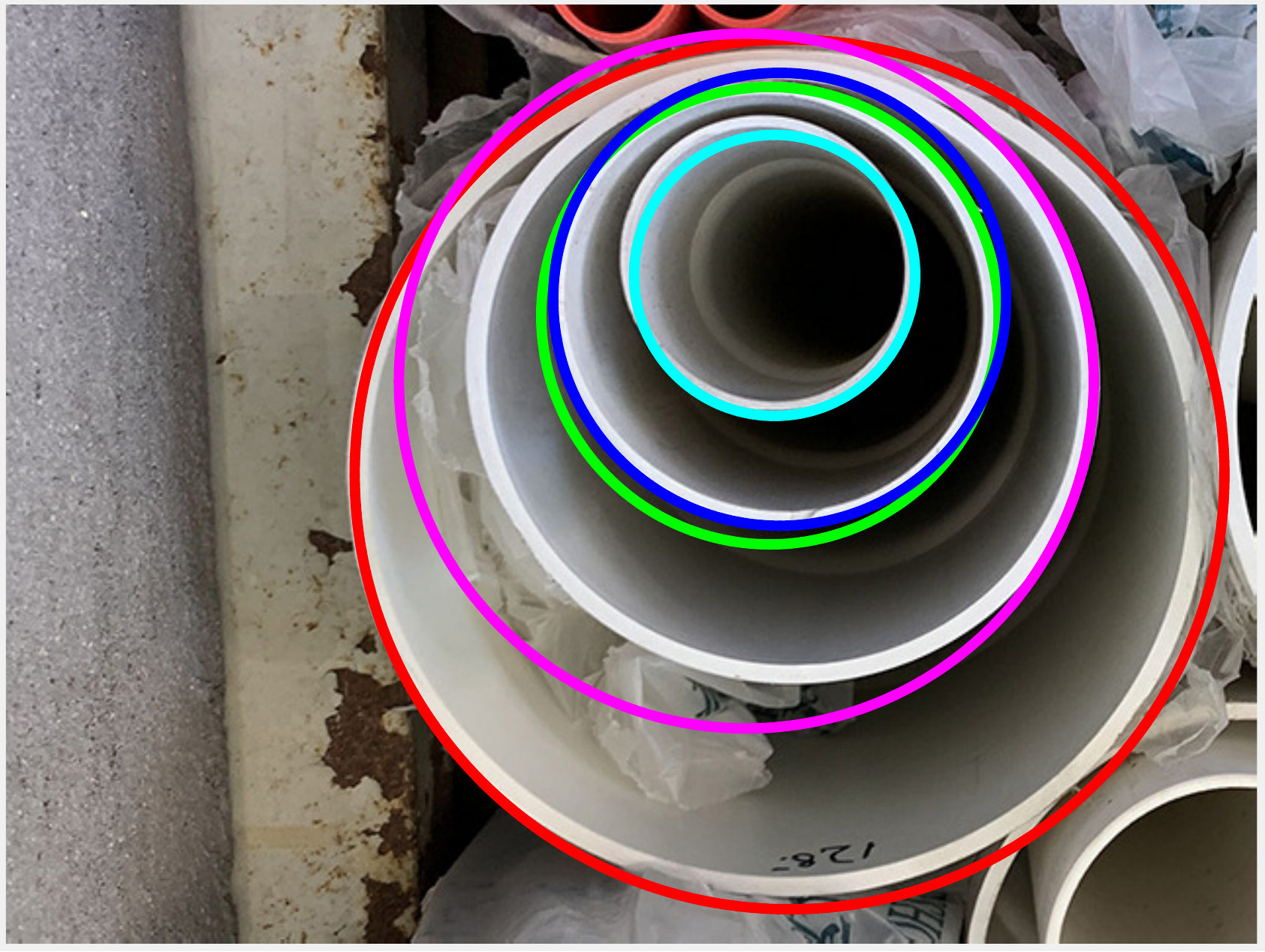}\vspace{0.2ex}
					\includegraphics[width=1.\textwidth]{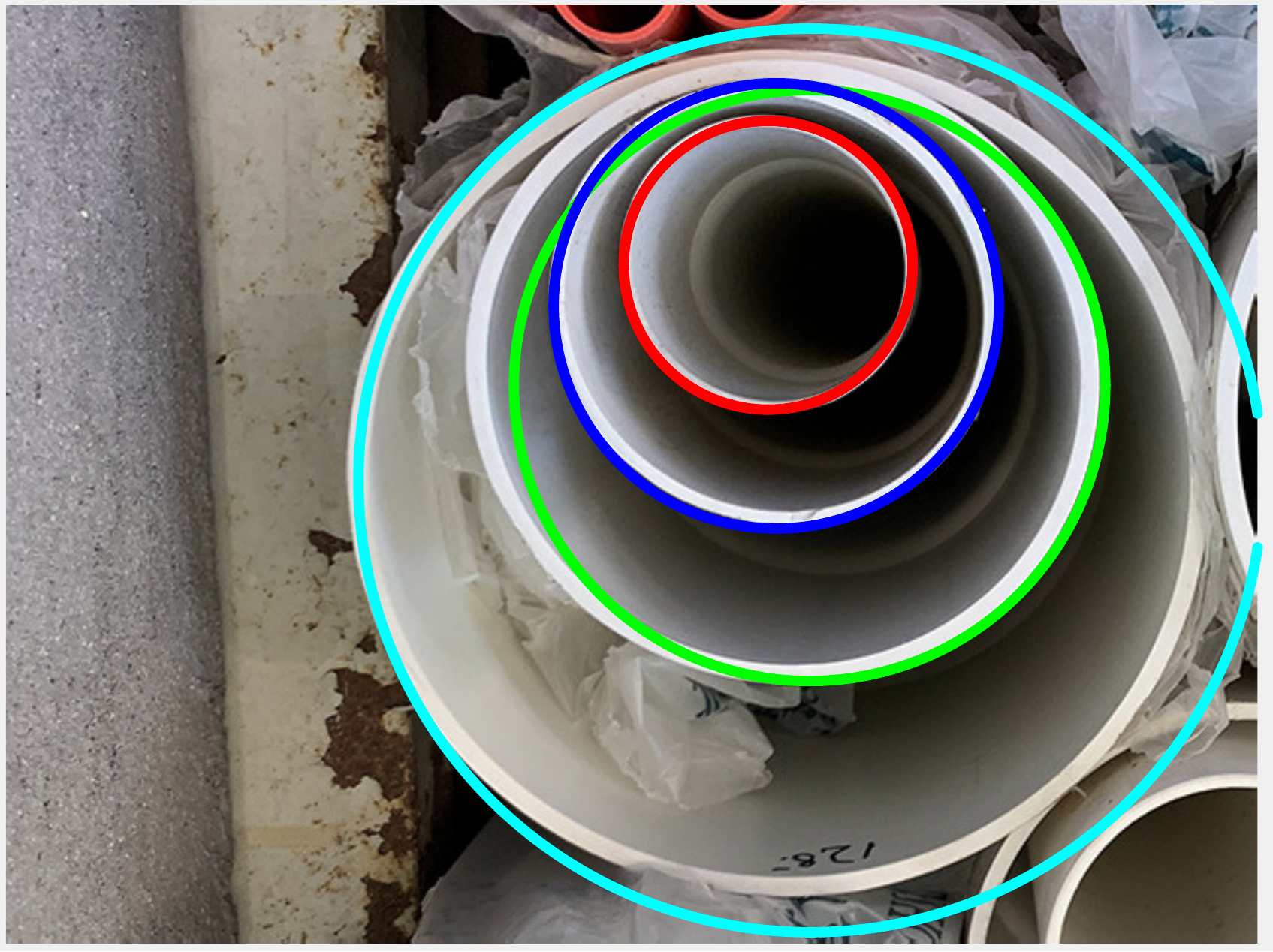}\vspace{0.2ex}
					\includegraphics[width=1.\textwidth]{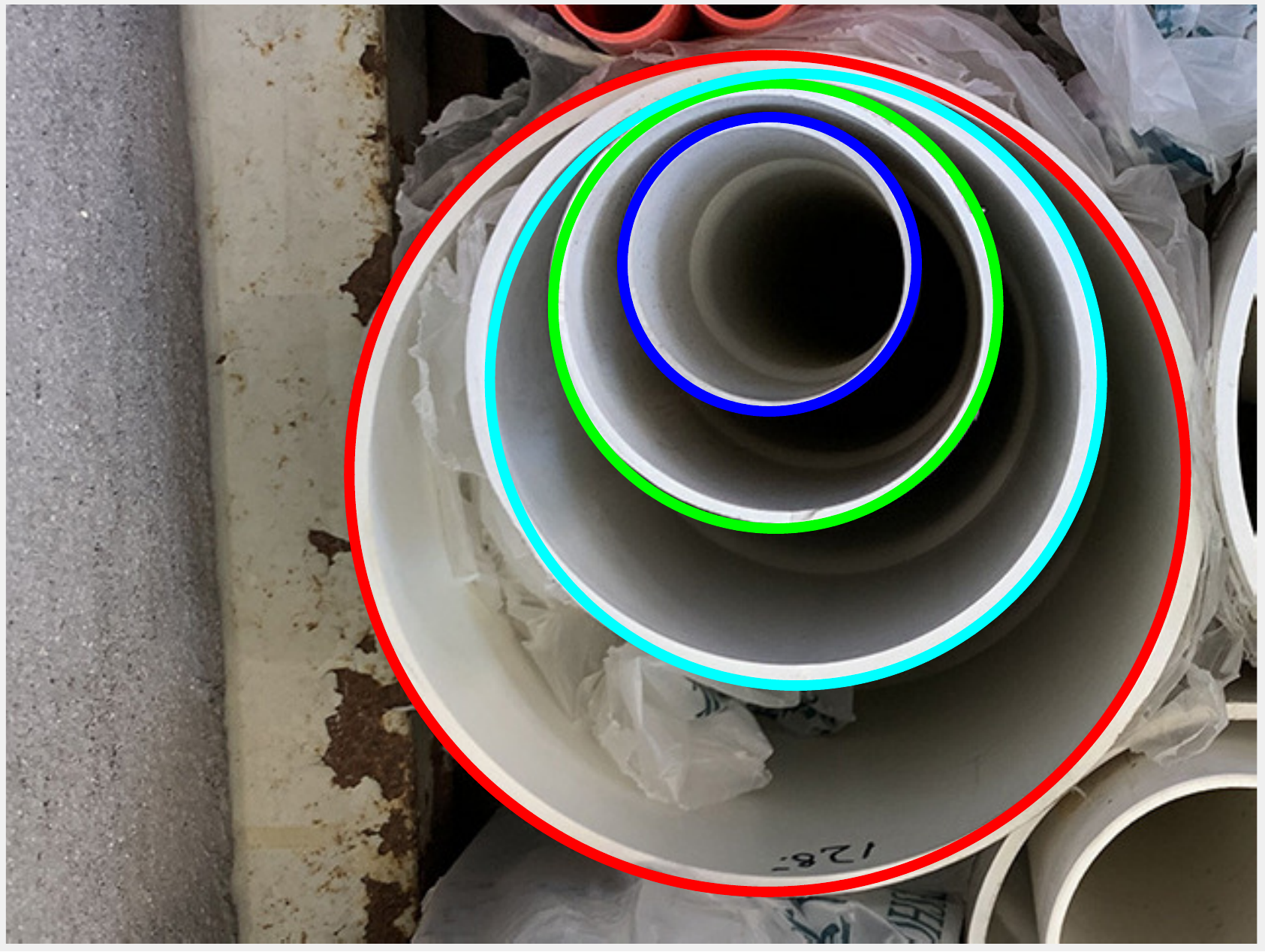}\vspace{0.2ex}
					\includegraphics[width=1.\textwidth]{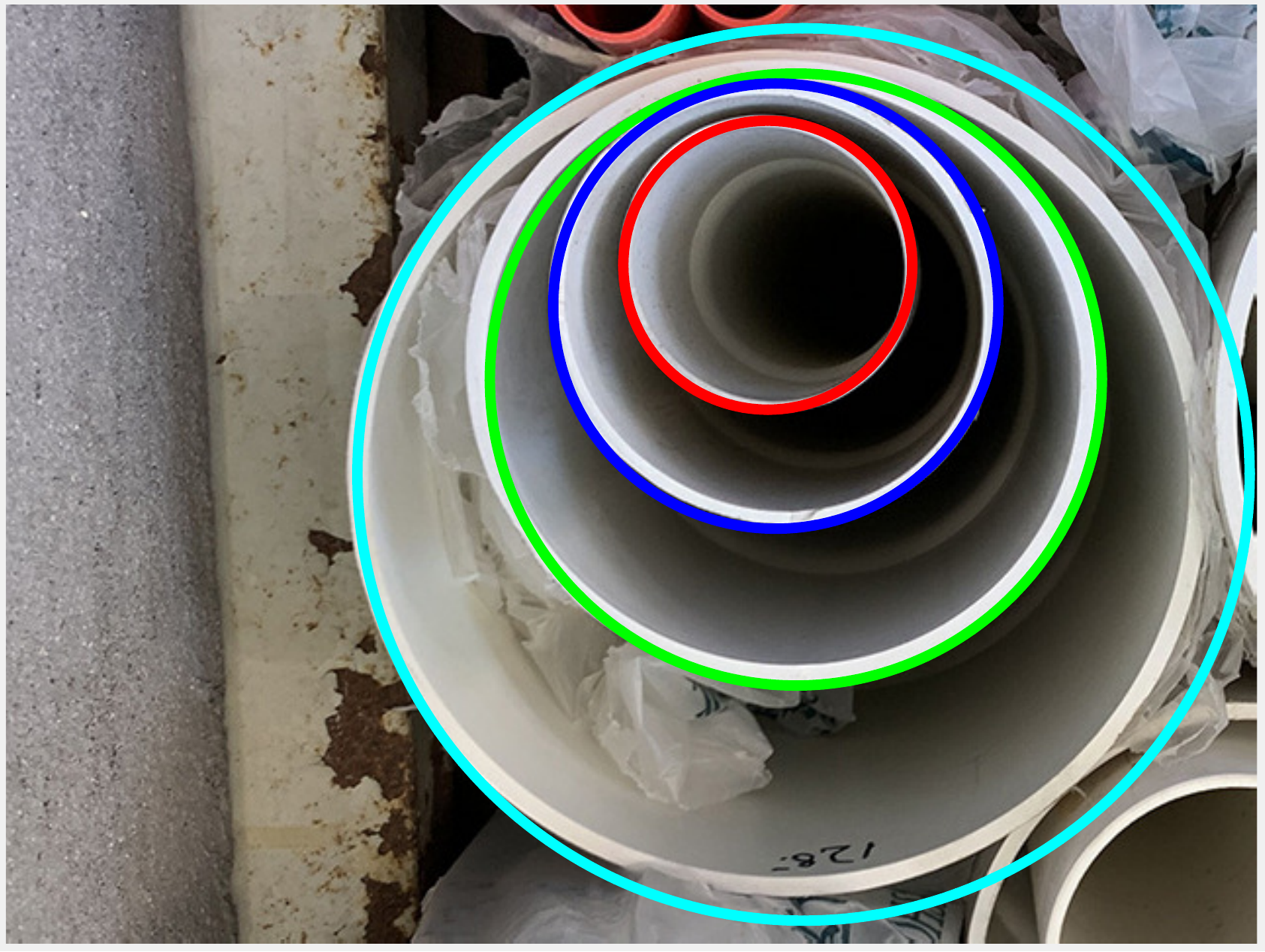}\vspace{0.2ex}
					\includegraphics[width=1.\textwidth]{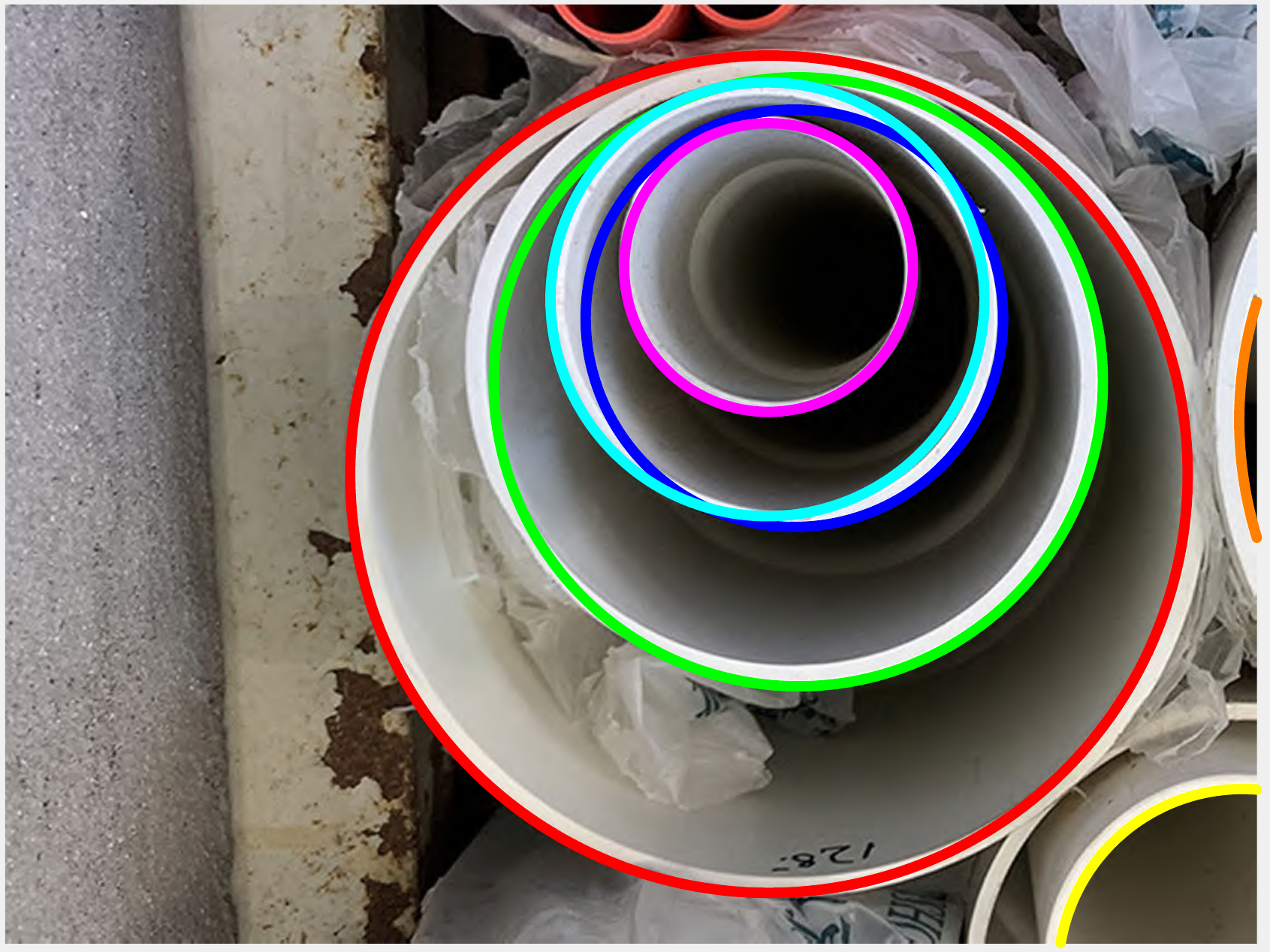}\vspace{0.2ex}
					\includegraphics[width=1.\textwidth]{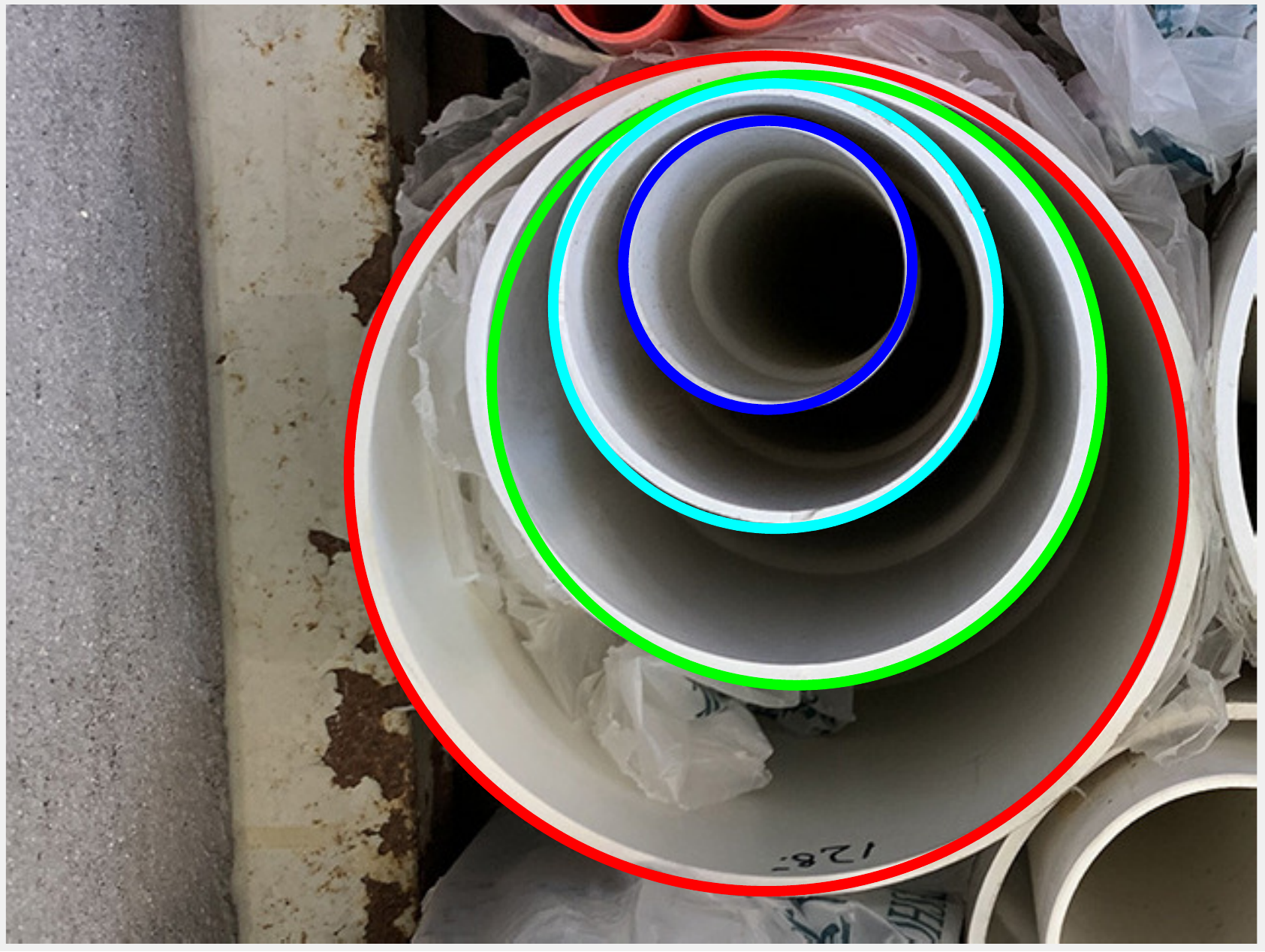}\vspace{0.2ex}
					\includegraphics[width=1.\textwidth]{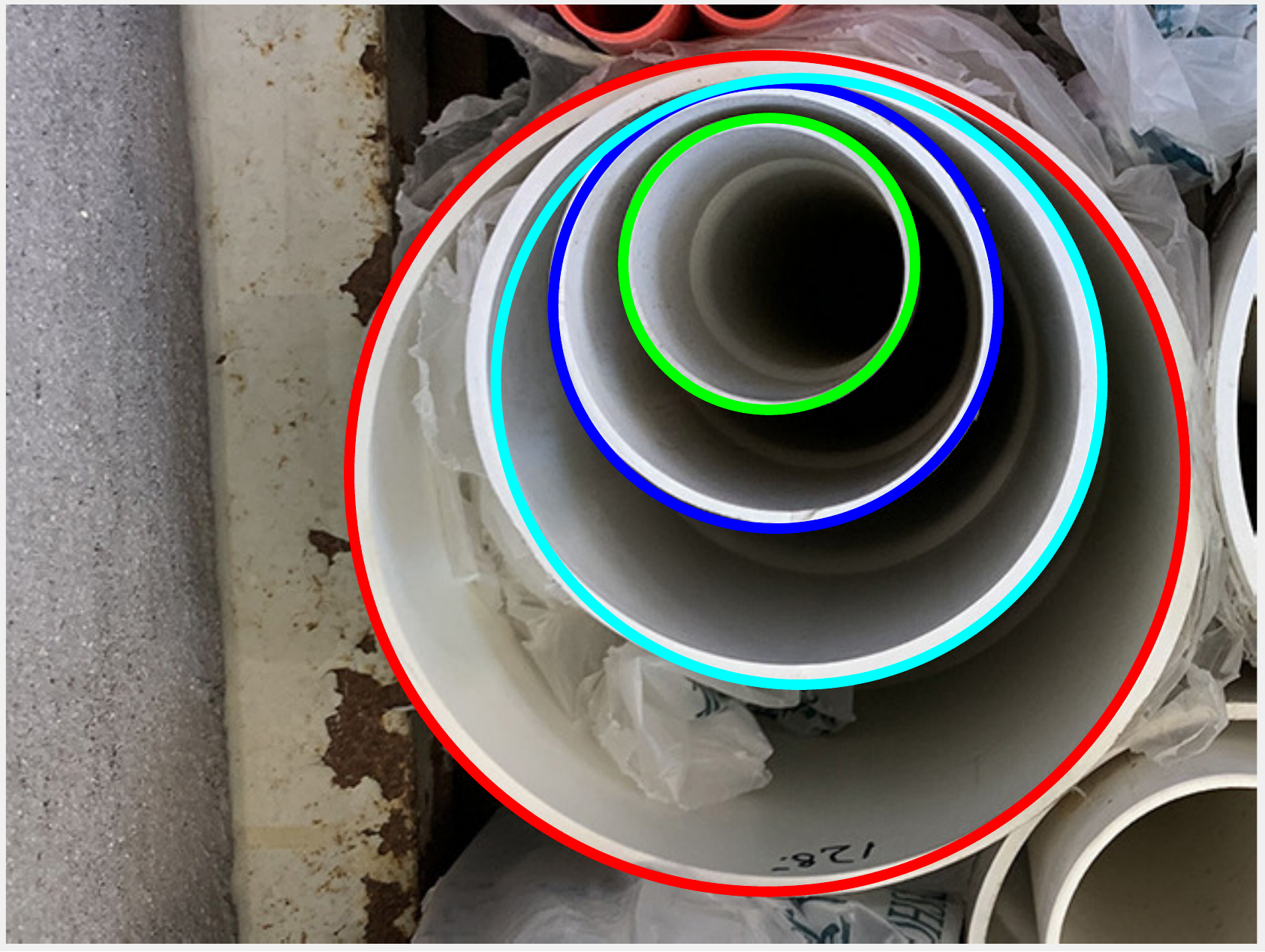}
			\end{minipage}}\hspace{-1.5ex}	
			\subfigure[Screw nuts]{
				\begin{minipage}[b]{0.125\textwidth}
					\includegraphics[width=1.\textwidth]{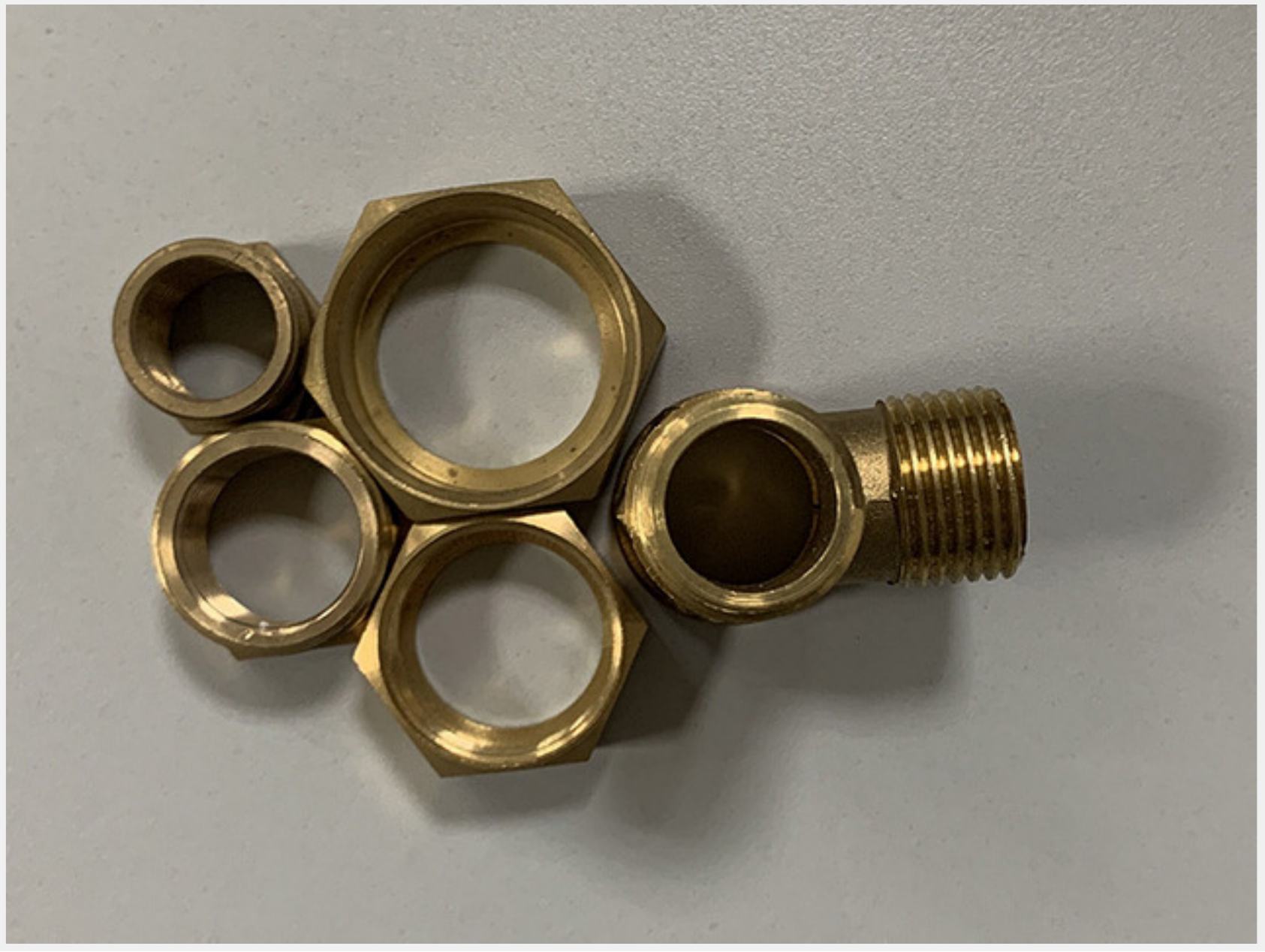}\vspace{0.2ex}
					\includegraphics[width=1.\textwidth]{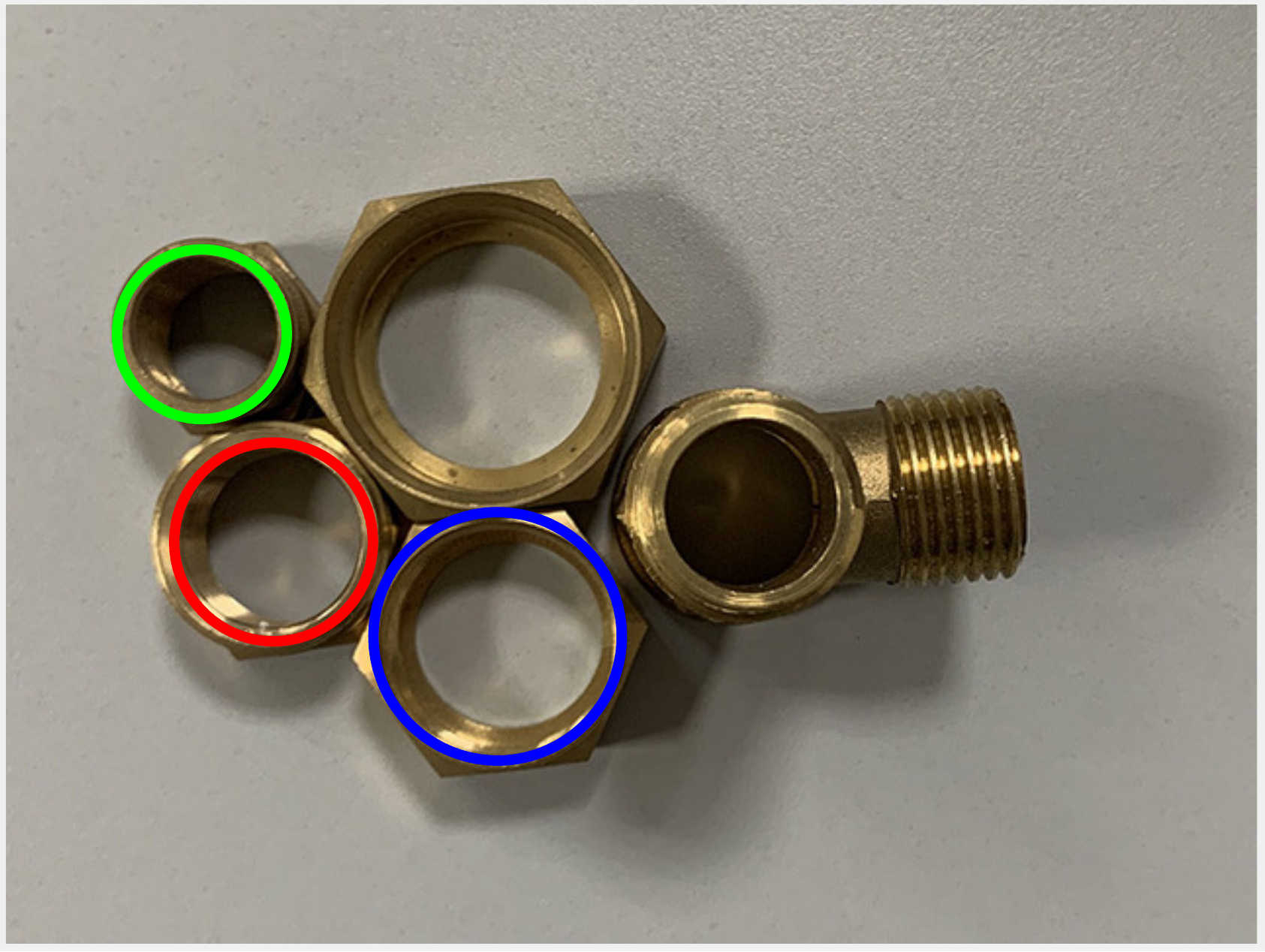}\vspace{0.2ex}
					\includegraphics[width=1.\textwidth]{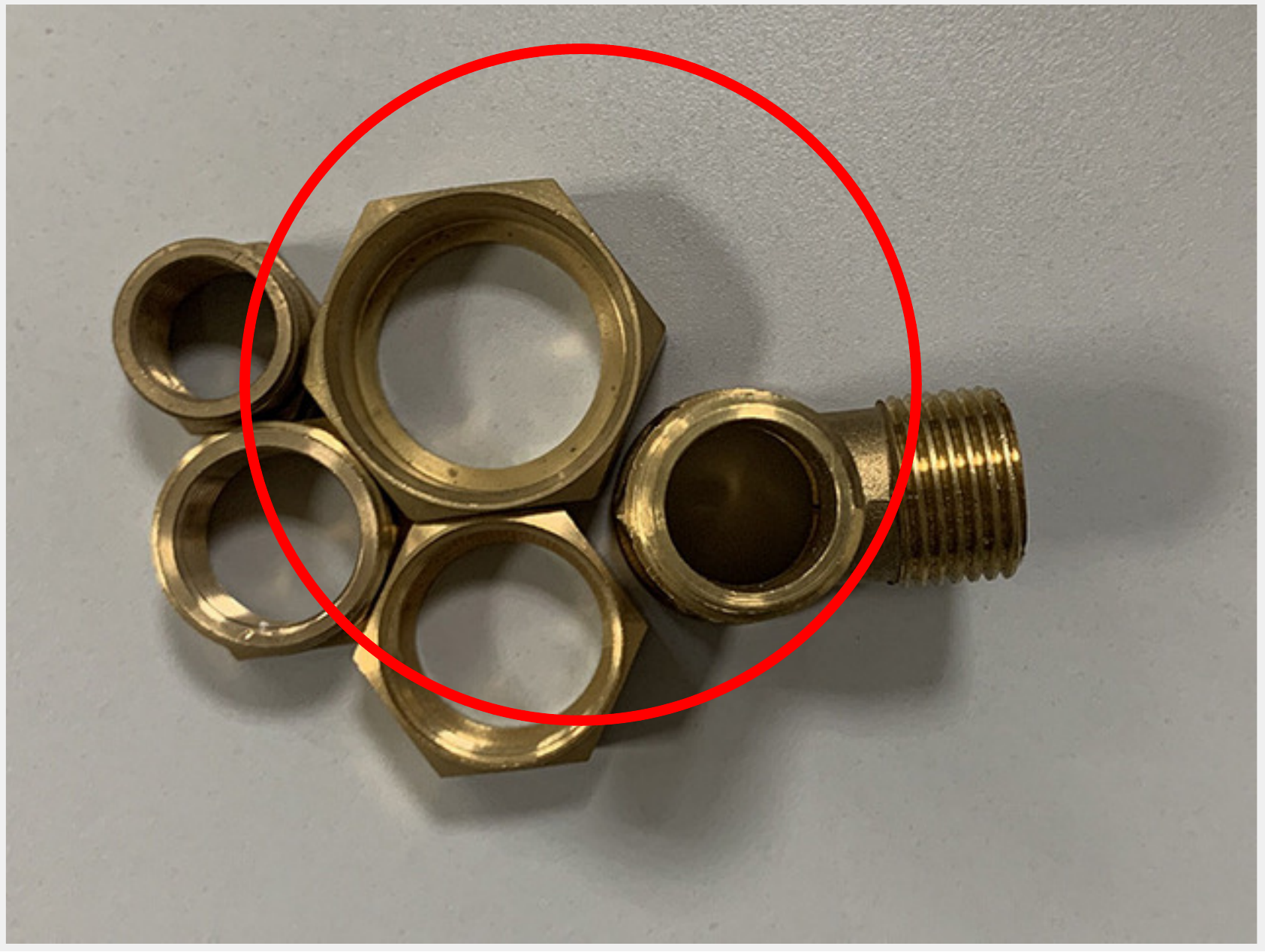}\vspace{0.2ex}
					\includegraphics[width=1.\textwidth]{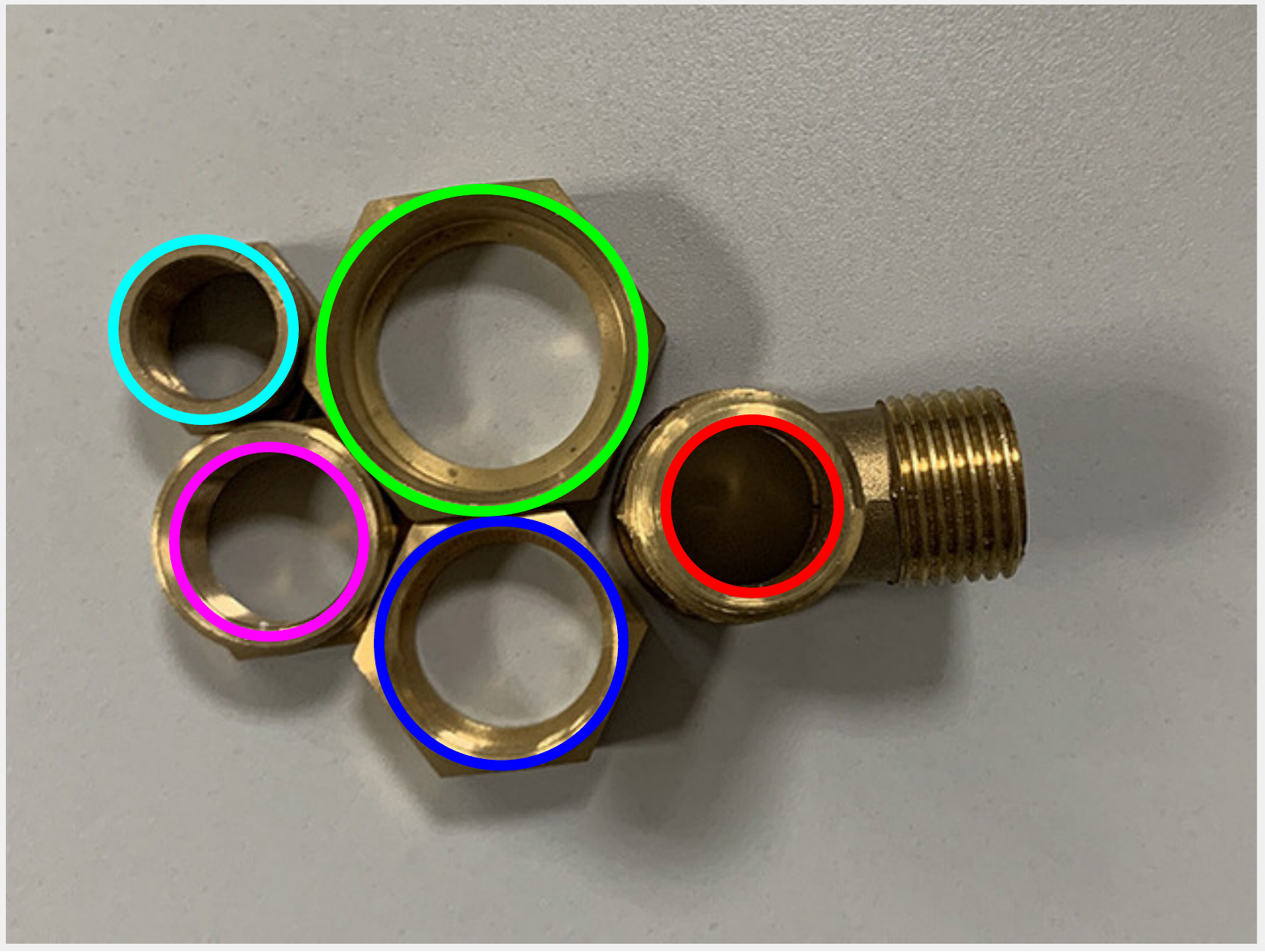}\vspace{0.2ex}
					\includegraphics[width=1.\textwidth]{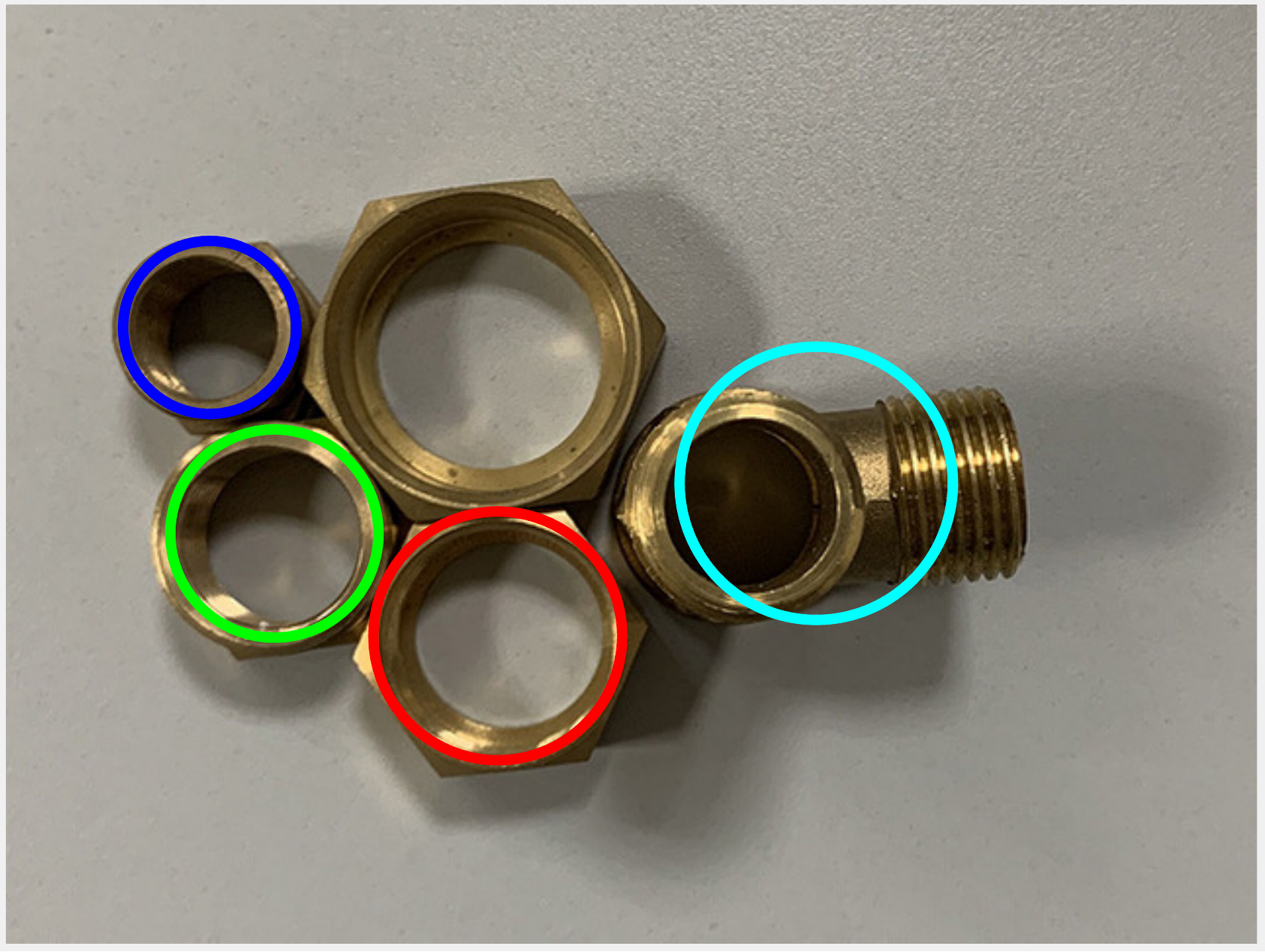}\vspace{0.2ex}
					\includegraphics[width=1.\textwidth]{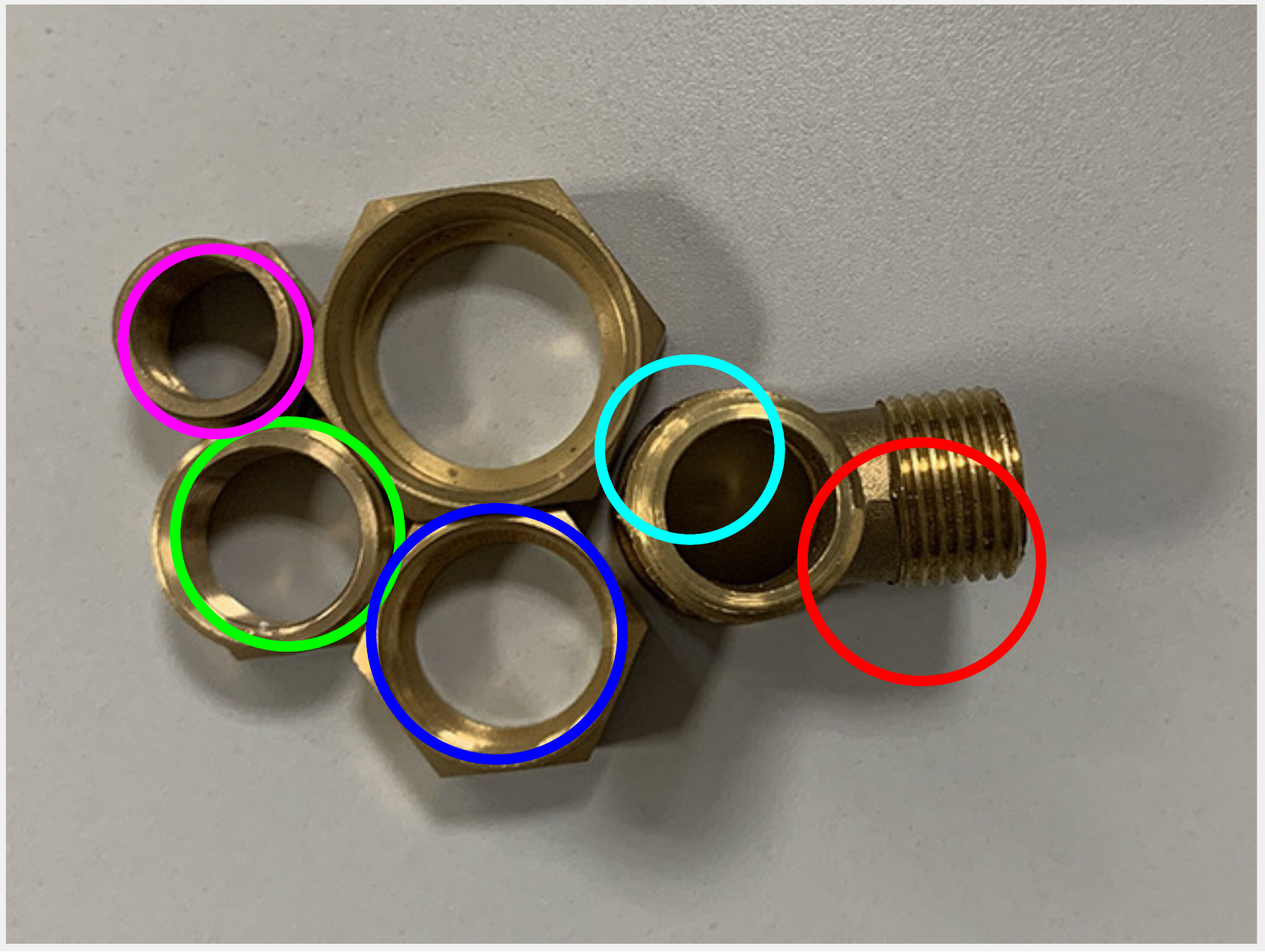}\vspace{0.2ex}
					\includegraphics[width=1.\textwidth]{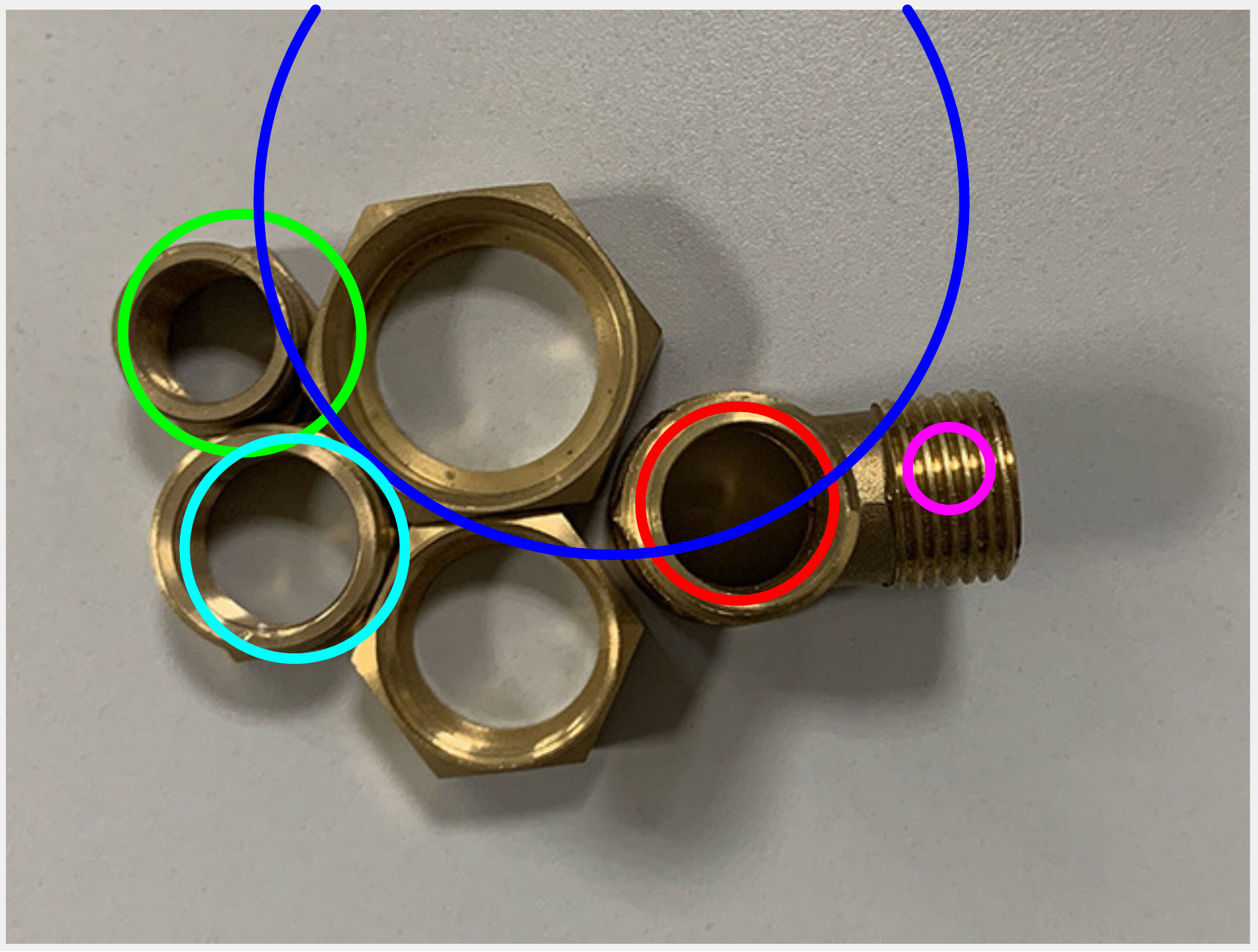}\vspace{0.2ex}
					\includegraphics[width=1.\textwidth]{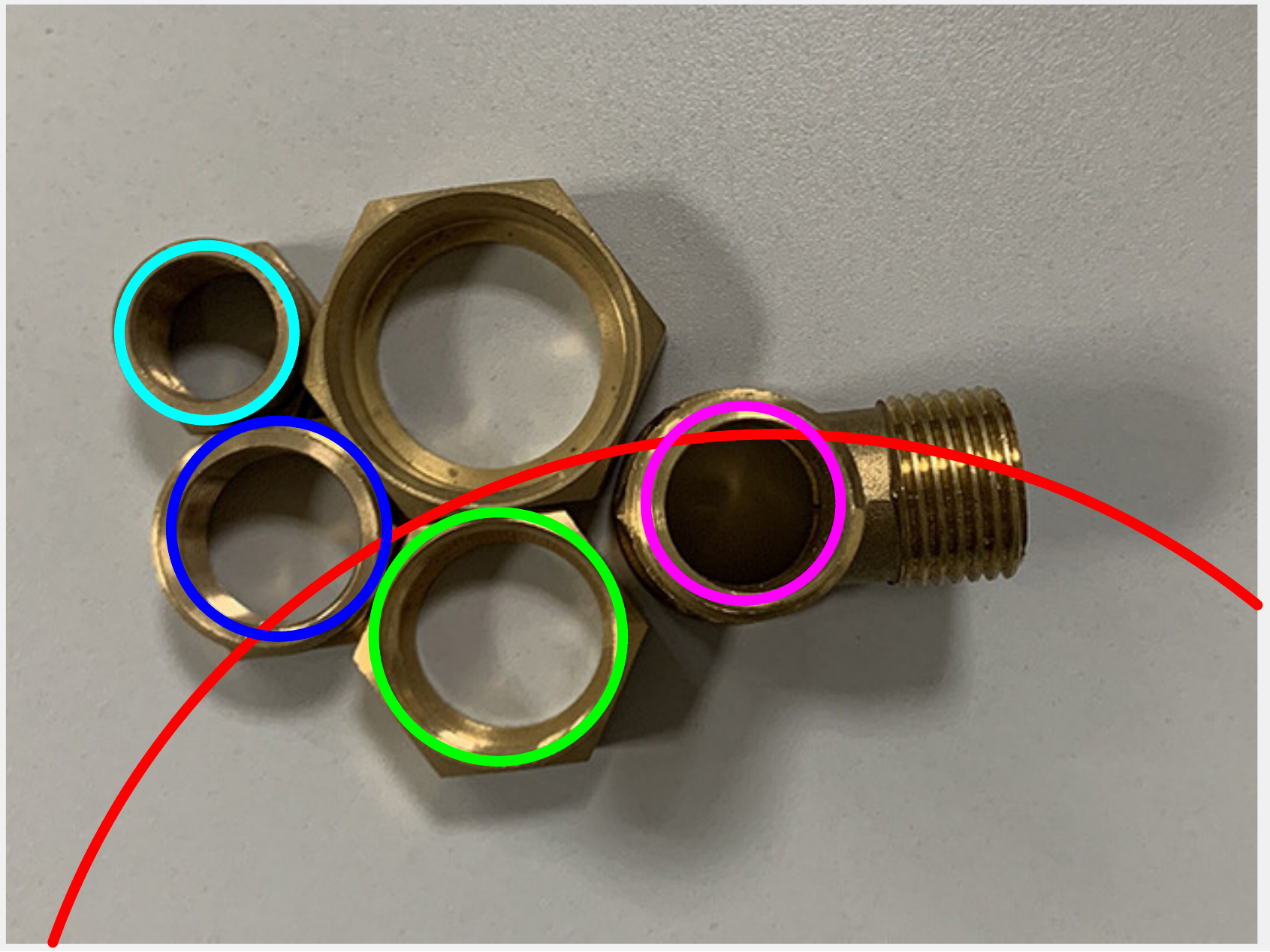}\vspace{0.2ex}
					\includegraphics[width=1.\textwidth]{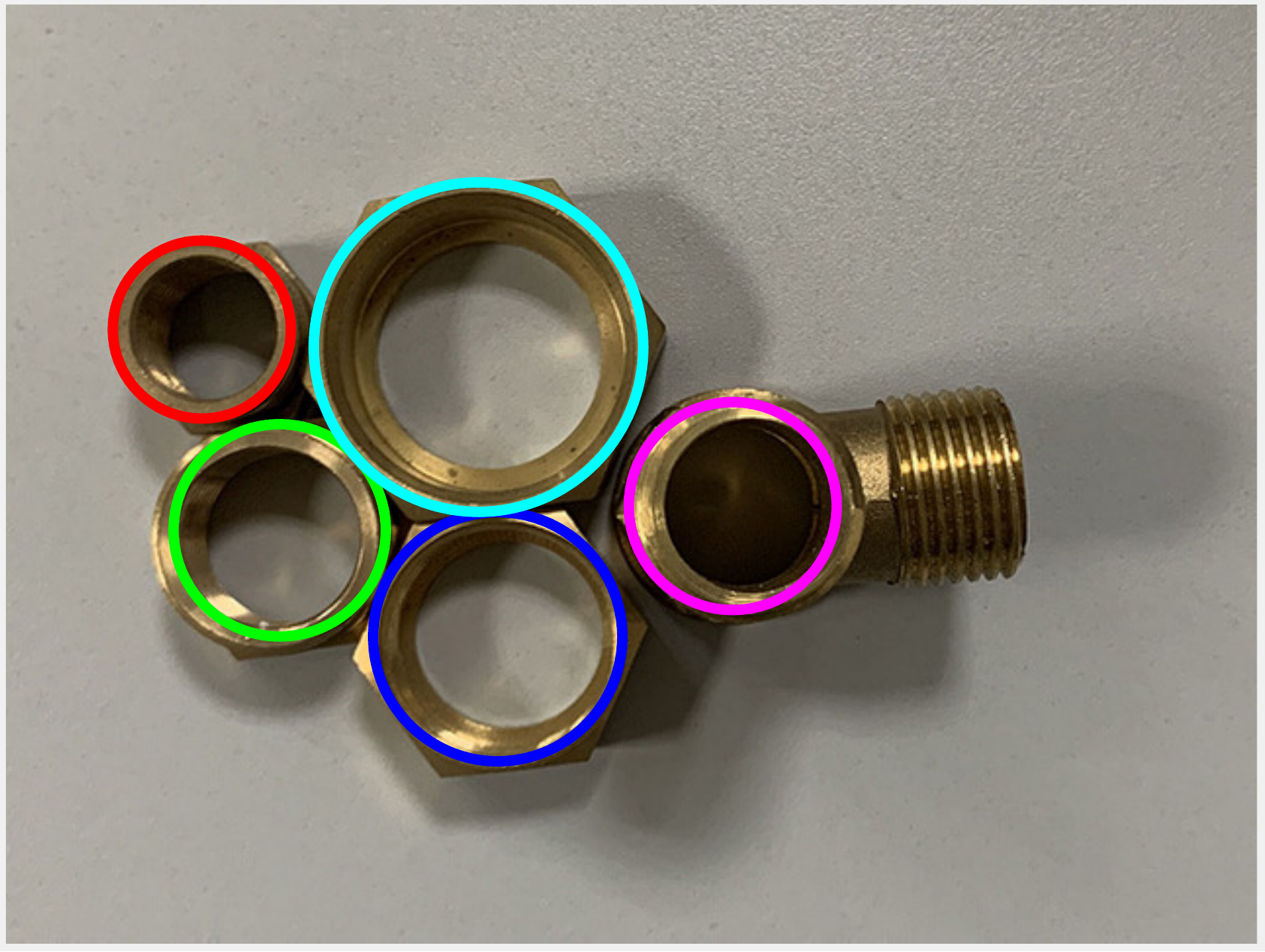}\vspace{0.2ex}
					\includegraphics[width=1.\textwidth]{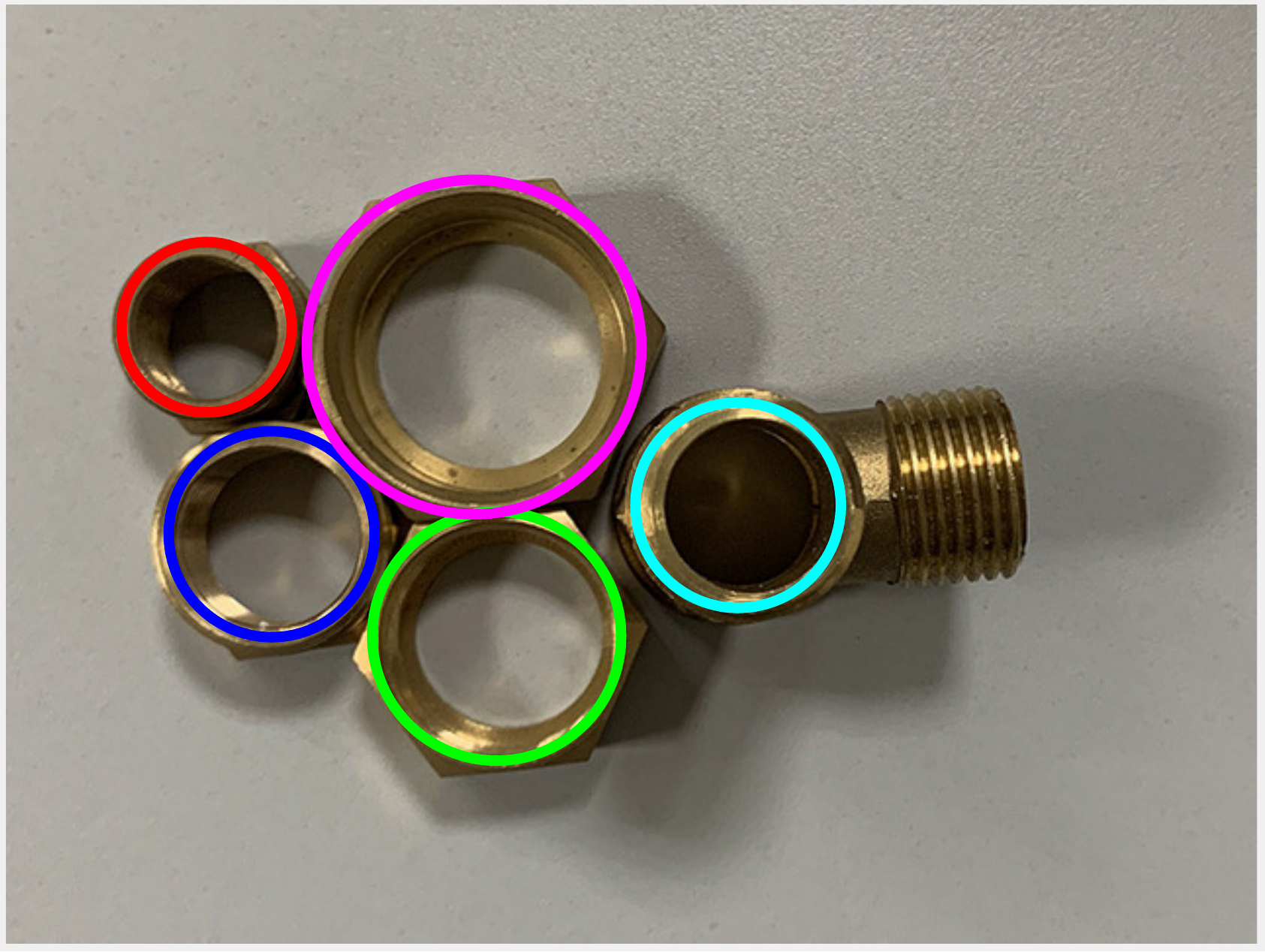}\vspace{0.2ex}
					\includegraphics[width=1.\textwidth]{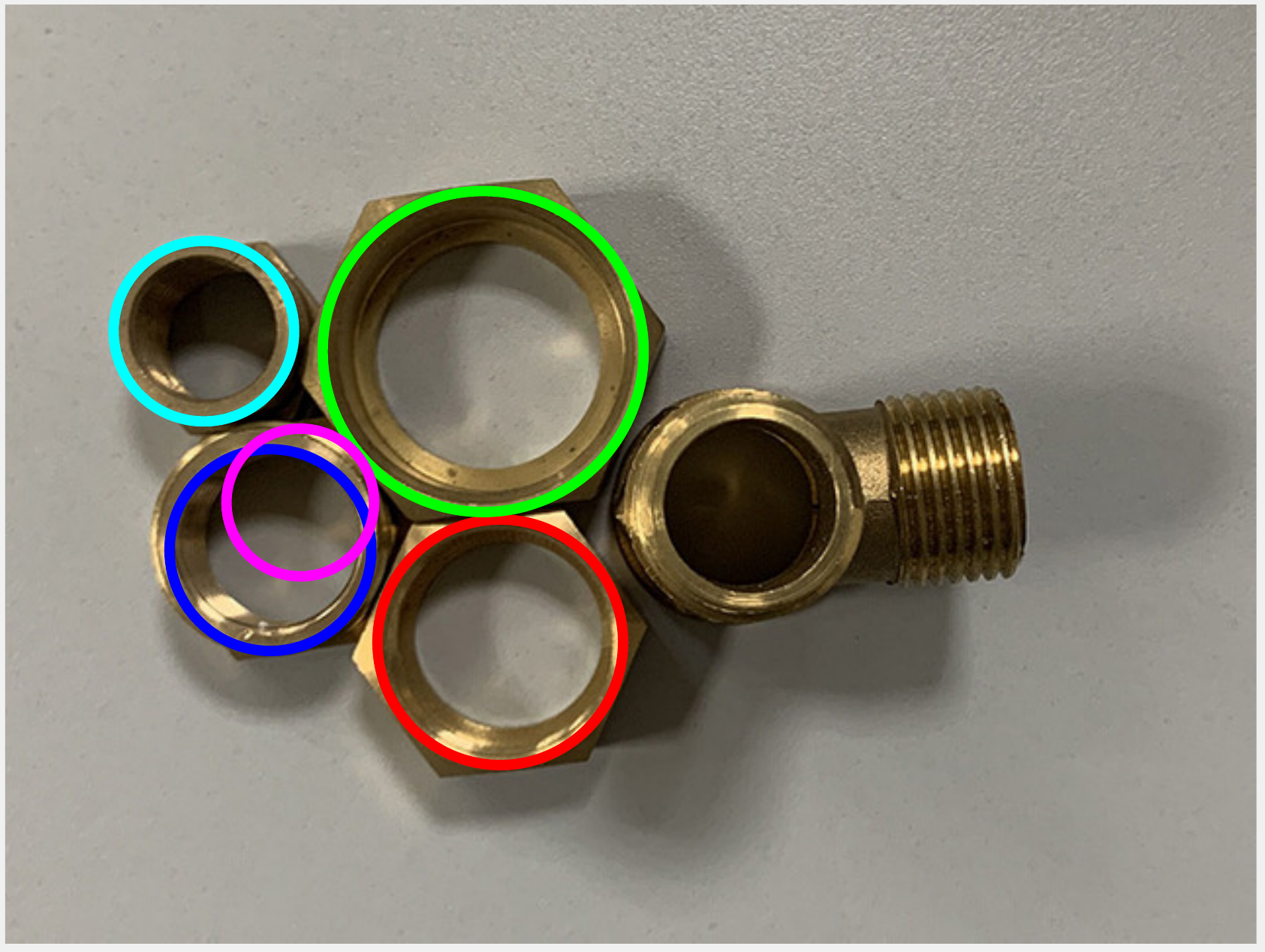}\vspace{0.2ex}
					\includegraphics[width=1.\textwidth]{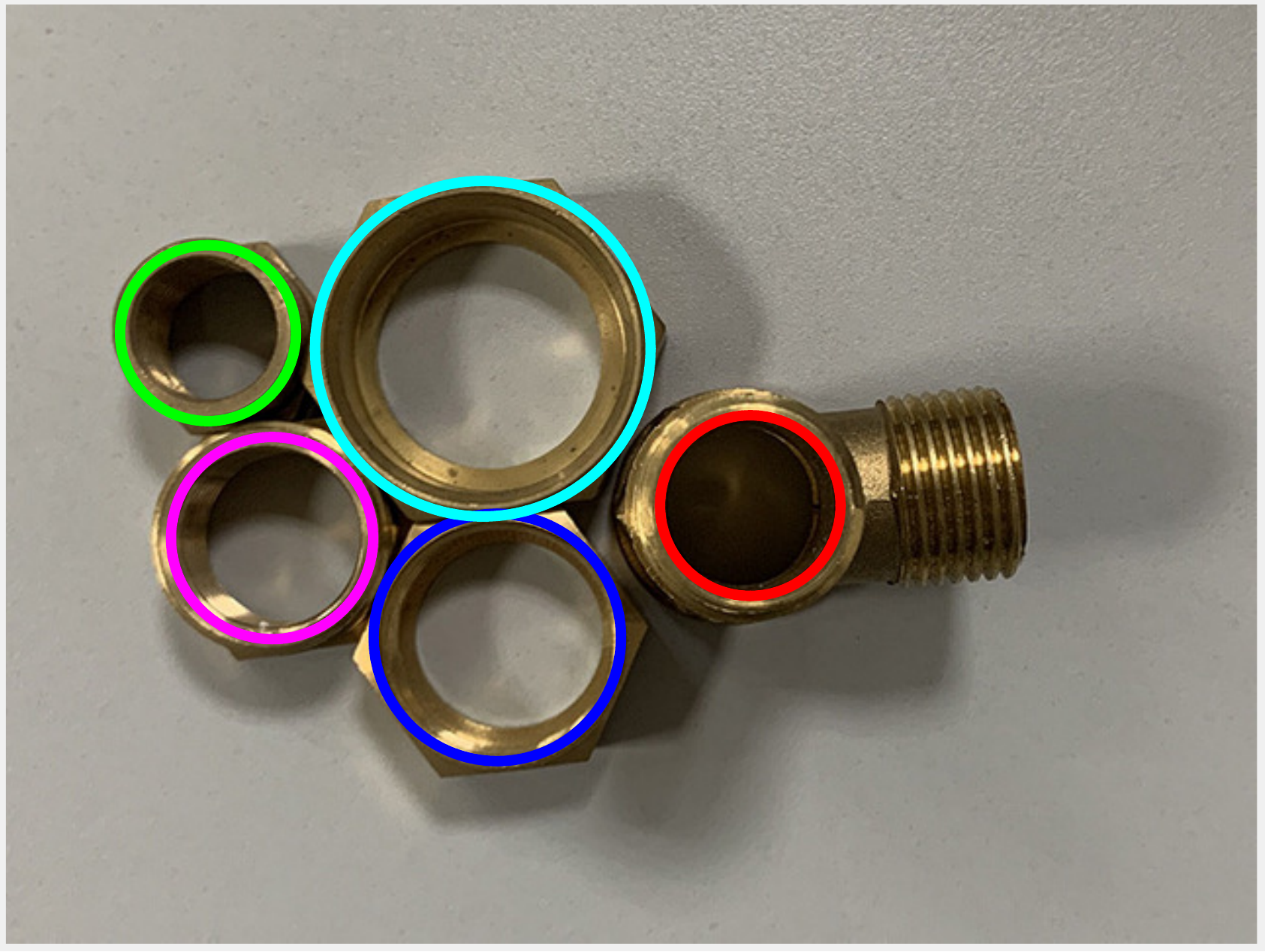}\vspace{0.2ex}
					\includegraphics[width=1.\textwidth]{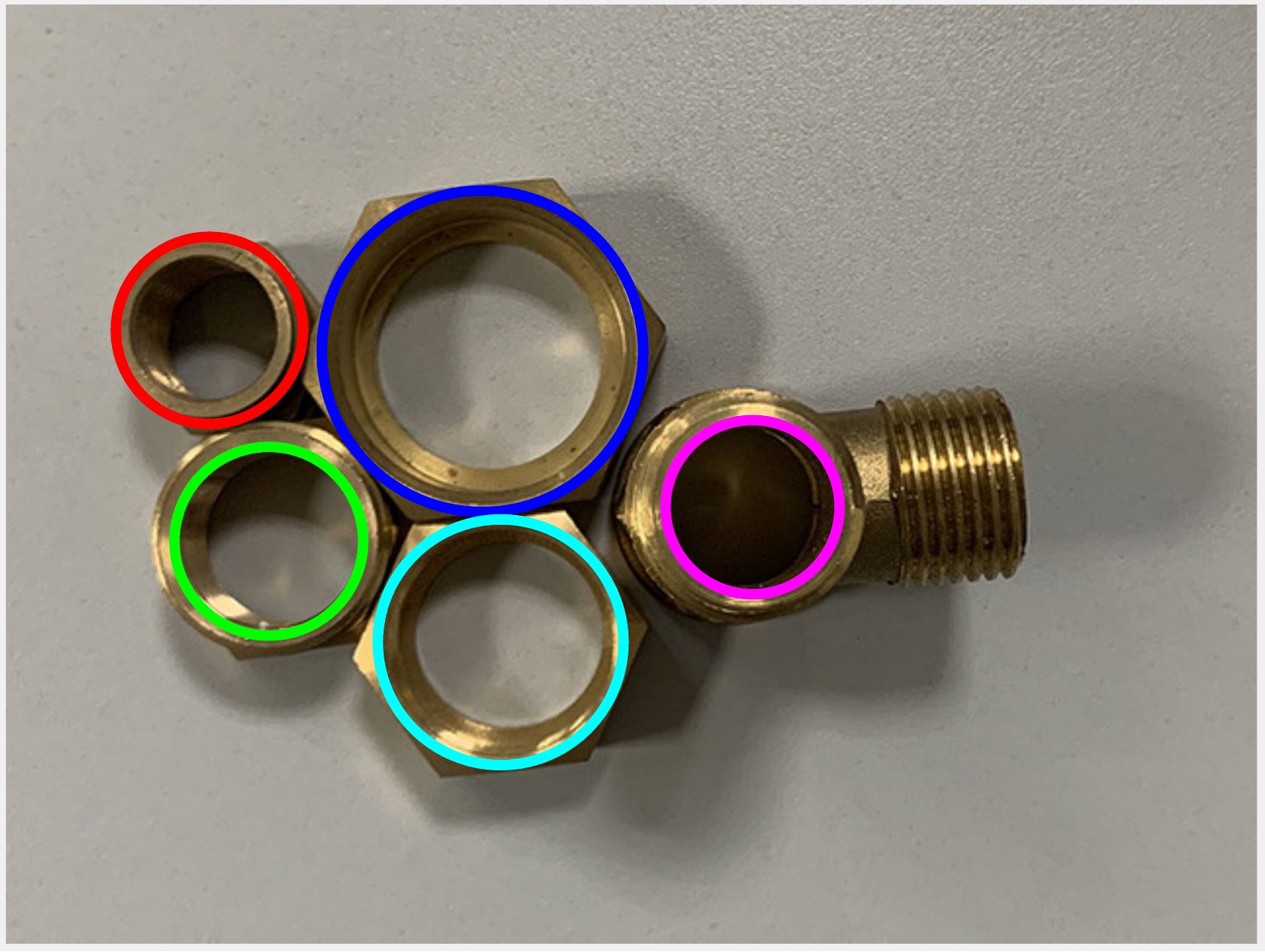}
			\end{minipage}}\hspace{-1.5ex}
			\vspace{-1ex}
			\caption{Examples for line and circle fitting on real industrial electronic images. $1^{st}$ to $13^{th}$ rows are the original images and the results obtained by J-Linkage, KF, AKSWH, T-Linkage, RPA, DPA, RansaCov, TSMP, HVF, DGSAC, MSHF and HRMP, respectively. The different model instances are marked in different colors. }
			\label{realimages}
		\end{center}
	\end{figure}
	
	Although DGSAC obtains good results for most of the real image pairs, it spends much more CPU time (about 26 times) than the proposed HRMP.
	Both J-Linkage and RPA obtain the lowest mean fitting errors in 1 out of 19 real image pairs, but RPA obtains the lower total average fitting error than J-Linkage.
	The reason behind this is that RPA effectively uses the low-rank approximation algorithm to compute the preference matrix. Among AKSWH, T-Linkage, DPA, RansaCov, TSMP and HVF, they obtain similar total average fitting errors, while AKSWH is significantly faster than T-Linkage, DPA and RansaCov (about 225, 31 and 60 times faster, respectively). We note that KF obtains the worst results for most of the real image pairs since a majority of insignificant model hypotheses may result in incorrect merging of model instances in KF. Fig. \ref{homo} shows some segmentation results obtained by HRMP on multi-homography segmentation.

	\subsection{Two-view Motion Segmentation}
	\label{TwoviewBasedMotionSegmentation}
	In this subsection, we test the performance of all the competing methods for two-view motion segmentation by using all the 19 real image pairs with single-structure and multi-structural data from the AdelaideRMF dataset. 
	The quatitative comparison results are reported in Table \ref{tableFundamental}. From Table \ref{tableFundamental}, we can see that HRMP achieves the lowest mean fitting errors in 8 out of the 19 real image pairs. HRMP also achieves the lowest total median fitting error and the fastest total average running time among all the competing methods, although its total average fitting error is slightly higher than DGSAC (HRMP is more than 70 times faster than DGSAC considering the total average CPU time).
	Both RPA and MSHF achieve better results than DPA for most of image pairs. TSMP and RansaCov achieve similar total average fitting errors, but TSMP is about 69 times faster than RansaCov.
	J-Linkage, KF, and T-Linkage achieve similar total average fitting errors, while AKSWH obtains the second fastest total average CPU time among all the competing methods. AKSWH and HVF fail in fitting most of the real image pairs since they are based on the parameter space, which is sensitive to the data distribution. Fig. \ref{fun} shows some motion segmentation results obtained by HRMP. In short, the proposed HRMP obtains significant superiority over the other competing methods for most image pairs.
	
	\begin{table*}[!t]
		\vspace{-7ex}
		\caption{The CPU time used by the twelve fitting methods (in seconds). \# indicates the true number of model instances included in data. The best results are boldfaced.}
		\label{realimages_tab}
		\vspace{-5.5ex}
		\begin{center}
			\scalebox{0.8}{\tabcolsep0.15in
				\begin{tabular}{crrrrrrrrrrrr}
					\toprule
					{Data} & \multicolumn{1}{c}{{J-Linkage}} & \multicolumn{1}{c}{{KF}} & \multicolumn{1}{c}{{AKSWH}} & \multicolumn{1}{c}{{T-Linkage}} & \multicolumn{1}{c}{{RPA}} & \multicolumn{1}{c}{{DPA}} & \multicolumn{1}{c}{{RansaCov}} & \multicolumn{1}{c}{{TSMP}} & \multicolumn{1}{c}{{HVF}} & \multicolumn{1}{c}{{DGSAC}} & \multicolumn{1}{c}{{MSHF}} & \multicolumn{1}{c}{{HRMP}} \\
					\midrule
					\multicolumn{1}{l}{{Masonry nails (\#4)}} & {6.63} & {5.55} & {25.94} & {301.25} & {37.31} & {41.66} & {8.83} & {4.97} & {11.48} & {3.01} & {15.45} & {\textbf{2.36}} \\
					\multicolumn{1}{l}{{Power lines (\#8)}} & {15.17} & {39.39} & {4.45} & {1413.14} & {319.22} & {128.56} & {15.84} & {48.66} & {14.53} & {3.35} & {9.64} & {\textbf{3.05}} \\
					\multicolumn{1}{l}{{PVC pipes (\#4)}} & {8.89} & {21.53} & {3.31} & {863.80} & {64.47} & {76.75} & {16.27} & {17.69} & {7.44} & {3.25} & {3.89} & {\textbf{2.80}} \\
					\multicolumn{1}{l}{{Screw nuts (\#5)}} & {12.41} & {14.42} & {4.61} & {1021.03} & {92.92} & {83.08} & {18.22} & {10.95} & {7.39} & {3.40} & {13.03} & {\textbf{3.25}} \\
					\midrule
					{Total mean time} & {10.77} & {20.22} & {9.58} & {899.80} & {128.48} & {82.51} & {14.79} & {20.57} & {10.21} & {3.25} & {10.50} & {\textbf{2.86}} \\
					\bottomrule
				\end{tabular}
			}
		\end{center}
		\vspace{-4.8ex}
	\end{table*}

	\vspace{-2ex}
	\subsection{Applications Oriented Towards Industrial Electronics}
	In this subsection, we evaluate the performance of all the 12 competing methods for line fitting and circle fitting by using 4 representative real industrial electronic images: i.e., the ``Masonry nails", ``Power lines", ``PVC pipes" and ``Screw nuts" images, which respectively include four lines, eight lines, four circles and five circles. These four images contain 393, 840, 631 and 674 feature points extracted by the Canny operator \cite{canny1986computational}, respectively. 
	As shown in Fig. \ref{realimages} and Table \ref{realimages_tab}, only AKSWH, MSHF and the proposed HRMP correctly fit all the model instances (i.e., the lines and the circles). AKSWH and MSHF determine the inliers of model instances by using an inlier noise scale estimator in the line and circle fitting, but they require higher CPU time than HRMP. J-Linkage misses some model instances in each image due to its sensitivity to pseudo-outliers.
	Although T-Linkage successfully fits all the eight power lines of the ``Power lines" image, it is very time-consuming, especially when the model hypotheses contain a large number of data points. KF fails in fitting the ``Screw nuts" image because it wrongly removes the inliers belonging to the other model instances. RPA and DPA also miss some model instances in each image, respectively. Moreover, RPA needs more time to construct and process the affinity matrix based on the Cauchy weighting function than DPA for the ``Power lines" and the ``Screw nuts" images. RansaCov succeeds in fitting all the four nails of the ``Masonry nails" image within a reasonable time. TSMP, HVF and DGSAC correctly estimate the five, five and four nuts of the ``Screw nuts" image, respectively. However, DGSAC achieves the second fastest total average CPU time among all the competing methods. It is worth noting that DGSAC cannot effectively select one real structure of multiple model instances for the ``Masonry nails" image and it also misses some model instances of the ``Power lines" image. This is because that the model selection used by DGSAC is sensitive to the disparity between inliers and outliers. Overall, the proposed HRMP performs favorably against all the competitors in terms of both accuracy and speed.

	\section{Conclusion}
	\label{sec:Conclusion}
	In this paper, we present a novel hierarchical representation via message propagation (HRMP) method, which combines the merits of both the consensus analysis and the preference analysis for robust geometric model fitting. Moreover, a new hierarchical message propagation (HMP) algorithm is proposed based on the hierarchical representation. On one hand, the consensus information and the preference information are propagated to effectively prune the two layers of the hierarchical representation. On the other hand, the improved affinity propagation (IAP) algorithm is used to cluster the remaining data points by employing the Tanimoto-like similarity in the pruned data point layer. The two parts of the proposed HRMP are tightly coupled to improve the accuracy and speed of model fitting. In addition, HRMP alleviates the sensitivity to gross outliers by the message propagation, and it can handle multi-structural data corrupted by a great number of outliers. Experimental results on both synthetic data and real images show that the proposed HRMP significantly improves the efficiency and robustness for geometric model fitting and it outperforms several state-of-the-art model fitting methods.

	\bibliographystyle{Bibliography/IEEEtranTIE}
	\bibliography{Bibliography/IEEEabrv,Bibliography/BIB_xx-TIE-xxxx}\ 
	
	\vspace{-2cm}
	\begin{IEEEbiography}[{\includegraphics[width=1in,height=1.25in,clip,keepaspectratio]{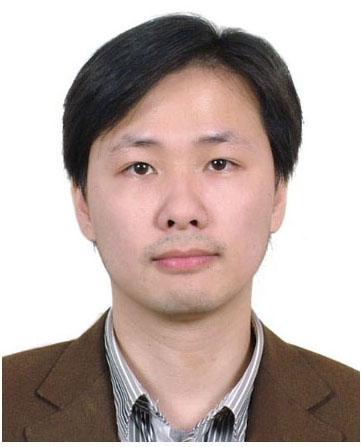}}]
		{Shuyuan Lin} is currently working toward the Ph.D. degree in the School of Informatics, Xiamen University, Xiamen, China. His research interests mainly focus on geometric model fitting, video analysis and motion detection. His research interests also focus on computer vision and machine learning.
	\end{IEEEbiography}
	
	\vspace{-1.5cm}
	\begin{IEEEbiography}[{\includegraphics[width=1in,height=1.25in,clip,keepaspectratio]{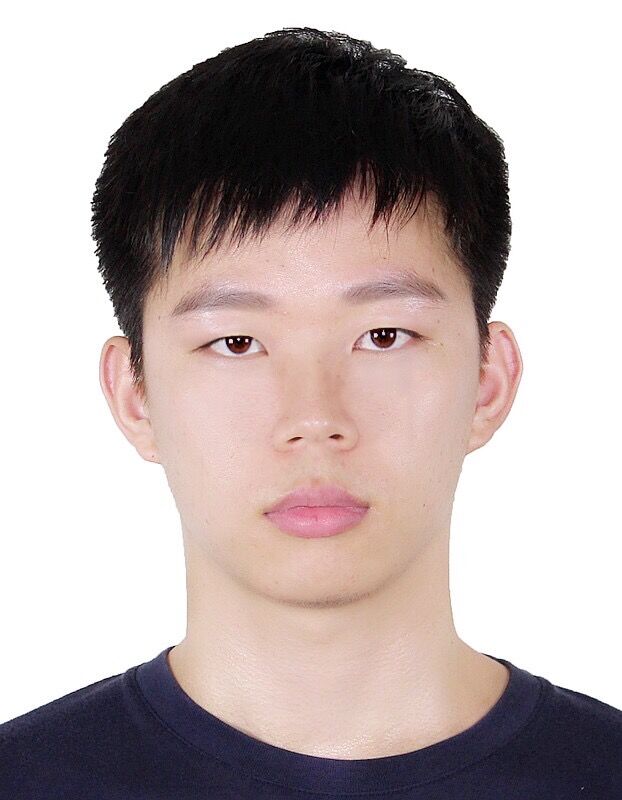}}]
		{Xing Wang} received the M.S. degree from Xiamen University, China, in 2016. His research interests mainly focus on geometric model fitting, video analysis and motion detection. His research interests also focus on computer vision and machine learning.
	\end{IEEEbiography}	
	
	\vspace{-1.5cm}
	\begin{IEEEbiography}[{\includegraphics[width=1in,height=1.25in,clip,keepaspectratio]{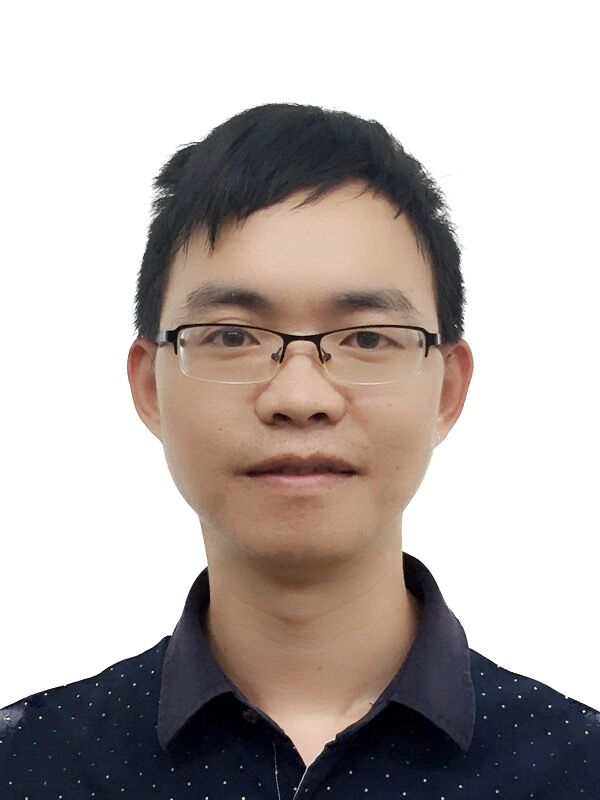}}]
		{Guobao Xiao} received the M.S. degree in information and computing science from Fujian Normal University, China, in 2013 and the Ph.D. degree in Computer Science and Technology from Xiamen University, China, in 2016. From 2016-2018, he was a Postdoctoral Fellow in the School of Aerospace Engineering at Xiamen University, China. He is currently a professor at Minjiang University, China. He has published over 30 papers in the international journals and conferences. His research interests include machine learning, computer vision, pattern recognition and bioinformatics.
	\end{IEEEbiography}		
	
	\vspace{-1.5cm}
	\begin{IEEEbiography}[{\includegraphics[width=1in,height=1.25in,clip,keepaspectratio]{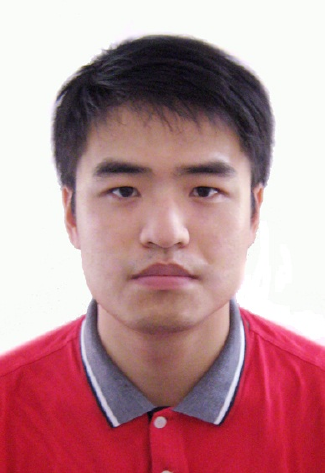}}]
		{Yan Yan} received the Ph.D. degree	in information and communication engineering from Tsinghua University, China, in 2009. From 2009 to 2010, he worked as a Research Engineer with the Research and Development Center, Nokia Japan, and a Project Leader with Panasonic Singapore Lab in 2011. He is currently an Associate Professor with the School of Informatics, Xiamen University,	China. He has published over 60 papers in international journals and conferences. His research interests include computer vision and pattern recognition.
	\end{IEEEbiography}
	
	\vspace{-1.5cm}
	\begin{IEEEbiography}[{\includegraphics[width=1in,height=1.25in,clip,keepaspectratio]{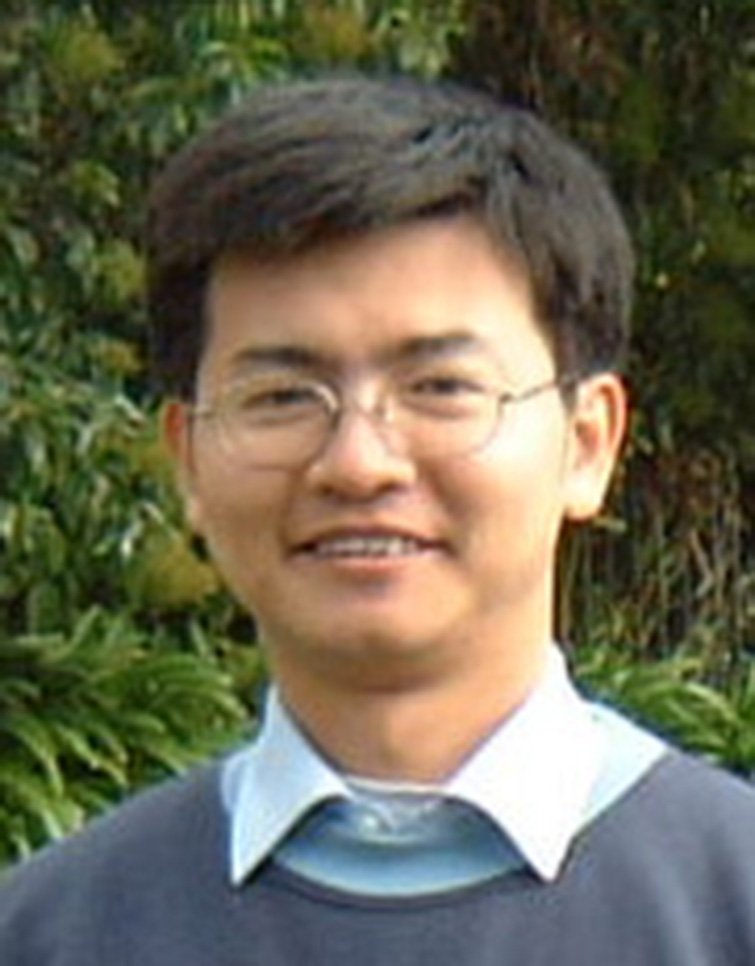}}]
		{Hanzi Wang} received the Ph.D. (Hons.) degree in Computer Vision from Monash University, Australia. He is currently a Distinguished Professor of "Minjiang Scholars" in Fujian province and a Founding Director of the Center for Pattern Analysis and Machine Intelligence (CPAMI) at Xiamen University in China. His research interests are concentrated on computer vision and pattern recognition including visual tracking, robust statistics, object detection, video segmentation, model fitting, optical flow calculation, 3D structure from motion, image segmentation and related fields.
	\end{IEEEbiography}

\end{document}